\documentclass[10pt,journal,compsoc]{IEEEtran}

\usepackage[T1]{fontenc}
\usepackage{amsmath,amssymb,amsfonts}
\usepackage{amsthm}
\usepackage{mathtools}

\usepackage{graphicx}
\usepackage{tikz}
\usetikzlibrary{shapes.geometric,arrows.meta,positioning,fit,backgrounds,calc,patterns,decorations.pathreplacing}
\usepackage{pgfplots}
\pgfplotsset{compat=1.17}

\usepackage{array}
\usepackage{booktabs}
\usepackage{multirow}
\usepackage{makecell}
\usepackage{tabularx}
\usepackage{colortbl}
\newcolumntype{P}[1]{>{\raggedright\arraybackslash}p{#1}}

\usepackage[caption=false,font=footnotesize,labelfont=sf,textfont=sf]{subfig}
\usepackage{textcomp}
\usepackage{stfloats}
\usepackage{url}
\usepackage{balance}
\usepackage{enumitem}
\usepackage{xcolor}
\usepackage[nocompress]{cite}
\usepackage{pifont}

\newcommand{\cmark}{\textcolor{green!50!black}{\ding{51}}}
\newcommand{\halfmark}{\textcolor{orange!90!black}{$\triangle$}}
\newcommand{\xmark}{\textcolor{red!75!black}{\ding{55}}}

\usepackage[colorlinks=true,citecolor=blue!60!black,linkcolor=blue!60!black,urlcolor=blue!60!black]{hyperref}

\newtheorem{definition}{Definition}
\newtheorem{theorem}{Theorem}
\newtheorem{proposition}{Proposition}

\newtheorem{remark}{Remark}

\newcommand{\X}{\mathcal{X}}
\newcommand{\Y}{\mathcal{Y}}
\newcommand{\D}{\mathcal{D}}
\newcommand{\Hcal}{\mathcal{H}}
\newcommand{\Ps}{P_{S}}
\newcommand{\Pt}{P_{T}}
\newcommand{\E}{\mathbb{E}}

\begin{document}

\title{A Unified View of Distribution Shift in Machine Learning:\\
From Classical Foundations to Foundation Models}

\author{Behraj~Khan and Tahir~Qasim~Syed%
\IEEEcompsocitemizethanks{\IEEEcompsocthanksitem B.~Khan and T.~Q.~Syed are with the
Institute of Business Administration (IBA), Karachi, Pakistan.}%
\thanks{Manuscript received XXX; revised XXX.}}

\markboth{IEEE Transactions on Pattern Analysis and Machine Intelligence}%
{Khan and Syed: A Unified View of Distribution Shift in Machine Learning}

\IEEEtitleabstractindextext{%
\begin{abstract}
Machine learning systems are trained under the assumption that training and
deployment data are drawn from the same distribution, an assumption that is
routinely violated in practice. The resulting problem, \emph{distribution
shift}, has generated a large but fragmented literature spanning statistics,
causal inference, computer vision, natural language processing, federated
learning and, most recently, foundation models. Existing surveys typically
address one slice of this landscape---covariate/label shift correction, domain
adaptation and generalization, test-time adaptation, out-of-distribution (OOD)
detection, or the robustness of vision--language models (VLMs)---and rarely
reconcile their terminologies. This survey provides a unified, technically
grounded treatment of the full spectrum. We (i) formalize a single taxonomy in
which covariate, label, conditional, concept, sub-population, feature,
open-set and temporal shift are distinct factorizations of a change in
$P(X,Y)$, and connect it to a causal-graph view with an explicit
identifiability statement; (ii) review classical correction (importance
weighting, label-shift estimation, distributionally robust optimization),
detection (two-sample tests, drift detectors), label-free performance
estimation and conformal guarantees, together with domain-adaptation theory
including its impossibility results; (iii) synthesize recent large-scale
evidence that $Y\mid X$ shift, not covariate shift, dominates real tabular
benchmarks; (iv) organize deep domain adaptation, domain generalization,
sub-population robustness, source-free and test-time adaptation by the target
information available and when; (v) treat federated learning as distribution
shift across clients and across time; (vi) make explicit the mapping between
the shift taxonomy and generalized OOD, anomaly and open-set detection; and
(vii) analyze how foundation models reshape the landscape, from data-driven
effective robustness and its contested origins to robust fine-tuning,
parameter-efficient adaptation and training-free test-time adaptation of VLMs;
and (viii) extend the framework to world models, where shift becomes
endogenous because the agent's own actions generate its test distribution:
we bound model-based decision error by a policy-induced density ratio and a
horizon-amplified model error, show that interaction makes dynamics shift
identifiable on the visited support, and organize the 2024--2026 literature
on physical generalization, test-time adaptation and uncertainty-aware
safety of world models.
We consolidate benchmarks and metrics, compare against 29 existing surveys, and
close with open problems that we believe define the next phase of research on
learning under distributional non-stationarity.
\end{abstract}

}

\maketitle
\IEEEdisplaynontitleabstractindextext
\IEEEpeerreviewmaketitle

\IEEEraisesectionheading{\section{Introduction}\label{sec:introduction}}

\IEEEPARstart{A}{defining} assumption of classical statistical learning
theory is that training and test data are independent and identically
distributed (i.i.d.) draws from a common distribution $P(X,Y)$. This
assumption underlies generalization bounds for empirical risk minimization
(ERM) and is, in practice, routinely false. A model trained on hospital A's
patient population is deployed at hospital B; a perception stack trained on
daytime, clear-weather driving footage is deployed at night, in fog, in a
different city; a recommender trained on last year's users is deployed on
this year's, whose preferences have already moved. In every case the
distribution that generates deployment data, $\Pt(X,Y)$, differs from the one
that generated the training data, $\Ps(X,Y)$. This phenomenon---\emph{distribution
shift}, also called \emph{dataset shift} or \emph{non-stationarity}---is one of
the most consistent causes of the gap between benchmark performance and
real-world reliability~\cite{quinonero2009dataset,moreno2012unifying,koh2021wilds}.

The research response has been large, technically deep and---this is the
central observation motivating this survey---remarkably fragmented. A
statistician correcting covariate shift by importance
weighting~\cite{shimodaira2000improving,huang2007correcting}, a vision
researcher building domain-generalization
benchmarks~\cite{zhou2022domain,wang2023generalizing}, a practitioner running
test-time adaptation~\cite{liang2024comprehensive,wang2021tent}, a safety
researcher building an out-of-distribution (OOD)
detector~\cite{yang2024generalized,hendrycks2017baseline}, an engineer
aggregating models over heterogeneous federated
clients~\cite{kairouz2021advances,zhu2021federated}, and a practitioner
fine-tuning CLIP~\cite{wortsman2022robust,kumar2022finetuning} are, in a
precise technical sense, all responding to the same object: a change in the
data-generating distribution between training and deployment. Yet these
communities use different vocabularies, benchmarks and theoretical tools, and
cite each other only sporadically. A reader entering the field must reconcile
at least half a dozen largely disjoint surveys, and even then the connections
between, e.g., classical label-shift correction and modern test-time prior
adaptation, or between covariate shift and the generalized OOD detection
framework, are rarely made explicit in a single source.

\smallskip
\noindent\textbf{Why now.} Five developments make a unifying treatment
timely.

\smallskip
\noindent\emph{(1) Empirical evidence is overturning methodological priors.}
Large-scale studies of naturally occurring shift in tabular
data~\cite{liu2023need,wang2024rethinking,gardner2023tableshift} find that
shift in $P(Y\mid X)$---not the covariate shift in $P(X)$ that dominates the
classical and deep robustness
literature~\cite{shimodaira2000improving,ganin2016domain}---is the most
prevalent form of performance-degrading shift on their testbeds, and that
distributionally robust objectives are, on average, no better than ERM
there. In parallel, controlled re-evaluations in vision
(DomainBed~\cite{gulrajani2021search}, subpopulation-shift
benchmarks~\cite{yang2023change}) and test-time adaptation~\cite{zhao2023pitfalls,press2023rdumb}
have revised downward the benefit of many specialized methods once model
selection is held to the same standard as the baseline. A survey written
today should surface these tensions rather than reproduce older framings.

\smallskip
\noindent\emph{(2) Foundation models change both the symptoms and the cures.}
Zero-shot vision--language models (VLMs) such as CLIP~\cite{radford2021learning}
sit well above the ``accuracy-on-the-line'' trend of ImageNet-trained
models~\cite{taori2020measuring,miller2021accuracy}, an effect attributed
primarily to training-data distribution~\cite{fang2022data}. At the same time,
standard fine-tuning erodes exactly this
robustness~\cite{kumar2022finetuning,wortsman2022robust,andreassen2021evolution},
and a growing body of work questions how much of it is genuine
generalization rather than coverage of test-like data by web-scale
pretraining~\cite{udandarao2024no,mayilvahanan2025search,shi2023effective}.
This has produced new sub-literatures---robust fine-tuning,
parameter-efficient adaptation, training-free test-time adaptation of
VLMs---that sit awkwardly between classical domain adaptation and phenomena
specific to large pretrained models.

\smallskip
\noindent\emph{(3) The anomaly/OOD/shift literatures have only recently been
unified.} The generalized OOD detection
framework~\cite{yang2024generalized,salehi2022unified} unifies anomaly
detection, novelty detection, open-set recognition, outlier detection and OOD
detection around two primitive notions---\emph{covariate} and \emph{semantic}
shift---that are exactly the constructs studied, under other names, in the
distribution-shift literature proper. VLMs have since blurred these
boundaries again~\cite{miyai2025generalized}, and the unification has not yet
been folded into the broader distribution-shift narrative.

\smallskip
\noindent\emph{(4) Deployment is increasingly distributed and continual.}
Federated learning trains across clients whose data distributions differ by
construction~\cite{kairouz2021advances,zhao2018federated,li2021fedbn}, and
deployed models face streams whose distribution drifts over
time~\cite{gama2014survey,yao2022wildtime}. Both settings are distribution
shift problems, yet they are surveyed separately from the centralized
literature.

\smallskip
\noindent\emph{(5) Models increasingly act on the world they predict.}
Learned world models now underpin planning from pixels and zero-shot robot
control~\cite{hafner2025mastering,assran2025vjepa2,zhou2025dinowm} and are
pretrained at foundation-model scale~\cite{bruce2024genie,nvidia2025cosmos}.
When predictions drive actions, the deployed system generates its own test
distribution, and 2025--2026 evaluations show that such models generalize
well in distribution but poorly out of it~\cite{kang2025far,xue2026acwm,maes2026stable}.
This closed-loop setting has not been connected to the shift taxonomy.

\subsection{Contributions and Scope}
\label{sec:contributions}

This survey attempts a single, technically grounded narrative connecting the
strands above. Concretely, we:
\begin{enumerate}[leftmargin=*,itemsep=1pt]
\item formalize a taxonomy of shift (covariate, label, conditional, concept,
sub-population, feature, open-set, temporal) as factorizations of a change in
$P(X,Y)$, connect each to a structural causal model, and state precisely what
is and is not identifiable from unlabeled target data (\S\ref{sec:taxonomy});
\item review classical correction, detection, label-free performance
estimation and distribution-free uncertainty quantification under shift,
together with domain-adaptation theory including lower bounds and
impossibility results (\S\ref{sec:classical});
\item synthesize the tabular evidence on $Y\mid X$ shift and feature shift
and its methodological consequences (\S\ref{sec:lowdim});
\item organize deep domain adaptation, domain generalization, sub-population
robustness, source-free and test-time adaptation by the target information
available and when (\S\ref{sec:highdim});
\item treat federated learning as distribution shift across clients and
across time (\S\ref{sec:federated});
\item map the shift taxonomy onto generalized OOD, anomaly and open-set
detection, including the VLM era (\S\ref{sec:ood});
\item analyze foundation models: the origins of their effective robustness,
the fine-tuning robustness trade-off, parameter-efficient and test-time
adaptation, and LLM robustness (\S\ref{sec:foundation});
\item extend the framework to world models (\S\ref{sec:worldmodels}): we
show that model-based decision error is controlled by a policy-induced
density ratio times in-distribution model error, amplified by the horizon
(Proposition~\ref{prop:wm}), that interaction makes dynamics (concept) shift
identifiable on the visited support, and we map observation, dynamics,
occupancy and compositional physical shift onto the taxonomy;
\item consolidate benchmarks and metrics (\S\ref{sec:benchmarks}), compare
against 29 existing surveys (\S\ref{sec:gaps}), and propose open problems
(\S\ref{sec:challenges}).
\end{enumerate}

\noindent\textbf{Literature selection.} We assembled the corpus by (i)
backward and forward citation search from the foundational references of each
sub-area (e.g.,~\cite{quinonero2009dataset,bendavid2010theory,ganin2016domain,hendrycks2017baseline,wang2021tent,radford2021learning,mcmahan2017communication}),
(ii) the reference lists of the 29 surveys compared in \S\ref{sec:gaps}, and
(iii) keyword search of the proceedings of the major machine learning,
vision, language and robotics venues and of arXiv through November 2026,
including the proceedings of ICLR, CVPR, ICML and ACL 2026. We prioritized
peer-reviewed work; arXiv-only works are cited when they are widely used
benchmarks, surveys, or the only source of a result we discuss, and are
identified as such in the bibliography. All quantitative claims reported from
individual papers are stated as reported by their authors, with the
experimental conditions under which they hold.

\noindent\textbf{Scope.} We focus on supervised and self-supervised
learning in tabular, vision, vision--language and language data. We touch on,
but do not exhaustively cover, graph data~\cite{zhang2024survey,ma2024survey},
time series~\cite{wu2025tsood} and offline reinforcement
learning~\cite{levine2020offline,prudencio2024survey,samani2026distribution}, each of which has
dedicated surveys; for sequential decision-making we focus on the world-model
component and its shift (\S\ref{sec:worldmodels}) rather than on policy
optimization. Adversarial robustness to worst-case $\ell_p$ perturbations
is out of scope except where it intersects distributional robustness
(\S\ref{sec:dro}).

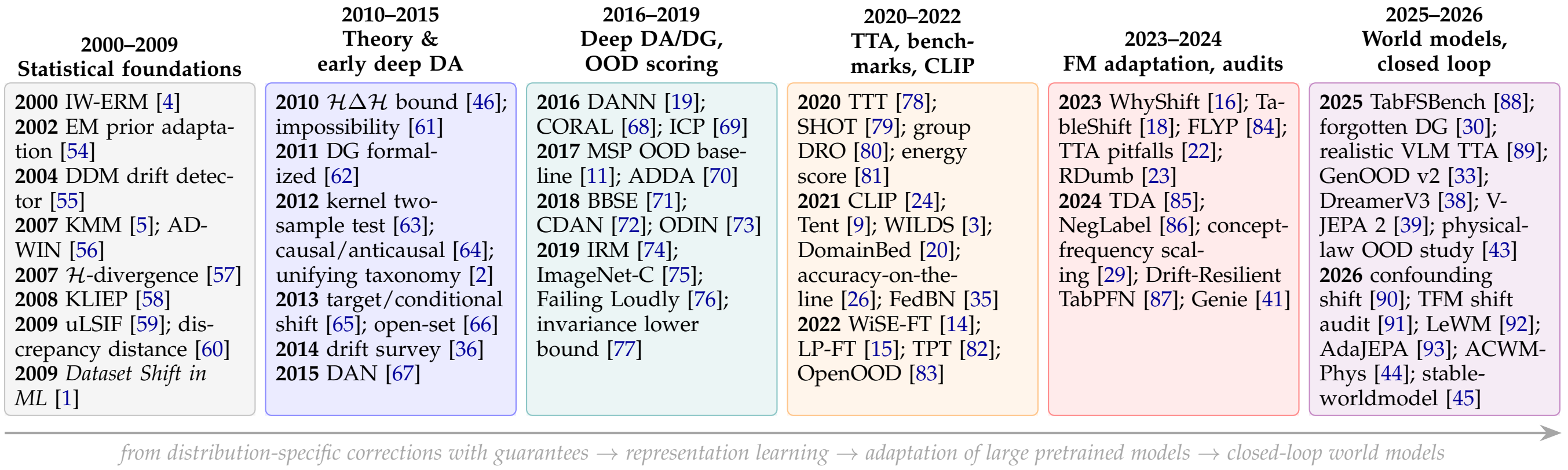
\begin{figure*}[t]
\centering
\begin{tikzpicture}[
  font=\scriptsize,
  era/.style={align=left, anchor=north west, text width=2.66cm, inner sep=2pt},
  head/.style={font=\scriptsize\bfseries, align=center, anchor=south west,
               text width=2.7cm, inner sep=2pt}
]
\def\w{2.98}
\def\h{3.8}
\foreach \i/\col in {0/gray, 1/blue, 2/teal, 3/orange, 4/red, 5/violet}{
  \fill[\col!8] (\i*\w,0) rectangle ++(2.86,-\h);
  \draw[\col!45, rounded corners=2pt] (\i*\w,0) rectangle ++(2.86,-\h);
}
\node[head] at (0*\w,0) {2000--2009\\Statistical foundations};
\node[head] at (1*\w,0) {2010--2015\\Theory \& early deep DA};
\node[head] at (2*\w,0) {2016--2019\\Deep DA/DG, OOD scoring};
\node[head] at (3*\w,0) {2020--2022\\TTA, benchmarks, CLIP};
\node[head] at (4*\w,0) {2023--2024\\FM adaptation, audits};
\node[head] at (5*\w,0) {2025--2026\\World models, closed loop};

\node[era] at (0*\w+0.04,-0.04) {%
\textbf{2000} IW-ERM~\cite{shimodaira2000improving}\\
\textbf{2002} EM prior adaptation~\cite{saerens2002adjusting}\\
\textbf{2004} DDM drift detector~\cite{gama2004learning}\\
\textbf{2007} KMM~\cite{huang2007correcting}; ADWIN~\cite{bifet2007learning}\\
\textbf{2007} $\Hcal$-divergence~\cite{bendavid2007analysis}\\
\textbf{2008} KLIEP~\cite{sugiyama2008direct}\\
\textbf{2009} uLSIF~\cite{kanamori2009ulsif}; discrepancy distance~\cite{mansour2009domain}\\
\textbf{2009} \emph{Dataset Shift in ML}~\cite{quinonero2009dataset}};

\node[era] at (1*\w+0.04,-0.04) {%
\textbf{2010} $\Hcal\Delta\Hcal$ bound~\cite{bendavid2010theory}; impossibility~\cite{bendavid2010impossibility}\\
\textbf{2011} DG formalized~\cite{blanchard2011generalizing}\\
\textbf{2012} kernel two-sample test~\cite{gretton2012kernel}; causal/anticausal~\cite{scholkopf2012causal}; unifying taxonomy~\cite{moreno2012unifying}\\
\textbf{2013} target/conditional shift~\cite{zhang2013domain}; open-set~\cite{scheirer2013open}\\
\textbf{2014} drift survey~\cite{gama2014survey}\\
\textbf{2015} DAN~\cite{long2015learning}};

\node[era] at (2*\w+0.04,-0.04) {%
\textbf{2016} DANN~\cite{ganin2016domain}; CORAL~\cite{sun2016deep}; ICP~\cite{peters2016causal}\\
\textbf{2017} MSP OOD baseline~\cite{hendrycks2017baseline}; ADDA~\cite{tzeng2017adversarial}\\
\textbf{2018} BBSE~\cite{lipton2018detecting}; CDAN~\cite{long2018conditional}; ODIN~\cite{liang2018enhancing}\\
\textbf{2019} IRM~\cite{arjovsky2019invariant}; ImageNet-C~\cite{hendrycks2019benchmarking}; Failing Loudly~\cite{rabanser2019failing}; invariance lower bound~\cite{zhao2019learning}};

\node[era] at (3*\w+0.04,-0.04) {%
\textbf{2020} TTT~\cite{sun2020test}; SHOT~\cite{liang2020we}; group DRO~\cite{sagawa2020distributionally}; energy score~\cite{liu2020energy}\\
\textbf{2021} CLIP~\cite{radford2021learning}; Tent~\cite{wang2021tent}; WILDS~\cite{koh2021wilds}; DomainBed~\cite{gulrajani2021search}; accuracy-on-the-line~\cite{miller2021accuracy}; FedBN~\cite{li2021fedbn}\\
\textbf{2022} WiSE-FT~\cite{wortsman2022robust}; LP-FT~\cite{kumar2022finetuning}; TPT~\cite{shu2022test}; OpenOOD~\cite{yang2022openood}};

\node[era] at (4*\w+0.04,-0.04) {%
\textbf{2023} WhyShift~\cite{liu2023need}; TableShift~\cite{gardner2023tableshift}; FLYP~\cite{goyal2023finetune}; TTA pitfalls~\cite{zhao2023pitfalls}; RDumb~\cite{press2023rdumb}\\
\textbf{2024} TDA~\cite{karmanov2024efficient}; NegLabel~\cite{jiang2024negative}; concept-frequency scaling~\cite{udandarao2024no}; Drift-Resilient TabPFN~\cite{helli2024driftresilient}; Genie~\cite{bruce2024genie}};

\node[era] at (5*\w+0.04,-0.04) {%
\textbf{2025} TabFSBench~\cite{cheng2025tabfsbench}; forgotten DG~\cite{mayilvahanan2025search}; realistic VLM TTA~\cite{zanella2025realistic}; GenOOD v2~\cite{miyai2025generalized}; DreamerV3~\cite{hafner2025mastering}; V-JEPA~2~\cite{assran2025vjepa2}; physical-law OOD study~\cite{kang2025far}\\
\textbf{2026} confounding shift~\cite{reddy2026when}; TFM shift audit~\cite{loza2026empirical}; LeWM~\cite{maes2026leworldmodel}; AdaJEPA~\cite{wang2026adajepa}; ACWM-Phys~\cite{xue2026acwm}; stable-worldmodel~\cite{maes2026stable}};

\draw[-{Stealth[length=2.5mm]}, thick, gray!70] (0,-\h-0.2) -- (6*\w-0.1,-\h-0.2)
  node[midway, below, font=\scriptsize\itshape]{from distribution-specific corrections with guarantees
  $\rightarrow$ representation learning $\rightarrow$ adaptation of large pretrained models
  $\rightarrow$ closed-loop world models};
\end{tikzpicture}
\caption{Timeline of representative milestones in learning under
distribution shift. Each column lists works that introduced a setting,
algorithm, theoretical result or benchmark used throughout this survey.}
\label{fig:timeline}
\end{figure*}

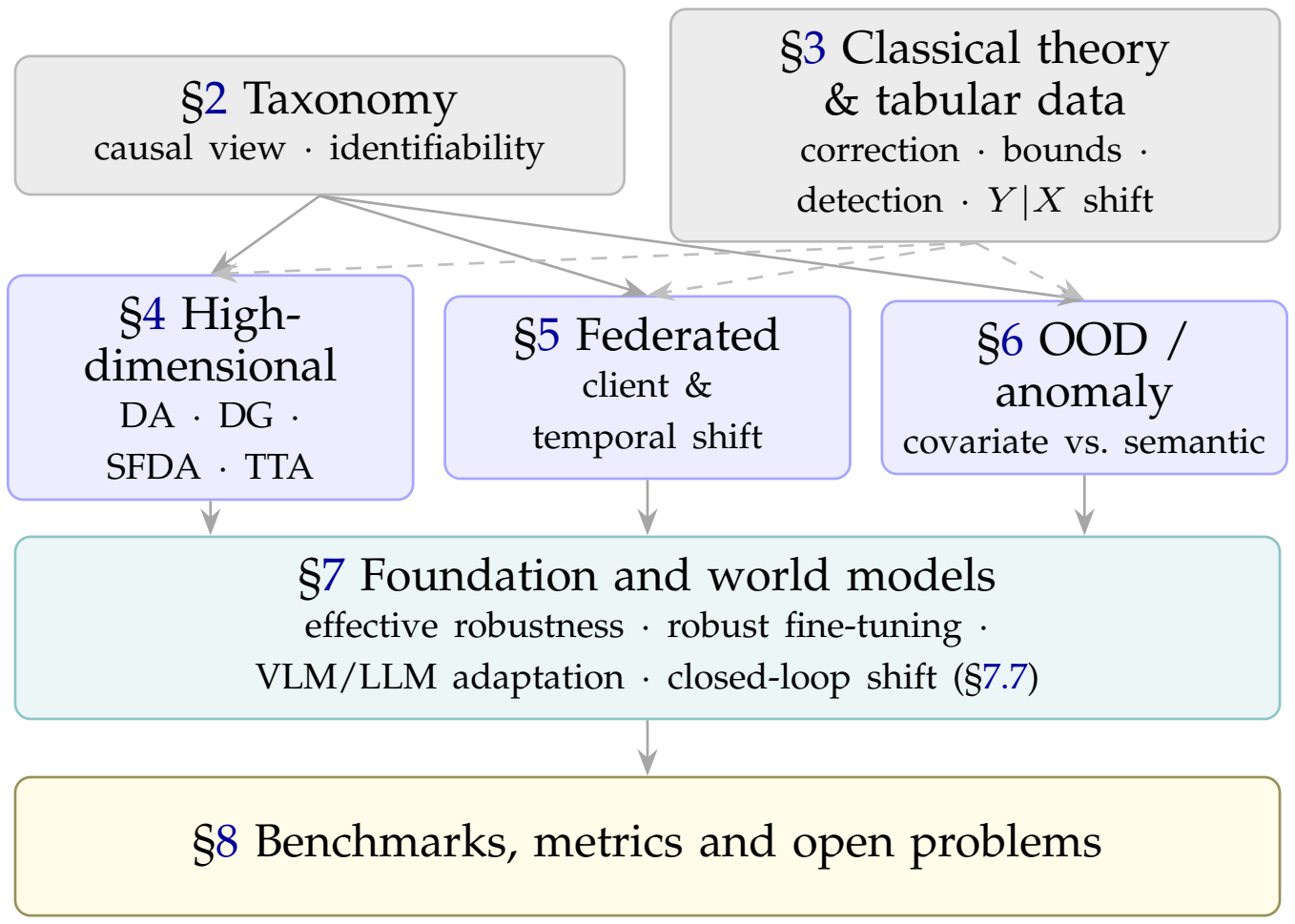
\begin{figure}[t]
\centering
\resizebox{\columnwidth}{!}{%
\begin{tikzpicture}[
  font=\small, >=Stealth, line width=0.5pt,
  base/.style={draw, rounded corners=3pt, align=center, inner sep=4pt,
               minimum height=0.95cm},
  found/.style={base, fill=gray!15, draw=gray!55, text width=3.9cm},
  core/.style={base, fill=blue!7, draw=blue!35, text width=2.5cm},
  wide/.style={base, text width=8.4cm}
]
\node[found] (tax) at (-2.25,0) {\S\ref{sec:taxonomy} Taxonomy\\[-1pt]{\scriptsize causal view $\cdot$ identifiability}};
\node[found] (cls) at (2.25,0) {\S\ref{sec:classical} Classical theory \& tabular data\\[-1pt]{\scriptsize correction $\cdot$ bounds $\cdot$ detection $\cdot$ $Y|X$ shift}};

\node[core] (vis) at (-3.0,-1.8) {\S\ref{sec:highdim} High-dimensional\\[-1pt]{\scriptsize DA $\cdot$ DG $\cdot$ SFDA $\cdot$ TTA}};
\node[core] (fed) at (0,-1.8)   {\S\ref{sec:federated} Federated\\[-1pt]{\scriptsize client \& temporal shift}};
\node[core] (ood) at (3.0,-1.8) {\S\ref{sec:ood} OOD / anomaly\\[-1pt]{\scriptsize covariate vs.\ semantic}};

\node[wide, fill=teal!7, draw=teal!45] (fm) at (0,-3.45) {\S\ref{sec:foundation} Foundation and world models\\[-1pt]{\scriptsize effective robustness $\cdot$ robust fine-tuning $\cdot$ VLM/LLM adaptation $\cdot$ closed-loop shift (\S\ref{sec:worldmodels})}};
\node[wide, fill=yellow!10, draw=yellow!50!black] (bm) at (0,-4.95) {\S\ref{sec:benchmarks} Benchmarks, metrics and open problems};

\foreach \n in {vis,fed,ood}{\draw[->, gray!70] (tax.south) -- (\n.north);}
\foreach \n in {vis,fed,ood}{\draw[->, gray!50, dashed] (cls.south) -- (\n.north);}
\foreach \n in {vis,fed,ood}{\draw[->, gray!70] (\n.south) -- (fm.north -| \n.south);}
\draw[->, gray!70] (fm) -- (bm);
\end{tikzpicture}}
\caption{Organization of the survey. Solid arrows: a section builds directly
on concepts introduced in its predecessor; dashed arrows: the classical
theory of \S\ref{sec:classical} underlies the design objectives of the
methods in the subsequent sections. The comparison with 29 existing surveys
is in \S\ref{sec:gaps}.}
\label{fig:roadmap}
\end{figure}

Fig.~\ref{fig:timeline} places the main developments on a timeline, and
Fig.~\ref{fig:roadmap} summarizes the organization: \S\ref{sec:gaps}
positions the survey against existing ones; \S\ref{sec:taxonomy} and
\S\ref{sec:classical} give the taxonomy, classical theory and the tabular
evidence; \S\ref{sec:highdim}--\S\ref{sec:ood} cover high-dimensional
adaptation, federated learning and OOD detection;
\S\ref{sec:foundation} treats foundation models and world models; and
\S\ref{sec:benchmarks} consolidates evaluation and open problems.

\subsection{Existing Surveys and Gap Analysis}
\label{sec:gaps}

Table~\ref{tab:gaps} compares the scope of this survey with 29 existing
surveys that address overlapping parts of the landscape. Coverage is assessed
from each survey's stated scope (abstract, taxonomy and table of contents);
benchmark-and-findings papers such as WhyShift and TableShift are discussed
in \S\ref{sec:lowdim} rather than listed here.

\begin{table*}[t]
\centering
\footnotesize
\setlength{\tabcolsep}{3.2pt}
\renewcommand{\arraystretch}{1.05}
\caption{Scope comparison with existing surveys. \cmark\ substantial coverage;
\halfmark\ partial or tangential coverage; \xmark\ not covered.
``Classical'' = importance weighting, label-shift estimation, drift detection
and domain-adaptation theory (\S\ref{sec:classical}); ``Tabular'' = the
natural-shift tabular evidence of \S\ref{sec:lowdim}; ``FL'' = federated
shift (\S\ref{sec:federated}); ``FM/VLM'' = foundation-model-specific
robustness and adaptation (\S\ref{sec:foundation}); ``WM'' = world models
and closed-loop shift (\S\ref{sec:worldmodels}); ``Unifies'' = an explicit
mapping across all of these axes and OOD detection.}
\label{tab:gaps}
\begin{tabular}{P{4.3cm}ccccccccc}
\toprule
\textbf{Survey (year)} & \textbf{Classical} & \textbf{Tabular} & \textbf{DA / DG} & \textbf{TTA / SFDA} & \textbf{FL} & \textbf{OOD / AD} & \textbf{FM / VLM} & \textbf{WM} & \textbf{Unifies}\\
\midrule
\multicolumn{10}{l}{\emph{Dataset shift and transfer learning}}\\
Qui\~nonero-Candela et al.~\cite{quinonero2009dataset} (2009) & \cmark & \xmark & \xmark & \xmark & \xmark & \xmark & \xmark & \xmark & \xmark\\
Pan \& Yang~\cite{pan2010survey} (2010) & \halfmark & \xmark & \halfmark & \xmark & \xmark & \xmark & \xmark & \xmark & \xmark\\
Moreno-Torres et al.~\cite{moreno2012unifying} (2012) & \cmark & \xmark & \xmark & \xmark & \xmark & \xmark & \xmark & \xmark & \xmark\\
Wu \& He~\cite{wu2025trustworthy} (2025) & \cmark & \xmark & \halfmark & \xmark & \halfmark & \xmark & \xmark & \xmark & \xmark\\
\multicolumn{10}{l}{\emph{Concept drift}}\\
Gama et al.~\cite{gama2014survey} (2014) & \halfmark & \xmark & \xmark & \xmark & \xmark & \xmark & \xmark & \xmark & \xmark\\
Lu et al.~\cite{lu2018learning} (2019) & \halfmark & \xmark & \xmark & \xmark & \xmark & \xmark & \xmark & \xmark & \xmark\\
\multicolumn{10}{l}{\emph{Domain adaptation and generalization}}\\
Redko et al.~\cite{redko2020survey} (2020) & \cmark & \xmark & \halfmark & \xmark & \xmark & \xmark & \xmark & \xmark & \xmark\\
Wilson \& Cook~\cite{wilson2020survey} (2020) & \xmark & \xmark & \cmark & \xmark & \xmark & \xmark & \xmark & \xmark & \xmark\\
Kouw \& Loog~\cite{kouw2021review} (2021) & \cmark & \xmark & \halfmark & \xmark & \xmark & \xmark & \xmark & \xmark & \xmark\\
Zhou et al.~\cite{zhou2022domain} (2023) & \xmark & \xmark & \cmark & \halfmark & \xmark & \xmark & \xmark & \xmark & \xmark\\
Wang et al.~\cite{wang2023generalizing} (2023) & \halfmark & \xmark & \cmark & \xmark & \xmark & \xmark & \xmark & \xmark & \xmark\\
\multicolumn{10}{l}{\emph{Test-time and source-free adaptation}}\\
Fang et al.~\cite{fang2024sourcefree} (2024) & \xmark & \xmark & \halfmark & \cmark & \xmark & \xmark & \xmark & \xmark & \xmark\\
Li et al.~\cite{li2024comprehensive} (2024) & \xmark & \xmark & \halfmark & \cmark & \halfmark & \xmark & \xmark & \xmark & \xmark\\
Liang et al.~\cite{liang2024comprehensive} (2024) & \xmark & \xmark & \halfmark & \cmark & \xmark & \xmark & \halfmark & \xmark & \xmark\\
Xiao \& Snoek~\cite{xiao2024beyond} (2024) & \xmark & \xmark & \halfmark & \cmark & \xmark & \xmark & \halfmark & \xmark & \xmark\\
Wang et al.~\cite{wang2025search} (2025) & \xmark & \xmark & \xmark & \cmark & \xmark & \xmark & \xmark & \xmark & \xmark\\
Maharana et al.~\cite{maharana2026continual} (2026, CTTA) & \xmark & \xmark & \xmark & \cmark & \xmark & \xmark & \xmark & \xmark & \xmark\\
\multicolumn{10}{l}{\emph{OOD, anomaly and open-set detection}}\\
Salehi et al.~\cite{salehi2022unified} (2022) & \xmark & \xmark & \xmark & \xmark & \xmark & \cmark & \xmark & \xmark & \xmark\\
Yang et al.~\cite{yang2024generalized} (2024) & \xmark & \xmark & \xmark & \xmark & \xmark & \cmark & \halfmark & \xmark & \xmark\\
Miyai et al.~\cite{miyai2025generalized} (2025) & \xmark & \xmark & \xmark & \xmark & \xmark & \cmark & \cmark & \xmark & \xmark\\
Xu \& Ding~\cite{xu2025large} (2025) & \xmark & \xmark & \xmark & \xmark & \xmark & \cmark & \cmark & \xmark & \xmark\\
\multicolumn{10}{l}{\emph{Federated learning}}\\
Kairouz et al.~\cite{kairouz2021advances} (2021) & \xmark & \xmark & \xmark & \xmark & \cmark & \xmark & \xmark & \xmark & \xmark\\
Zhu et al.~\cite{zhu2021federated} (2021) & \xmark & \xmark & \xmark & \xmark & \cmark & \xmark & \xmark & \xmark & \xmark\\
\multicolumn{10}{l}{\emph{Language models, other modalities and sequential decision-making}}\\
Zhang et al.~\cite{zhang2025evaluating} (2025, LLMs) & \xmark & \xmark & \xmark & \xmark & \xmark & \halfmark & \cmark & \xmark & \xmark\\
Zhang et al.~\cite{zhang2024survey} (2024, graphs) & \xmark & \xmark & \halfmark & \halfmark & \xmark & \xmark & \xmark & \xmark & \xmark\\
Samani et al.~\cite{samani2026distribution} (2026, offline RL) & \halfmark & \xmark & \xmark & \xmark & \xmark & \halfmark & \xmark & \halfmark & \xmark\\
Li et al.~\cite{li2025comprehensivewm} (2025, world models) & \xmark & \xmark & \xmark & \xmark & \xmark & \xmark & \halfmark & \cmark & \xmark\\
Wang et al.~\cite{wang2026towards_safe} (2026, safe world models) & \xmark & \xmark & \xmark & \xmark & \xmark & \halfmark & \xmark & \cmark & \xmark\\
Wu et al.~\cite{wu2025tsood} (2025, time series) & \xmark & \xmark & \halfmark & \xmark & \xmark & \xmark & \halfmark & \xmark & \xmark\\
\midrule
\textbf{This survey} & \cmark & \cmark & \cmark & \cmark & \cmark & \cmark & \cmark &  \cmark & \cmark\\
\bottomrule
\end{tabular}
\end{table*}

\subsubsection{What Existing Surveys Cover}
\label{sec:survey-by-survey}

\noindent\textbf{Dataset shift and transfer learning.} The edited volume of
Qui\~nonero-Candela et al.~\cite{quinonero2009dataset} and the taxonomy of
Moreno-Torres et al.~\cite{moreno2012unifying} established the vocabulary of
\S\ref{sec:taxonomy}; both predate representation learning. Pan and
Yang~\cite{pan2010survey} treat domain adaptation as one transfer setting
among several, and Wu and He~\cite{wu2025trustworthy} survey transferability
theory together with trustworthiness (robustness, fairness, privacy) of
transfer.

\smallskip
\noindent\textbf{Concept drift.} Gama et al.~\cite{gama2014survey} and Lu et
al.~\cite{lu2018learning} are the standard references for streaming drift
detection and adaptation; they do not address representation-level
adaptation or vision benchmarks.

\smallskip
\noindent\textbf{Domain adaptation and generalization.} Redko et
al.~\cite{redko2020survey} survey DA theory; Kouw and Loog~\cite{kouw2021review}
review DA without target labels from a statistical perspective (sample-,
feature- and inference-based); Wilson and Cook~\cite{wilson2020survey} survey
deep UDA; and Zhou et al.~\cite{zhou2022domain} and Wang et
al.~\cite{wang2023generalizing} are the principal DG surveys. None connects
DA/DG to OOD detection, federated shift or the tabular evidence.

\smallskip
\noindent\textbf{Test-time and source-free adaptation.} Liang et
al.~\cite{liang2024comprehensive} define the TTA sub-settings we adopt;
Wang et al.~\cite{wang2025search} benchmark online TTA with ViTs; Xiao and
Snoek~\cite{xiao2024beyond} organize TTA by the adapted component; and Li et
al.~\cite{li2024comprehensive} and Fang et al.~\cite{fang2024sourcefree} survey
SFDA. Maharana et al.~\cite{maharana2026continual} survey continual TTA and
restrict themselves explicitly to covariate shift. These treat classical
correction and OOD detection as context rather than as part of a common
framework.

\smallskip
\noindent\textbf{OOD and anomaly detection.} Yang et
al.~\cite{yang2024generalized} and Salehi et al.~\cite{salehi2022unified}
provide the unified detection frameworks of \S\ref{sec:ood}; Miyai et
al.~\cite{miyai2025generalized} revise them for VLMs; and Xu and
Ding~\cite{xu2025large} survey LLM-based detection. Their connection to the
dataset-shift literature is primarily terminological.

\smallskip
\noindent\textbf{Federated learning.} Kairouz et al.~\cite{kairouz2021advances}
and Zhu et al.~\cite{zhu2021federated} discuss non-IID data extensively but
not its correspondence with the centralized shift taxonomy, test-time shift,
or the evaluation lessons of DG.

\smallskip
\noindent\textbf{Adjacent modalities and LLMs.} Surveys on graphs~\cite{zhang2024survey,ma2024survey},
time series~\cite{wu2025tsood} and LLM robustness~\cite{zhang2025evaluating}
address structurally distinct settings with analogous machinery; we treat
them as adjacent rather than subsumed.

\smallskip
\noindent\textbf{World models and offline RL.} Li et
al.~\cite{li2025comprehensivewm} organize world models for embodied AI by
functionality, temporal modeling and spatial representation, and list
physical consistency of evaluation among open challenges, but do not relate
world-model failures to a shift taxonomy. Wang et
al.~\cite{wang2026towards_safe} organize world-model safety along a lifecycle
from data and observations through dynamics, planning and feedback, covering
non-adversarial failures and attacks; their perspective is safety rather
than the statistical structure of the shift. Samani et
al.~\cite{samani2026distribution} survey distribution shift and OOD actions in
offline RL, where model-based pessimism is one family among many. Neither
connects the closed-loop shift of world models to the static taxonomy,
identifiability results or TTA literature of the preceding sections.

\subsubsection{Identified Gaps}
\label{sec:identified-gaps}

\begin{enumerate}[leftmargin=*,itemsep=1pt]
\item \textbf{No single source connects classical correction theory and its
negative results to deep, federated and foundation-model methods.} Why
alignment losses take their form (Theorem~\ref{thm:bendavid}) and when they
provably fail~\cite{zhao2019learning,johansson2019support} are rarely presented
alongside the methods they govern.
\item \textbf{The $Y\mid X$-shift evidence~\cite{liu2023need,wang2024rethinking}
has not been integrated into DA/DG surveys}, despite its implications for
which methods transfer to tabular deployment.
\item \textbf{Streaming drift detection and online TTA remain disconnected}
(\S\ref{sec:tta}), with no shared benchmark spanning data-mining and vision
venues.
\item \textbf{OOD detection and the shift taxonomy are adjacent but not
merged}, obscuring why a score's effectiveness depends on whether shift is
covariate or semantic and how detection should interact with adaptation
(\S\ref{sec:ood-connection}).
\item \textbf{Federated heterogeneity is not mapped onto the shift
taxonomy}, so FL methods are rarely compared with their centralized
counterparts (e.g., agnostic FL versus group DRO, FedBN versus BN-based TTA).
\item \textbf{The origin of foundation-model robustness is contested}
(\S\ref{sec:pretraining-robustness}), and existing surveys do not confront the
evidence on coverage, contamination and measurement artifacts with the
classical notions of support overlap and diversity shift.
\item \textbf{Feature shift in tabular data lacks a methodological
literature} beyond the benchmark establishing its relevance~\cite{cheng2025tabfsbench}.
\item \textbf{World models are absent from distribution-shift surveys}, and
world-model surveys do not use the shift vocabulary, although observation,
dynamics and policy-induced shift are instances of conditional, concept and
covariate shift and the 2026 world-model TTA methods re-derive ideas from
\S\ref{sec:tta} (\S\ref{sec:worldmodels}).
\end{enumerate}

\begin{remark}[What a survey cannot resolve]
Gaps (3), (5) and (7) are primarily empirical and infrastructural---the
absence of shared benchmarks or mature method families---rather than gaps a
synthesis alone can close. We return to them as open problems in
\S\ref{sec:challenges}.
\end{remark}
\section{Foundations and Taxonomy of Distribution Shift}
\label{sec:taxonomy}

We fix notation and give a single formal taxonomy into which the
terminology of the remainder of the survey is mapped. Readers familiar with
the classical taxonomies of Storkey~\cite{storkey2009training} and
Moreno-Torres et al.~\cite{moreno2012unifying} will recognize much of
\S\ref{sec:basic-types}; our contribution is to connect these types to (i) a
causal-graph view (\S\ref{sec:causal-view}), (ii) the sub-population, feature,
open-set and temporal notions prominent in modern deep and tabular learning
(\S\ref{sec:modern-types}), and (iii) an explicit statement of what is
identifiable from the data typically available at adaptation time
(\S\ref{sec:identifiability}). Table~\ref{tab:shift-types} summarizes the
result.

\subsection{Formal Setup}
\label{sec:setup}

Let $X\in\X$ denote inputs and $Y\in\Y$ the target. A learning problem is
specified by a joint distribution $P(X,Y)$, which factorizes in two
equivalent ways:
\begin{equation}
P(X,Y) = \underbrace{P(X)}_{\text{marginal}}\,\underbrace{P(Y\mid X)}_{\text{posterior}}
= \underbrace{P(Y)}_{\text{prior}}\,\underbrace{P(X\mid Y)}_{\text{class-conditional}}.
\label{eq:factorization}
\end{equation}
A model $h:\X\to\Y$ is trained on samples from a \emph{source} $\Ps(X,Y)$ and
deployed on a \emph{target} $\Pt(X,Y)$. For a loss $\ell$, the source and
target risks are $\epsilon_S(h)=\E_{\Ps}[\ell(h(X),Y)]$ and
$\epsilon_T(h)=\E_{\Pt}[\ell(h(X),Y)]$. \emph{Distribution shift} is the
statement $\Ps\neq\Pt$. Because $P(X,Y)$ admits both factorizations in
\eqref{eq:factorization}, a difference can be localized to one or more of
$P(X)$, $P(Y\mid X)$, $P(Y)$ and $P(X\mid Y)$; the taxonomy below enumerates
which components are allowed to change.

\begin{definition}[Basic shift types]
\label{def:shift}
Assume $\Ps\neq\Pt$. The shift is
\textbf{covariate shift} if $\Ps(X)\neq\Pt(X)$ and $\Ps(Y\mid X)=\Pt(Y\mid X)$;
\textbf{label (prior) shift} if $\Ps(Y)\neq\Pt(Y)$ and $\Ps(X\mid Y)=\Pt(X\mid Y)$;
\textbf{conditional shift} if $\Ps(X\mid Y)\neq\Pt(X\mid Y)$ and $\Ps(Y)=\Pt(Y)$;
\textbf{concept shift} if $\Ps(Y\mid X)\neq\Pt(Y\mid X)$, with or without a
change in $P(X)$; and \textbf{compound} (or general) shift if several
components change with no constraint relating $\Ps$ and $\Pt$.
\end{definition}

Covariate and label shift come from the early importance-weighting and
prior-adaptation literatures~\cite{shimodaira2000improving,saerens2002adjusting};
conditional and the combined ``target and conditional'' shift were
formalized by Zhang et al.~\cite{zhang2013domain}. A large fraction of the
methodological literature is organized around exactly one of these cases,
most often the first; as we discuss in \S\ref{sec:lowdim}, this
organization is in tension with what is observed in deployed systems.

\subsection{Basic Shift Types}
\label{sec:basic-types}

\begin{description}[leftmargin=0pt,style=nextline,itemsep=2pt]
\item[Covariate shift.] The input distribution moves but the labeling
function does not. This is the setting of importance-weighted
ERM~\cite{shimodaira2000improving,huang2007correcting,sugiyama2008direct} and
is implicitly assumed by most deep feature-alignment
methods~\cite{ganin2016domain,sun2016deep,long2015learning}, which seek a
representation under which $P(X)$ is aligned while trusting that $P(Y\mid X)$
is preserved by the alignment.

\item[Label shift.] The class-conditional generative process is unchanged
but class frequencies move---e.g., disease prevalence differs between the
hospital where a diagnostic model was trained and the one where it is
deployed. This is the natural assumption under an \emph{anticausal}
generative story $Y\to X$~\cite{scholkopf2012causal} and licenses correction
by prior re-estimation~\cite{saerens2002adjusting,lipton2018detecting}
(\S\ref{sec:bbse}).

\item[Conditional shift and generalized label shift.] The appearance of each
class changes (a new sensor, a new rendering style) while class frequencies
may or may not. In representation space, Tachet des Combes et
al.~\cite{tachet2020domain} define \emph{generalized label shift} (GLS):
there exists a representation $Z=g(X)$ such that $\Ps(Z\mid Y)=\Pt(Z\mid Y)$,
while $P(Y)$ may change. GLS is the assumption under which class-conditional
feature alignment combined with label-proportion re-weighting is
provably sound; classical label shift is the special case $Z=X$.

\item[Concept shift / drift.] The input--label relationship itself changes.
No reweighting of source data can correct it without additional structure
(e.g., an invariance assumption across environments, \S\ref{sec:causal-view}),
because the source posterior is the wrong target. In streaming settings this
is called \emph{concept drift}~\cite{widmer1996learning,gama2014survey,lu2018learning}
(\S\ref{sec:driftdetection}).

\item[Domain shift.] An umbrella term, mostly used in vision and NLP, for a
typically compound shift between a \emph{source domain} and one or more
\emph{target domains} arising from a change in the data-generating mechanism
(sensor, geography, style, modality). It is the subject of domain adaptation
and generalization (\S\ref{sec:dadtheory}, \S\ref{sec:highdim}).
\end{description}

\begin{table*}[t]
\centering
\footnotesize
\setlength{\tabcolsep}{3.5pt}
\renewcommand{\arraystretch}{1.12}
\caption{Taxonomy of distribution shift. ``Invariant'' is the component
assumed shared by source and target. ``Identifiable'' states whether the
target risk can be estimated from labeled source data plus unlabeled target
data under the stated assumption (\S\ref{sec:identifiability}).}
\label{tab:shift-types}
\begin{tabular}{P{2.15cm}P{3.1cm}P{2.0cm}P{2.85cm}P{3.15cm}P{3.0cm}}
\toprule
\textbf{Shift type} & \textbf{What changes} & \textbf{Invariant} & \textbf{Identifiable?} & \textbf{Typical remedy} & \textbf{Representative work}\\
\midrule
Covariate & $P(X)$ & $P(Y\mid X)$ & Yes, if $\mathrm{supp}\,\Pt(X)\subseteq\mathrm{supp}\,\Ps(X)$ & Importance weighting; feature alignment & \cite{shimodaira2000improving,huang2007correcting,sugiyama2008direct,ganin2016domain}\\
Label / prior & $P(Y)$ & $P(X\mid Y)$ & Yes, if confusion matrix invertible & Prior re-estimation, logit adjustment & \cite{saerens2002adjusting,lipton2018detecting,azizzadenesheli2019regularized,garg2020unified}\\
Conditional; GLS & $P(X\mid Y)$ (and $P(Y)$) & $P(Z\mid Y)$ for some $Z=g(X)$ & Only under structural assumptions & Class-conditional alignment + reweighting & \cite{zhang2013domain,tachet2020domain}\\
Concept & $P(Y\mid X)$ & none (w/o assumptions) & No from unlabeled target & Target labels, invariance, retraining & \cite{widmer1996learning,peters2016causal,arjovsky2019invariant,liu2023need}\\
Sub-population & group weights $\pi(g)$ & $P(X,Y\mid g)$ & Worst case boundable & Group DRO, reweighting, last-layer retraining & \cite{sagawa2020distributionally,liu2021just,kirichenko2023last,yang2023change}\\
Feature / schema & feature set $\X$ & partial & Not by reweighting & Imputation, schema-aware models & \cite{cheng2025tabfsbench}\\
Open-set / semantic & $\mathrm{supp}\,P(Y)$ grows & $P(X,Y)$ on known $\Y$ & Detection only & OOD detection, rejection & \cite{scheirer2013open,hendrycks2017baseline,yang2024generalized}\\
Temporal / drift & any of the above over $t$ & varies & Sequentially, with delay & Drift detection, online / test-time adaptation & \cite{gama2014survey,bifet2007learning,wang2022continual,yao2022wildtime}\\
\bottomrule
\end{tabular}
\end{table*}

\subsection{A Causal-Graph View}
\label{sec:causal-view}

\begin{figure}[t]
\centering
\begin{tikzpicture}[
  var/.style={draw, circle, minimum size=6.5mm, font=\small, inner sep=0pt},
  env/.style={draw, rectangle, rounded corners, minimum size=5mm, font=\scriptsize, fill=orange!15, inner sep=1.5pt},
  >=Stealth, thick
]
\begin{scope}
  \node[var] (Xa) at (0,0) {$X$};
  \node[var] (Ya) at (1.8,0) {$Y$};
  \node[env] (Ea) at (0,1.05) {$E$};
  \node[env] (Fa) at (0.9,1.05) {$E'$};
  \draw[->] (Xa) -- node[above, font=\scriptsize]{$P(Y|X)$} (Ya);
  \draw[->, dashed] (Ea) -- (Xa);
  \draw[->, dashed, red!70!black] (Fa) -- (0.9,0.08);
  \node[font=\scriptsize, align=center] at (0.9,-0.75) {(a) causal: $X\to Y$};
\end{scope}
\begin{scope}[xshift=4.0cm]
  \node[var] (Yb) at (0,0) {$Y$};
  \node[var] (Xb) at (1.8,0) {$X$};
  \node[env] (Eb) at (0,1.05) {$E$};
  \node[env] (Fb) at (0.9,1.05) {$E'$};
  \draw[->] (Yb) -- node[above, font=\scriptsize]{$P(X|Y)$} (Xb);
  \draw[->, dashed] (Eb) -- (Yb);
  \draw[->, dashed, red!70!black] (Fb) -- (0.9,0.08);
  \node[font=\scriptsize, align=center] at (0.9,-0.75) {(b) anticausal: $Y\to X$};
\end{scope}
\end{tikzpicture}
\caption{Causal-graph view. (a) If $X$ causes $Y$, an environment $E$ acting
only on $P(X)$ produces covariate shift with $P(Y\mid X)$ invariant; an
environment $E'$ acting on the mechanism (red) produces concept shift.
(b) If $Y$ causes $X$, $E$ acting on $P(Y)$ produces label shift with
$P(X\mid Y)$ invariant; $E'$ acting on the mechanism produces conditional
shift. Dashed edges denote interventions that differ between source and
target.}
\label{fig:causal-graphs}
\end{figure}
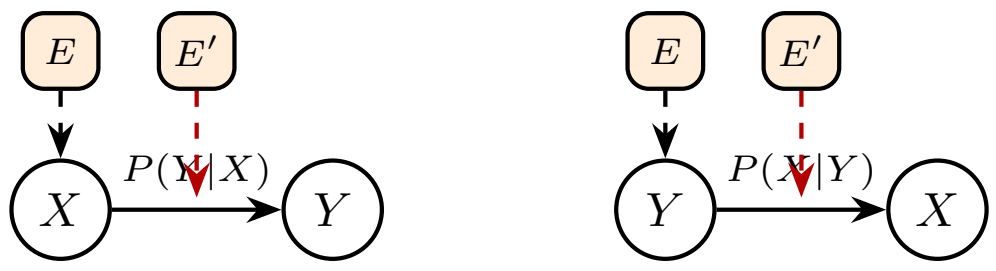

The classical taxonomy becomes considerably more transparent when each shift
type is read off a structural causal model (SCM) over $X$, $Y$ and an
environment index $E$ (Fig.~\ref{fig:causal-graphs}).

This view, developed in the causal-learning
literature~\cite{scholkopf2012causal,peters2016causal,scholkopf2021toward},
clarifies a point that is easy to lose in the purely distributional
taxonomy: \emph{covariate and label shift are not alternatives chosen by
problem type; they are consequences of the assumed causal direction combined
with which node the environment acts on.} If $X$ causes $Y$ and the
environment acts only on $X$'s generation (a different camera, a different
acquisition protocol), the mechanism $P(Y\mid X)$ is invariant by causal
modularity---an assumption with real content, not a convenience. If $Y$
causes $X$ (a disease generating symptoms) and the environment acts on
prevalence, it is $P(X\mid Y)$ that is stable. In the anticausal case a change in $P(Y)$ changes $P(Y\mid X)$ as well, so
covariate-shift correction applied there violates its own central
assumption~\cite{scholkopf2012causal}; this mismatch is a principal
motivation for the invariant-prediction literature
(\S\ref{sec:dg})~\cite{peters2016causal,arjovsky2019invariant}.

\begin{remark}
Concept shift corresponds to an intervention on the mechanism itself. No
reweighting of $\Ps$ recovers $\Pt$ unless the change is structured---e.g.,
a sparse intervention affecting only non-causal parents of $Y$, which is
precisely the condition exploited by invariant causal
prediction~\cite{peters2016causal}. More generally, different
domain-generalization objectives impose different conditional-independence
constraints, and Kaur et al.~\cite{kaur2023modeling} prove that any algorithm
based on a single fixed constraint cannot succeed across all shift types
realizable by a canonical causal graph.
\end{remark}

\subsection{Modern Extensions}
\label{sec:modern-types}

The classical taxonomy predates both distributionally robust optimization
and the deep learning era, and several shift types prominent in recent work
do not map cleanly onto it.

\smallskip
\noindent\textbf{Sub-population shift.} The target is a reweighting of
groups already present (possibly with vanishing weight) in the source:
$\Pt(X,Y)=\sum_g\pi_T(g)P(X,Y\mid g)$ with $\pi_S\neq\pi_T$. This is the
setting of group DRO~\cite{sagawa2020distributionally} and fairness-motivated
robust losses~\cite{hashimoto2018fairness}. It reduces to label shift when
groups are defined by $Y$, but is genuinely distinct when groups are defined
by an attribute $A$: then neither $P(X)$ nor $P(Y)$ alone localizes the shift.
Yang et al.~\cite{yang2023change} further decompose it into \emph{spurious
correlations} (an attribute correlated with $Y$ in training but not at test),
\emph{attribute imbalance}, \emph{class imbalance} and \emph{attribute
generalization} (attribute values unseen in training), and show that
algorithms improve some of these but not others (\S\ref{sec:subpop}).

\smallskip
\noindent\textbf{Diversity versus correlation shift.} Ye et
al.~\cite{ye2022ood} propose an orthogonal, measurable decomposition of
domain shift into \emph{diversity shift} (features whose support differs
between domains) and \emph{correlation shift} (features whose correlation with
$Y$ differs). They find that methods that beat ERM on one dimension
typically fail to on the other, a useful empirical counterpart to the
causal argument above.

\smallskip
\noindent\textbf{Feature shift.} The set or schema of observed features
changes: a feature available at training is missing, corrupted or replaced
at deployment. This changes $\X$ itself rather than the distribution over a
fixed $\X$, and has only recently received a dedicated
benchmark~\cite{cheng2025tabfsbench} (\S\ref{sec:feature-shift-tabular}).

\smallskip
\noindent\textbf{Open-set / semantic shift.} The target places mass on
labels $y\notin\mathrm{supp}\,\Ps(Y)$. No reweighting of a fixed label space can
produce mass on an unobserved label; the shift underlies open-set recognition
and OOD detection (\S\ref{sec:ood}). Label-set mismatch also defines the
partial, open-set and universal domain-adaptation settings of
\S\ref{sec:dda}.

\smallskip
\noindent\textbf{Temporal shift.} Any of the above can be \emph{static} or
evolve over deployment time. The dynamic case is classified by rate and
structure into \emph{sudden}, \emph{gradual}, \emph{incremental} and
\emph{recurring} drift~\cite{gama2014survey,ditzler2015learning}, and is the
setting of drift detection (\S\ref{sec:driftdetection}), continual test-time
adaptation (\S\ref{sec:tta}) and temporal benchmarks such as
Wild-Time~\cite{yao2022wildtime}.

\subsection{What Is Identifiable?}
\label{sec:identifiability}

Most adaptation methods have access to labeled source data and unlabeled
target data, i.e., to $\Ps(X,Y)$ and $\Pt(X)$. The following elementary
statement explains why every method must commit to a structural assumption.

\begin{proposition}[Non-identifiability without structure]
\label{prop:nonid}
Let $\Ps(X,Y)$ and $\Pt(X)$ be given. For every conditional $Q(Y\mid X)$, the
joint $\Pt^{Q}(X,Y)=\Pt(X)Q(Y\mid X)$ is consistent with all observable data.
Hence, for a fixed $h$, $\epsilon_T(h)$ can take any value in
$[\inf_Q \E_{\Pt^Q}\ell,\ \sup_Q\E_{\Pt^Q}\ell]$, which for the 0-1 loss and
non-degenerate $h$ is $[0,1]$.
\end{proposition}

The proof is immediate: the observables constrain only the $X$-marginal of
the target. The proposition is the formal reason why (i) covariate shift
assumes $Q=\Ps(Y\mid X)$ and then requires support overlap for the
importance weights of \S\ref{sec:classical-correction} to exist; (ii) label
shift assumes $\Pt(X\mid Y)=\Ps(X\mid Y)$, under which $\Pt(Y)$ is identified
whenever the confusion matrix of a fixed predictor is
invertible~\cite{lipton2018detecting}; and (iii) concept shift cannot be
corrected from unlabeled target data alone. More generally, Ben-David et
al.~\cite{bendavid2010impossibility} show that natural assumptions taken
individually (covariate shift, or a small divergence between marginals) do
not suffice for successful adaptation. It also explains the role of
\emph{multiple} source environments in domain generalization: they replace a
direct assumption about $\Pt(Y\mid X)$ with an assumption about what is
invariant across environments (\S\ref{sec:dg}).

The invariance assumption can itself fail when the shift originates in an
unobserved variable. Reddy et al.~\cite{reddy2026when} analyze
\emph{confounding shift}, in which a latent confounder $U$ of $X$ and $Y$
changes its distribution across environments, so that every environment has a
different joint and the invariances assumed by many OOD-generalization methods
no longer hold. They argue that ERM can then match or exceed specialized
methods, that generalization requires environment-specific rather than only
invariant relationships, and that adding observed proxies of the confounder
reduces concept shift; experiments on eight TableShift tasks support these
claims. This is the causal counterpart of the tabular evidence of
\S\ref{sec:yx-shift}.

\subsection{Relation to Transfer and Multi-Task Learning}
\label{sec:relation-transfer}

Transfer learning~\cite{pan2010survey,wu2025trustworthy} is the general
problem of using knowledge from a source task or domain to improve learning
on a target. In \emph{domain adaptation} the task ($\Y$ and the optimal
decision rule up to shift) is fixed while the domain differs---exactly our
notion of distribution shift. In \emph{multi-task learning} the task differs
while the domain may be shared, and in few-shot and meta-learning a small
labeled target sample is available, which most shift-handling methods do not
assume. Distribution shift is therefore the sub-problem of transfer learning
in which the predictive task is fixed and only the data-generating
distribution moves---the restriction that makes the fine-grained taxonomy of
Table~\ref{tab:shift-types} possible.
\section{Classical Theory and the Tabular Regime}
\label{sec:classical}

This section reviews the statistical machinery for correcting, bounding and
detecting the shift types of \S\ref{sec:taxonomy}. We organize it into
correction under a specified shift type (\S\ref{sec:classical-correction}),
distributionally robust optimization (\S\ref{sec:dro}), domain-adaptation
theory that bounds target risk without fixing the shift type
(\S\ref{sec:dadtheory}), detection and diagnosis
(\S\ref{sec:twosample}--\S\ref{sec:driftdetection}), and label-free
performance estimation and uncertainty quantification
(\S\ref{sec:perf-est}). It closes with low-dimensional tabular data
(\S\ref{sec:lowdim}), the regime in which these tools are most directly
applicable and in which recent natural-shift benchmarks have tested their
assumptions most sharply. A recurring theme is that correction is only as good
as the shift-type assumption it encodes (Proposition~\ref{prop:nonid}).

\subsection{Correction under a Specified Shift Type}
\label{sec:classical-correction}

\subsubsection{Covariate shift: importance weighting}
Under covariate shift and $\mathrm{supp}\,\Pt(X)\subseteq\mathrm{supp}\,\Ps(X)$,
the target risk is a reweighted source risk with density ratio
$w(x)=\Pt(x)/\Ps(x)$:
\begin{equation}
\epsilon_T(h)=\E_{\Ps}\!\left[w(X)\,\ell(h(X),Y)\right].
\label{eq:iw}
\end{equation}
Minimizing $\frac1n\sum_i\hat w(x_i)\ell(h(x_i),y_i)$ is
\emph{importance-weighted ERM}~\cite{shimodaira2000improving}; model selection
must use the same weights (importance-weighted cross-validation,
IWCV~\cite{sugiyama2007covariate}). The central difficulty is estimating $w$
without estimating two high-dimensional densities. Kernel Mean Matching
(KMM)~\cite{huang2007correcting} matches the reweighted source mean embedding
to the target embedding in an RKHS; KLIEP~\cite{sugiyama2008direct} minimizes
$\mathrm{KL}(\Pt\Vert w\Ps)$ subject to normalization; and
uLSIF~\cite{kanamori2009ulsif} replaces KL by a squared loss, yielding a
closed-form estimator. A practical alternative is to train a probabilistic
domain classifier $c(x)\approx\Pr(\text{target}\mid x)$ and set
$\hat w(x)\propto c(x)/(1-c(x))$.

Two caveats are central. First, the variance of \eqref{eq:iw} is governed by
the second moment of $w$: Cortes et al.~\cite{cortes2010learning} show that
generalization bounds for importance weighting depend on the R\'enyi
divergence $d_2(\Pt\Vert\Ps)=\E_{\Pt}[w]$, and that learning can fail when
weights are unbounded---the statistical signature of what the deep
literature calls a ``large domain gap.'' Second, for over-parameterized
networks that can interpolate the training data, Byrd and
Lipton~\cite{byrd2019effect} find that the effect of importance weights on
the learned classifier diminishes over training unless combined with
regularization, early stopping or other mechanisms that prevent
separability; Fang et al.~\cite{fang2020rethinking} address the resulting
circularity (weights estimated on fixed features that themselves need to be
learned) by alternating weight estimation and representation learning.

\subsubsection{Label shift: prior re-estimation}
\label{sec:bbse}
Under label shift, $\Pt(y\mid x)\propto \frac{\Pt(y)}{\Ps(y)}\Ps(y\mid x)$, so
correction reduces to estimating the weight vector
$\boldsymbol{w}_y=\Pt(y)/\Ps(y)$. The EM procedure of Saerens et
al.~\cite{saerens2002adjusting} (maximum-likelihood label shift, MLLS)
alternates between re-estimating $\Pt(Y)$ and recalibrating posteriors.
Black Box Shift Estimation (BBSE)~\cite{lipton2018detecting} instead exploits
the invariance of the confusion matrix $C_{ij}=\Ps(\hat Y=i,Y=j)$ of any fixed
predictor: since $\Pt(\hat Y=i)=\sum_j C_{ij}\boldsymbol{w}_j$, the weights
solve the linear system $\hat{\boldsymbol\mu}_T=\hat C\boldsymbol w$ and are
consistent whenever $C$ is invertible. RLLS~\cite{azizzadenesheli2019regularized}
regularizes this system for small target samples. Garg et
al.~\cite{garg2020unified} show that BBSE and MLLS are instances of a common
distribution-matching framework and that MLLS's advantage stems from using
calibrated soft predictions rather than hard confusion matrices, consistent
with the empirical finding of Alexandari et al.~\cite{alexandari2020maximum}
that MLLS with bias-corrected calibration is hard to beat. All of these
estimators degrade---potentially silently---when the label-shift assumption is
wrong, e.g., under compound or concept shift (\S\ref{sec:lowdim}).

Recent deep variants move the correction into representation space:
RDALS~\cite{shi2026regularized} adapts under the label-shift assumption by a
regularized, discriminative alignment of source and target distributions in
a learned latent space, with accompanying theoretical analysis.

\subsubsection{Conditional and generalized label shift}
When $P(X\mid Y)$ changes, Zhang et al.~\cite{zhang2013domain} model the
change as a location--scale transformation of the class-conditionals and
estimate it jointly with the prior by kernel mean matching. In the deep
setting, GLS~\cite{tachet2020domain} motivates aligning \emph{class-conditional}
feature distributions after reweighting by estimated label proportions,
rather than aligning marginals as in DANN (\S\ref{sec:dda}).

\subsection{Distributionally Robust Optimization}
\label{sec:dro}

Rather than correcting for a specific shift, DRO minimizes the worst-case
risk over an \emph{uncertainty set} $\mathcal{U}(\Ps)$ of plausible targets,
\begin{equation}
\min_h\ \sup_{Q\in\mathcal{U}(\Ps)}\ \E_{Q}[\ell(h(X),Y)].
\label{eq:dro}
\end{equation}
For $f$-divergence balls, $\mathcal{U}=\{Q: D_f(Q\Vert\Ps)\le\rho\}$, the
inner supremum has a dual form that upweights high-loss examples; Duchi and
Namkoong~\cite{duchi2021learning} show that this controls performance on all
sufficiently large sub-populations, and Hashimoto et
al.~\cite{hashimoto2018fairness} use the $\chi^2$ version to protect latent
minority groups. Wasserstein balls allow $Q$ to move mass to nearby but
unseen inputs and admit tractable reformulations~\cite{mohajerin2018data};
Sinha et al.~\cite{sinha2018certifying} use a Lagrangian relaxation to
certify robustness to Wasserstein perturbations, connecting DRO with
adversarial training. Group DRO~\cite{sagawa2020distributionally} restricts
$\mathcal{U}$ to mixtures of known groups and minimizes the maximum group
risk, but requires strong regularization or early stopping: heavily
over-parameterized models otherwise drive all group training losses to zero,
removing the worst-case signal. The uncertainty set is a modeling
assumption; when it does not contain the actual shift (e.g., an
$f$-divergence ball cannot create support outside $\Ps$), DRO is not
guaranteed to help---a point the tabular evidence of \S\ref{sec:yx-shift}
makes concrete.

\subsection{Domain-Adaptation Theory}
\label{sec:dadtheory}

A complementary line bounds the target risk of \emph{any} hypothesis in
terms of a distance between domains, without fixing the shift type. This
theory underlies the design of most deep domain-adaptation algorithms
(\S\ref{sec:highdim}); Table~\ref{tab:theory} summarizes its main results.

\begin{definition}[$\Hcal\Delta\Hcal$-divergence~\cite{bendavid2007analysis,bendavid2010theory}]
For a class $\Hcal$ of binary classifiers, let
$\Hcal\Delta\Hcal=\{h\oplus h':h,h'\in\Hcal\}$. Then
$d_{\Hcal\Delta\Hcal}(\Ps,\Pt)=2\sup_{g\in\Hcal\Delta\Hcal}
\big|\Pr_{\Ps}[g(X)=1]-\Pr_{\Pt}[g(X)=1]\big|$.
\end{definition}

\begin{theorem}[Target risk bound~\cite{bendavid2010theory}]
\label{thm:bendavid}
For any $h\in\Hcal$,
\begin{equation}
\epsilon_T(h)\le\epsilon_S(h)+\tfrac12\,d_{\Hcal\Delta\Hcal}(\Ps,\Pt)+\lambda^*,
\label{eq:bendavid}
\end{equation}
where $\lambda^*=\min_{h\in\Hcal}\{\epsilon_S(h)+\epsilon_T(h)\}$ is the joint
risk of the ideal hypothesis.
\end{theorem}

Theorem~\ref{thm:bendavid} directly motivates the two-term objective
(source risk $+$ domain-alignment loss) of adversarial methods such as
DANN~\cite{ganin2016domain}, which approximates the divergence by a learned
domain discriminator, and of moment-matching methods such as
DAN~\cite{long2015learning} and Deep CORAL~\cite{sun2016deep}, which
substitute MMD~\eqref{eq:mmd} or covariance alignment.

\smallskip
\noindent\textbf{Refinements.} Mansour et al.~\cite{mansour2009domain}
generalize the bound to bounded losses satisfying a triangle inequality via
the \emph{discrepancy distance}, and Cortes and Mohri~\cite{cortes2014domain}
extend it to regression. Zhang et al.~\cite{zhang2019bridging} introduce the
\emph{margin disparity discrepancy} (MDD), a scoring-function-based
divergence that depends on the hypothesis being evaluated and yields a
tighter, margin-aware bound with a corresponding adversarial algorithm.
Acuna et al.~\cite{acuna2021f} generalize the discrepancy to arbitrary
$f$-divergences, showing that the original DANN objective corresponds to a
Jensen--Shannon instance. Shui et al.~\cite{shui2022novel} observe that the
$\Hcal$-divergence assumed in theory is not the Jensen--Shannon divergence
actually optimized in adversarial training, and derive upper \emph{and}
lower bounds in terms of the joint-distribution JS divergence that decompose
into marginal and conditional terms. Optimal-transport formulations replace
the divergence by a Wasserstein distance between joint
distributions~\cite{courty2017optimal}. We refer to Redko et
al.~\cite{redko2020survey} for a comprehensive treatment.

\smallskip
\noindent\textbf{Negative results.} The bound is not a license to align as
aggressively as possible. Ben-David et al.~\cite{bendavid2010impossibility}
prove impossibility results showing that small marginal divergence, or
covariate shift, taken alone does not guarantee successful adaptation.
Zhao et al.~\cite{zhao2019learning} prove an information-theoretic
\emph{lower} bound on the joint error of source and target when a
representation $Z$ makes the marginals $\Ps(Z)$ and $\Pt(Z)$ identical but the
label marginals differ: perfect marginal alignment then \emph{forces} error
on at least one domain. Johansson et al.~\cite{johansson2019support} show that
invariance can destroy information needed for prediction when the
representation is not invertible and that support overlap, not merely small
divergence, is essential. These results formalize why aggressive alignment
can increase $\lambda^*$ by collapsing class-discriminative structure
correlated with domain, and motivate the class-conditional and
label-shift-aware objectives of \S\ref{sec:classical-correction}.

\begin{table*}[t]
\centering
\footnotesize
\setlength{\tabcolsep}{4pt}
\renewcommand{\arraystretch}{1.12}
\caption{Theoretical results for learning under shift and the algorithmic
principle each motivates.}
\label{tab:theory}
\begin{tabular}{P{3.3cm}P{3.6cm}P{4.6cm}P{5.2cm}}
\toprule
\textbf{Result} & \textbf{Quantity controlled} & \textbf{Algorithmic consequence} & \textbf{Main limitation}\\
\midrule
IW generalization~\cite{shimodaira2000improving,cortes2010learning} & $\epsilon_T$ via $w=\Pt/\Ps$; variance via $d_2(\Pt\Vert\Ps)$ & IW-ERM, IWCV~\cite{sugiyama2007covariate} & Requires covariate shift and support overlap; high variance when $w$ is heavy-tailed\\
BBSE identifiability~\cite{lipton2018detecting,garg2020unified} & $\Pt(Y)$ from $\Pt(\hat Y)$ & Prior correction of a black-box model & Requires label shift and an invertible confusion matrix\\
$\Hcal\Delta\Hcal$ bound~\cite{bendavid2010theory} & $\epsilon_S + d_{\Hcal\Delta\Hcal} + \lambda^*$ & Adversarial / moment alignment~\cite{ganin2016domain,long2015learning} & Blind to label/concept shift; $\lambda^*$ unobservable\\
Discrepancy / MDD~\cite{mansour2009domain,zhang2019bridging} & Loss- and hypothesis-dependent divergence & MDD adversarial training & Still a marginal-only quantity\\
$f$-divergence / JS~\cite{acuna2021f,shui2022novel} & Joint divergence, marginal + conditional terms & Conditional matching + label reweighting & Conditional term needs pseudo-labels\\
Invariance lower bound~\cite{zhao2019learning} & Joint error under aligned marginals & Avoid pure marginal alignment under label shift; GLS~\cite{tachet2020domain} & Lower bound, no algorithm by itself\\
Impossibility~\cite{bendavid2010impossibility} & Necessity of assumptions & Make assumptions explicit (Prop.~\ref{prop:nonid}) & Negative result\\
$f$-DRO / Wasserstein DRO~\cite{duchi2021learning,mohajerin2018data} & Worst case over $\mathcal{U}(\Ps)$ & Loss upweighting; adversarial training~\cite{sinha2018certifying} & Uncertainty set may not contain real shift\\
Weighted conformal~\cite{tibshirani2019conformal} & Coverage under covariate shift & Calibrated prediction sets & Needs known or well-estimated $w$\\
\bottomrule
\end{tabular}
\end{table*}

\subsection{Two-Sample Testing and Shift Detection}
\label{sec:twosample}

Detecting \emph{whether} shift has occurred is the hypothesis test
$H_0:\Ps=\Pt$. Classical univariate tools include the Kolmogorov--Smirnov
(KS) test, applied per feature, and the Population Stability Index, popular
in credit-risk monitoring for its interpretability. For multivariate data,
the kernel two-sample test based on the Maximum Mean
Discrepancy~\cite{gretton2012kernel},
\begin{equation}
\mathrm{MMD}^2(\Ps,\Pt)=\big\|\E_{\Ps}[\phi(X)]-\E_{\Pt}[\phi(X)]\big\|^2_{\mathcal H},
\label{eq:mmd}
\end{equation}
is zero iff $\Ps=\Pt$ for a characteristic kernel and is estimable without
density estimation; MMD also serves as a domain-adaptation loss
(\S\ref{sec:dda}). Classifier two-sample tests~\cite{lopezpaz2017revisiting}
use the held-out accuracy of a domain classifier as a test statistic, and
deep-kernel MMD~\cite{liu2020learning} learns the kernel to maximize test
power. For high-dimensional inputs, Rabanser et al.~\cite{rabanser2019failing}
find empirically that testing on low-dimensional representations---in
particular the softmax outputs of a pretrained classifier (``black-box shift
detection'') combined with univariate tests and Bonferroni
aggregation---detects a broad range of synthetic shifts well. Detection of
\emph{any} shift is, however, not the operationally relevant question;
Podkopaev and Ramdas~\cite{podkopaev2022tracking} instead sequentially test
whether the \emph{risk} of a deployed model has increased beyond a tolerance,
yielding monitors that ignore benign shifts and raise alarms only on harmful
ones.

\subsection{Explaining and Diagnosing Shift}
\label{sec:diagnosis}

Moving beyond detection, recent work attributes an observed change to
specific mechanisms. Budhathoki et al.~\cite{budhathoki2021why} attribute a
change in a joint distribution to changes in individual causal mechanisms
via Shapley values over a known causal graph; Kulinski and
Inouye~\cite{kulinski2023towards} seek interpretable transport maps that
explain how source moves to target. For model performance, Chen et
al.~\cite{chen2022estimating} estimate and decompose performance change when
both covariates and labels shift, and Cai et al.~\cite{cai2023diagnosing}
decompose a performance drop into contributions from covariate and
conditional shift. Diagnosis matters because the correct remediation
(reweighting, retraining, feature-specific intervention or abstention)
depends on \emph{which} component moved (\S\ref{sec:lowdim}).

\subsection{Concept-Drift Detection in Streams}
\label{sec:driftdetection}

In streaming settings the goal is to detect \emph{when} the distribution
changes, often using only the online error signal because labels arrive
with delay. Following the surveys of Gama et al.~\cite{gama2014survey} and Lu
et al.~\cite{lu2018learning}, detectors fall into three families.
\emph{Sequential analysis} methods such as CUSUM/Page--Hinkley~\cite{page1954continuous}
accumulate $g_t=\max(0,g_{t-1}+x_t-\delta)$ and signal when $g_t$ exceeds a
threshold. \emph{Error-rate} methods such as DDM~\cite{gama2004learning}
model the online error as a Bernoulli process and signal when it exceeds its
historical minimum by a multiple of its standard deviation;
EDDM~\cite{baena2006early} monitors the distance between successive errors
to improve sensitivity to gradual drift. \emph{Window-based} methods such as
ADWIN~\cite{bifet2007learning} maintain an adaptive window that shrinks
whenever two sub-windows differ significantly, with guarantees on false
positive and false negative rates. After detection, adaptation typically
retrains on a recent window, updates incrementally (risking catastrophic
forgetting of recurring concepts), or maintains an ensemble of concept-specific
models~\cite{ditzler2015learning}. This literature is largely disjoint, in
venues and benchmarks, from online test-time adaptation
(\S\ref{sec:tta}), despite addressing a closely related problem.

\subsection{Label-Free Performance Estimation and Uncertainty}
\label{sec:perf-est}

A deployed system often needs to know \emph{how much} accuracy it has lost
without target labels. Average Thresholded Confidence
(ATC)~\cite{garg2022leveraging} learns a confidence threshold on source data
and predicts target accuracy as the fraction of target points above it;
Garg et al.\ also prove that, without assumptions, accuracy estimation under
shift is impossible, in line with Proposition~\ref{prop:nonid}. Difference
of Confidences (DoC)~\cite{guillory2021predicting} predicts the change in
accuracy from the change in average confidence between source and target, and Deng and Zheng~\cite{deng2021labels} regress accuracy
on a distributional distance computed over synthetic meta-datasets.
\emph{Agreement-on-the-line}~\cite{baek2022agreement} observes that, when the
ID--OOD accuracy of a model family lies on a line (\S\ref{sec:pretraining-robustness}),
the ID--OOD \emph{agreement} between pairs of models lies on the same line,
which permits label-free OOD accuracy prediction and, conversely, signals
when the linear trend fails.

Distribution-free uncertainty quantification extends to shift as well.
Weighted conformal prediction~\cite{tibshirani2019conformal} restores
marginal coverage under covariate shift by reweighting calibration scores
with $w$; Podkopaev and Ramdas~\cite{podkopaev2021distribution} do the
analogous correction under label shift; Barber et
al.~\cite{barber2023conformal} bound the coverage gap for non-exchangeable
data in terms of the total-variation distance between calibration and test
points; and adaptive conformal inference~\cite{gibbs2021adaptive} updates the
miscoverage level online, guaranteeing long-run coverage under arbitrary
drift.
\subsection{Distribution Shift in Low-Dimensional and Tabular Data}
\label{sec:lowdim}

Tabular data---records with a modest number of heterogeneous (continuous,
categorical, ordinal) features---remains the dominant modality in high-stakes
settings such as credit risk, healthcare records, public-sector
administrative data and industrial monitoring~\cite{gardner2023tableshift}.
Recent empirical work in this setting has produced findings that are, in our
view, under-appreciated by the largely vision-centric shift literature and
that bear directly on which of the corrections above is appropriate in
practice.

\subsubsection{The $Y\mid X$-Shift Finding}
\label{sec:yx-shift}

Most methodology---nearly all of \S\ref{sec:classical-correction} and the
feature-alignment methods of \S\ref{sec:dda}---is designed under an explicit
or implicit covariate-shift assumption. Its empirical prevalence had, until
recently, not been systematically measured.

Liu, Wang, Cui and Namkoong~\cite{liu2023need} address this directly with
\emph{WhyShift}, a testbed of naturally occurring (not synthetically
injected) shifts in tabular data spanning income prediction, public
coverage, health surveys and recidivism, with shifts across demographic
groups, geography and time. They attribute each observed performance drop to
covariate-shift and $Y\mid X$-shift components. In the conference version,
covering five tabular datasets and roughly 86{,}000 model configurations,
they report that of the 169 source--target pairs with a large performance
degradation (more than 8 accuracy points), about 80\% are primarily
attributable to $Y\mid X$ shift~\cite{liu2023need}. The extended
version~\cite{wang2024rethinking} enlarges the testbed to eight datasets,
172 distribution pairs, 45 methods and 90{,}000 configurations spanning ERM,
imbalanced-learning and DRO methods, and reports that (i) $Y\mid X$ shifts
are the most prevalent, in contrast to the literature's focus on covariate
shift; (ii) the performance of robust algorithms varies widely across shift
types and is, on average, no better than vanilla methods; and (iii)
implementation details---notably the underlying model class (e.g., gradient
boosted trees) and hyperparameter selection---affect performance more than
the choice of ambiguity set or its radius.

\begin{remark}[Methodological consequence]
If the dominant real-world shift is in $P(Y\mid X)$, then methods that
correct $P(X)$ (importance weighting, marginal feature alignment) target the
wrong quantity by construction, however well executed
(Proposition~\ref{prop:nonid}). This does not make them useless---compound
shifts have covariate components---but it means gains on benchmarks that
isolate covariate shift should not be assumed to transfer to settings where
$Y\mid X$ shift dominates. Liu et al.\ further find that a small labeled
validation sample from the target is disproportionately valuable, and that
non-algorithmic interventions (targeted data collection in the covariate
regions with the largest $Y\mid X$ shift) deserve greater
consideration~\cite{liu2023need}.
\end{remark}

TableShift~\cite{gardner2023tableshift} corroborates this from a different
angle. It comprises 15 binary classification tasks spanning finance,
education, public policy, healthcare and civic participation, each paired
with a naturally occurring shift. Its analysis finds an approximately
linear relation between in-distribution and out-of-distribution accuracy
across models, and a strong association between the shift gap and the
magnitude of \emph{label-distribution} shift---again indicating that the
structure practitioners encounter differs from the covariate-dominant
structure assumed by much of the methodological literature.

\subsubsection{Feature Shift: A Largely Unaddressed Axis}
\label{sec:feature-shift-tabular}

A second, more structural form of tabular shift is \emph{feature shift}
(\S\ref{sec:modern-types}): the set of available features changes between
training and deployment because of a sensor failure, a regulatory change or
integration of a new data source. TabFSBench~\cite{cheng2025tabfsbench} is
the first benchmark dedicated to this setting. It evaluates four feature-shift
scenarios (a single shifted feature, removal of the most and least relevant
features, and random removal) across classification and regression datasets
and across tree-based models, deep tabular models, and large language
models, including tabular-specific LLMs. Its main observations are that (i)
most tabular models have limited applicability under feature shift, (ii)
the importance of the shifted feature set is linearly related to the
resulting performance degradation, and (iii) closed-environment performance
correlates with feature-shift performance. Because the feature \emph{space}
changes, the covariate- and label-shift corrections of
\S\ref{sec:classical-correction}, which assume a fixed $\X$, do not directly
apply.

\subsubsection{Remediation Strategies}
\label{sec:tabular-remediation}

\noindent\textbf{Distributionally robust optimization.} The $\chi^2$-DRO of
Hashimoto et al.~\cite{hashimoto2018fairness} and group
DRO~\cite{sagawa2020distributionally} (\S\ref{sec:dro}) are the most common
robust objectives in tabular work. The WhyShift evidence that their
benefit over ERM is inconsistent on natural shift~\cite{wang2024rethinking}
suggests that implementation details (model class, regularization, group
granularity) matter more than the robust objective \emph{per se}, mirroring
DomainBed's conclusion for vision (\S\ref{sec:bench-image}).

\smallskip
\noindent\textbf{Causal feature selection.} If the shift is concept-level, a
natural remedy is to restrict the model to features whose relation to $Y$ is
stable across environments. Invariant causal prediction~\cite{peters2016causal}
uses multiple source environments to identify the set of predictors $S$ such
that $P(Y\mid X_S)$ is invariant, with confidence guarantees under the SCM
assumptions of \S\ref{sec:causal-view}; the guarantee extends to unseen
targets in which only non-causal associations have moved.

\smallskip
\noindent\textbf{In-context tabular foundation models.} TabPFN, a
transformer pretrained on synthetic datasets drawn from a prior over SCMs,
performs Bayesian-style prediction in a single forward pass by conditioning on
the full labeled training set~\cite{hollmann2025accurate}. Drift-Resilient
TabPFN~\cite{helli2024driftresilient} extends the prior to SCMs whose
parameters drift over time (modeled by a secondary SCM), so that the
in-context learner can extrapolate temporal shift without hyperparameter
tuning. On its 18-dataset evaluation, the authors report improving accuracy
from 0.688 to 0.744 and ROC-AUC from 0.786 to 0.832 over the strongest
baselines while maintaining better calibration.

\smallskip
\noindent\textbf{Test-time adaptation for tables.}
AdapTable~\cite{kim2024adaptable} adapts a pretrained tabular classifier at
test time without source data by (i) a shift-aware uncertainty calibrator,
a graph neural network over columns that produces per-sample temperatures,
and (ii) a label-distribution handler that re-estimates the target label
distribution of the current batch and rescales predictions accordingly---a
direct operationalization of the observation that label-distribution shift
drives much tabular degradation~\cite{gardner2023tableshift}. The authors
report gains of up to 16 points on HELOC. PFT3A~\cite{he2026prior} removes
the remaining dependence on source statistics and priors: it estimates class
priors and aligns feature distributions using low- and high-entropy target
samples as proxies for the source and target distributions, and aligns them
only in a low-dimensional informative subspace, addressing label and feature
shift jointly from unlabeled target batches. The authors also show that
TabPFN degrades substantially under shift.

\smallskip
\noindent\textbf{Are tabular foundation models robust?} Loza et
al.~\cite{loza2026empirical} evaluate nine tabular foundation models (TabPFN
v2--v3, TabICL, Mitra, LimiX, TabFM and variants) on TableShift tasks and
report that all degrade systematically under shift regardless of
pretraining strategy; real-data pretraining raises absolute OOD scores but
does not shrink the ID--OOD gap, and the type of shift, not the pretraining
corpus, is the main factor. In-context learning thus inherits rather than
removes the $Y\mid X$ problem of \S\ref{sec:yx-shift}.

\smallskip
\noindent\textbf{Explaining concept shift.} Diagnosis rather than
correction is often the actionable goal. SGShift~\cite{lyu2026explaining}
attributes the performance drop under concept shift in tabular data to a
sparse set of features whose relation to the label has changed, casting it
as feature selection with generalized additive models and knockoffs, and
requires few target samples---a feature-level complement to the
covariate-region attribution of Liu et al.~\cite{liu2023need}.

\subsubsection{Detecting Shift in Low-Dimensional Settings}
\label{sec:lowdim-detection}

Because tabular features are low-to-moderate dimensional and individually
interpretable, detection can often be performed per feature (KS tests,
population stability index, \S\ref{sec:twosample}) and aggregated with a
multiple-testing correction~\cite{rabanser2019failing}. Per-feature testing
localizes \emph{which} variables moved, information an aggregate statistic
such as MMD does not provide, and is one reason statistical process control
remains widely used in tabular monitoring. The localization question that
matters for remediation, however, is not which feature moved but which
\emph{factor} of $P(X,Y)$ moved; decomposition methods that attribute a
performance drop to covariate versus conditional components~\cite{cai2023diagnosing,chen2022estimating}
(\S\ref{sec:diagnosis}) address this directly but require at least a small
labeled target sample.
\section{Distribution Shift in High-Dimensional Data}
\label{sec:highdim}

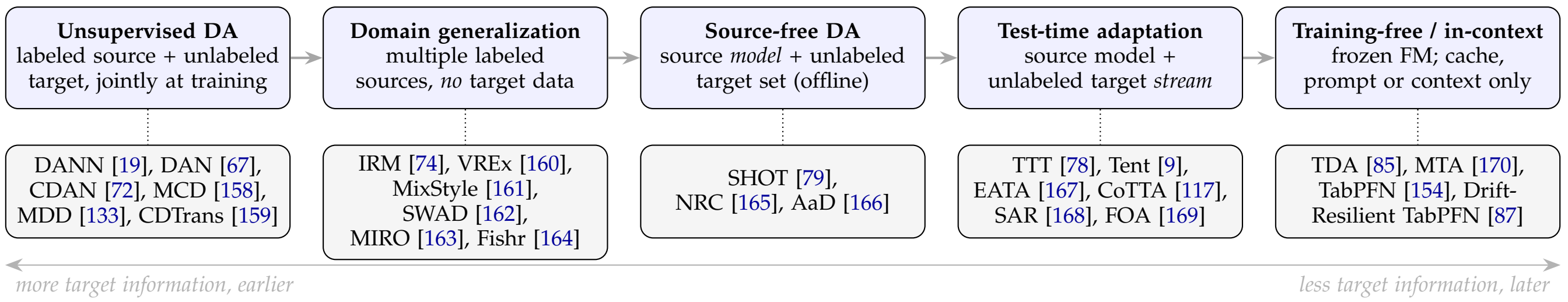
\begin{figure*}[t]
\centering
\begin{tikzpicture}[
  stage/.style={draw, rounded corners, font=\scriptsize, align=center, fill=blue!6,
                minimum width=3.15cm, minimum height=1.15cm, text width=3.0cm},
  mbox/.style={draw, rounded corners, font=\scriptsize, align=center, fill=gray!8,
               text width=3.0cm, minimum height=1.05cm},
  arr/.style={-{Stealth[length=2mm]}, thick, gray!70}
]
  \node[stage] (s1) at (0,0)    {\textbf{Unsupervised DA}\\labeled source + unlabeled target, jointly at training};
  \node[stage] (s2) at (3.6,0)  {\textbf{Domain generalization}\\multiple labeled sources, \emph{no} target data};
  \node[stage] (s3) at (7.2,0)  {\textbf{Source-free DA}\\source \emph{model} + unlabeled target set (offline)};
  \node[stage] (s4) at (10.8,0) {\textbf{Test-time adaptation}\\source model + unlabeled target \emph{stream}};
  \node[stage] (s5) at (14.4,0) {\textbf{Training-free / in-context}\\frozen FM; cache, prompt or context only};
  \foreach \a/\b in {s1/s2,s2/s3,s3/s4,s4/s5}{\draw[arr] (\a) -- (\b);}
  \node[mbox, below=4mm of s1] (m1) {DANN~\cite{ganin2016domain}, DAN~\cite{long2015learning}, CDAN~\cite{long2018conditional}, MCD~\cite{saito2018maximum}, MDD~\cite{zhang2019bridging}, CDTrans~\cite{xu2022cdtrans}};
  \node[mbox, below=4mm of s2] (m2) {IRM~\cite{arjovsky2019invariant}, VREx~\cite{krueger2021out}, MixStyle~\cite{zhou2021domain}, SWAD~\cite{cha2021swad}, MIRO~\cite{cha2022domain}, Fishr~\cite{rame2022fishr}};
  \node[mbox, below=4mm of s3] (m3) {SHOT~\cite{liang2020we}, NRC~\cite{yang2021exploiting}, AaD~\cite{yang2022attracting}};
  \node[mbox, below=4mm of s4] (m4) {TTT~\cite{sun2020test}, Tent~\cite{wang2021tent}, EATA~\cite{niu2022efficient}, CoTTA~\cite{wang2022continual}, SAR~\cite{niu2023towards}, FOA~\cite{niu2024test}};
  \node[mbox, below=4mm of s5] (m5) {TDA~\cite{karmanov2024efficient}, MTA~\cite{zanella2024test}, TabPFN~\cite{hollmann2025accurate}, Drift-Resilient TabPFN~\cite{helli2024driftresilient}};
  \foreach \i in {1,...,5}{\draw[densely dotted] (s\i) -- (m\i);}
  \draw[{Stealth[length=2mm]}-{Stealth[length=2mm]}, gray!60] ($(m1.south west)+(0,-0.3)$) -- ($(m5.south east)+(0,-0.3)$)
    node[pos=0.0, below right, font=\scriptsize\itshape]{more target information, earlier}
    node[pos=1.0, below left, font=\scriptsize\itshape]{less target information, later};
\end{tikzpicture}
\caption{Shift-handling settings for high-dimensional data, ordered by what
target information is available and when. Moving right removes access to
target data at training, then to source data, then to batch statistics and
gradients. Representative methods are listed below each setting.}
\label{fig:highdim-taxonomy}
\end{figure*}

High-dimensional data---images, video, audio and dense embeddings---differ
from the tabular setting in a way that has shaped the methodological
landscape: individual input dimensions carry no semantic meaning in
isolation, so neither per-feature detection (\S\ref{sec:lowdim-detection}) nor
per-feature causal selection (\S\ref{sec:tabular-remediation}) applies
directly. Shift is addressed predominantly through \emph{representation
learning}, and methods are best organized by a single distinction:
\emph{what target-domain information is available, and when}
(Fig.~\ref{fig:highdim-taxonomy} and Table~\ref{tab:settings}).

\begin{table}[t]
\centering
\footnotesize
\setlength{\tabcolsep}{2.5pt}
\caption{Data access by setting. $\mathcal{D}_S$: labeled source data; $f_S$:
source-trained model; $\mathcal{D}_T^u$: unlabeled target data;
$\mathcal{D}_T^\ell$: labeled target data. ``Online'': predictions must be
issued as data arrive.}
\label{tab:settings}
\begin{tabular}{lcccccc}
\toprule
\textbf{Setting} & $\mathcal{D}_S$ & $f_S$ & $\mathcal{D}_T^u$ & $\mathcal{D}_T^\ell$ & \textbf{Online} & \textbf{Multi-src.}\\
\midrule
Fine-tuning / few-shot & \halfmark & \cmark & \xmark & \cmark & \xmark & \xmark\\
Unsupervised DA & \cmark & -- & \cmark & \xmark & \xmark & \halfmark\\
Domain generalization & \cmark & -- & \xmark & \xmark & \xmark & \cmark\\
Source-free DA & \xmark & \cmark & \cmark & \xmark & \xmark & \xmark\\
Test-time adaptation & \xmark & \cmark & \cmark & \xmark & \cmark & \xmark\\
Federated DA/DG & \halfmark & \cmark & \halfmark & \xmark & \xmark & \cmark\\
\bottomrule
\end{tabular}
\end{table}

\subsection{Deep Domain Adaptation}
\label{sec:dda}

Unsupervised domain adaptation (UDA) assumes labeled source and unlabeled
target data at training time~\cite{wilson2020survey,kouw2021review}. Methods
are best organized by which proxy for the divergence in
Theorem~\ref{thm:bendavid} they optimize.

\smallskip
\noindent\textbf{Statistic matching.} DAN~\cite{long2015learning} minimizes
multi-kernel MMD~\eqref{eq:mmd} between source and target features at several
layers; Deep CORAL~\cite{sun2016deep} aligns second-order statistics; and
JAN~\cite{long2017deep} matches the \emph{joint} distribution of activations
across layers. These losses are cheap and stable but control only selected
moments.

\smallskip
\noindent\textbf{Adversarial alignment.} DANN~\cite{ganin2016domain} trains
a feature extractor to fool a domain discriminator through a gradient
reversal layer, implementing a learned approximation of
$d_{\Hcal\Delta\Hcal}$; ADDA~\cite{tzeng2017adversarial} uses untied encoders
and a GAN-style objective. CDAN~\cite{long2018conditional} conditions the
discriminator on the classifier's predictions (via multilinear conditioning),
aligning the joint distribution of features and predictions---an empirical
counterpart of the conditional terms in the JS-based
theory~\cite{shui2022novel}. MCD~\cite{saito2018maximum} replaces the domain
discriminator by the disagreement between two classifiers, and
MDD~\cite{zhang2019bridging} derives an adversarial objective from its
margin-based bound. Pixel-level methods such as CyCADA~\cite{hoffman2018cycada}
translate source images into the target style before training.

\smallskip
\noindent\textbf{Optimal transport and self-training.} DeepJDOT~\cite{damodaran2018deepjdot}
minimizes an OT distance between joint feature--label distributions, using
pseudo-labels on the target side. Self-training with class-balanced
pseudo-labels~\cite{zou2018unsupervised} and, more recently, teacher--student
consistency have become dominant in dense prediction.

\smallskip
\noindent\textbf{Transformers and dense prediction.} Transformer backbones
enable cross-attention between source and target
samples (CDTrans~\cite{xu2022cdtrans}). For semantic segmentation, adaptation
moved from output-space adversarial alignment
(AdaptSegNet~\cite{tsai2018learning}) to transformer-based self-training with
rare-class sampling and ImageNet feature distance
(DAFormer~\cite{hoyer2022daformer}) and masked-image consistency
(MIC~\cite{hoyer2023mic}); for detection, Domain Adaptive Faster
R-CNN~\cite{chen2018domain} aligns at image and instance levels.

\smallskip
\noindent\textbf{Label-set mismatch.} Closed-set UDA assumes
$\Y_S=\Y_T$. When the label sets differ, alignment of all target data to all
source data causes \emph{negative transfer}. Table~\ref{tab:labelsets}
summarizes the variants; their common principle is to weight or reject
samples before alignment, which connects UDA to the open-set detection
problem of \S\ref{sec:ood}.

\begin{table}[t]
\centering
\footnotesize
\setlength{\tabcolsep}{3pt}
\caption{Domain adaptation variants by relation between source and target
label sets.}
\label{tab:labelsets}
\begin{tabular}{P{1.5cm}P{1.55cm}P{3.0cm}P{1.6cm}}
\toprule
\textbf{Setting} & \textbf{Label sets} & \textbf{Key difficulty} & \textbf{Example}\\
\midrule
Closed-set & $\Y_T=\Y_S$ & covariate / conditional shift & \cite{ganin2016domain,long2018conditional}\\
Partial & $\Y_T\subset\Y_S$ & outlier source classes cause negative transfer & PADA~\cite{cao2018partial}\\
Open-set & $\Y_T\supset\Y_S$ & reject unknown target classes & OSBP~\cite{saito2018open}\\
Universal & unknown overlap & both private source and target classes & UAN~\cite{you2019universal}, DANCE~\cite{saito2020universal}\\
\bottomrule
\end{tabular}
\end{table}

\smallskip
\noindent\textbf{Shared limitation.} Marginal alignment implicitly targets
covariate shift. Under label shift, perfect marginal alignment provably
increases error on at least one domain~\cite{zhao2019learning}; under concept
shift it can destroy discriminative information~\cite{johansson2019support}.
Class-conditional alignment with label-proportion
reweighting~\cite{tachet2020domain} is the theoretically sound repair when
GLS holds.

\subsection{Domain Generalization}
\label{sec:dg}

Domain generalization (DG)~\cite{blanchard2011generalizing,muandet2013domain}
removes access to target data entirely: a model trained on several source
domains must generalize to an unseen target. The absence of target data
forces DG methods to assume \emph{what remains invariant} across domains.
Zhou et al.~\cite{zhou2022domain} and Wang et al.~\cite{wang2023generalizing}
provide comprehensive surveys; we summarize the main families and the
evaluation results that qualify them.

\smallskip
\noindent\textbf{Invariance and causality.} IRM~\cite{arjovsky2019invariant}
seeks a representation $\Phi$ on which a single classifier is simultaneously
optimal in every training environment, a representation-learning analogue of
invariant causal prediction~\cite{peters2016causal}.
Rosenfeld et al.~\cite{rosenfeld2021risks} show that in the linear case IRM
can recover the invariant predictor only when the number of environments
exceeds the dimension of the spurious features, and that in the non-linear
case it can fail catastrophically while appearing to succeed on training
environments. V-REx~\cite{krueger2021out} penalizes the variance of risks
across environments; Fishr~\cite{rame2022fishr} matches the variance of
per-domain gradients; causal matching~\cite{mahajan2021domain} uses an SCM
over domain, class and object to derive a contrastive matching objective;
CIRL~\cite{lv2022causality} synthesizes interventions on hypothesized
non-causal factors; and Salaudeen and Koyejo~\cite{salaudeen2024causally}
enforce graph-implied conditional independencies without requiring
observation of the spurious features.

\smallskip
\noindent\textbf{Data augmentation and style randomization.} Mixing
instance-level feature statistics across domains
(MixStyle~\cite{zhou2021domain}) or reducing style bias~\cite{nam2021reducing}
exposes the model to synthetic variation, on the premise that robustness to
simulated variation transfers to real variation.

\smallskip
\noindent\textbf{Meta-learning, flatness and ensembling.} MLDG~\cite{li2018learning}
simulates train/test domain splits episodically. A more consistently
effective family seeks flat minima: SWAD~\cite{cha2021swad} densely averages
weights along the training trajectory and bounds the DG gap by a robust-risk
term, and SAGM~\cite{wang2023sharpness} aligns the gradients of the empirical
and perturbed losses. Weight averaging across independently fine-tuned
models (model soups~\cite{wortsman2022model}, DiWA~\cite{rame2022diverse})
exploits the diversity of their errors. With pretrained backbones,
MIRO~\cite{cha2022domain} regularizes the mutual information between the
fine-tuned and pretrained features, foreshadowing the robust fine-tuning
methods of \S\ref{sec:finetuning-tradeoff}.

\smallskip
\noindent\textbf{What the evaluations show.} Under a consistent
model-selection protocol, ERM is competitive with most DG algorithms on
DomainBed~\cite{gulrajani2021search}; methods that beat ERM on
correlation shift typically fail to on diversity shift and vice
versa~\cite{ye2022ood}; and, theoretically, no fixed independence constraint
works across all realizable shifts~\cite{kaur2023modeling}. We read these
results as evidence that DG performance is governed by whether a method's
(often implicit) invariance assumption matches the data-generating process,
echoing \S\ref{sec:causal-view}.

\subsection{Sub-population Shift and Spurious Correlations}
\label{sec:subpop}

Deep networks readily exploit \emph{shortcuts}---features predictive in the
training distribution but not causally related to the
label~\cite{geirhos2020shortcut}. When group annotations are available,
group DRO~\cite{sagawa2020distributionally} minimizes worst-group risk.
Without them, JTT~\cite{liu2021just} first trains an ERM model and then
upweights its training errors as a proxy for minority groups, and
DFR~\cite{kirichenko2023last} shows that ERM features often already contain
the core features and that re-training only the last layer on a small
group-balanced set suffices for strong worst-group accuracy. Yang et
al.~\cite{yang2023change} benchmark these methods across spurious
correlations, attribute imbalance, class imbalance and attribute
generalization, finding that progress is concentrated on spurious
correlations and class imbalance while attribute generalization remains
largely unsolved, and that model selection with worst-class rather than
worst-group accuracy is a reasonable proxy when group labels are unavailable.

\subsection{Source-Free Domain Adaptation}
\label{sec:sfda}

Source-free domain adaptation (SFDA) assumes access only to a source-trained
model and an unlabeled target set that may be revisited offline, a setting
motivated by privacy and data-governance constraints~\cite{li2024comprehensive,fang2024sourcefree}.
SHOT~\cite{liang2020we} freezes the source classifier and adapts the feature
extractor by information maximization (confident individual predictions,
diverse batch predictions) and self-supervised pseudo-labeling.
Neighborhood-based methods exploit the cluster structure of target features:
NRC~\cite{yang2021exploiting} enforces consistency among reciprocal nearest
neighbors, and AaD~\cite{yang2022attracting} attracts predictions of
neighbors while dispersing those of non-neighbors. Li et
al.~\cite{li2024comprehensive} categorize SFDA into data-based (e.g.,
generating source-like samples) and model-based (self-training, consistency)
approaches and benchmark over 30 methods on Office-31, Office-Home and VisDA;
Fang et al.~\cite{fang2024sourcefree} additionally distinguish white-box from
black-box SFDA, in which only the source model's outputs are available.

\subsection{Test-Time Adaptation}
\label{sec:tta}

Test-time adaptation (TTA) adapts a source model using only unlabeled test
data encountered at inference. Liang et al.~\cite{liang2024comprehensive}
distinguish \emph{test-time domain adaptation} (the full unlabeled target set
is available; equivalent to SFDA), \emph{test-time batch adaptation}
(each batch or instance is adapted independently) and \emph{online} TTA (a
stream, possibly non-stationary), and treat test-time prior adaptation
(label shift at test time) separately. Table~\ref{tab:tta} summarizes
representative methods.

\smallskip
\noindent\textbf{Normalization and entropy.} Re-estimating batch-normalization
statistics on test batches already improves robustness to common
corruptions~\cite{schneider2020improving}---an observation consistent with,
but terminologically distinct from, the ``internal covariate shift'' that
motivated BN~\cite{ioffe2015batch}. Tent~\cite{wang2021tent} additionally
updates BN affine parameters by minimizing prediction entropy.
EATA~\cite{niu2022efficient} filters unreliable and redundant samples and adds
a Fisher-based anti-forgetting regularizer; SAR~\cite{niu2023towards}
identifies batch normalization and large-gradient samples as sources of
collapse under small batches, mixed shifts and label imbalance, and
proposes sharpness-aware, reliable entropy minimization;
DeYO~\cite{lee2024entropy} argues that entropy is insufficient and uses a
shape-sensitive confidence signal. Conjugate pseudo-labels~\cite{goyal2022test}
derive the adaptation loss from the convex conjugate of the training loss,
recovering entropy minimization for cross-entropy.

\smallskip
\noindent\textbf{Self-supervised test-time training.}
TTT~\cite{sun2020test} trains a self-supervised auxiliary task jointly with
the main task and optimizes it on each test input; TTT++~\cite{liu2021ttt}
analyzes when this helps and adds feature-alignment regularization; and
TTT-MAE~\cite{gandelsman2022test} uses masked autoencoding as the auxiliary
task.

\smallskip
\noindent\textbf{Backpropagation-free and output-level adaptation.}
T3A~\cite{iwasawa2021test} adjusts only the classifier with prototypes
built from test features; LAME~\cite{boudiaf2022parameter} adapts only the
outputs via Laplacian-regularized maximum likelihood, avoiding parameter
updates. FOA~\cite{niu2024test} learns an input prompt with a
derivative-free covariance-matrix-adaptation evolution strategy and shifts
activations toward source statistics, enabling adaptation of quantized
models that do not support backpropagation. MEMO~\cite{zhang2022memo}
minimizes the marginal entropy over augmentations of a single test point.

\smallskip
\noindent\textbf{Continual and dynamic shift.} CoTTA~\cite{wang2022continual}
uses weight-averaged and augmentation-averaged teacher predictions and
stochastic restoration of source weights to limit error accumulation;
NOTE~\cite{gong2022note} and RoTTA~\cite{yuan2023robust} target temporally
correlated (non-i.i.d.) streams.

\smallskip
\noindent\textbf{Re-evaluations.} Zhao et al.~\cite{zhao2023pitfalls} show
that TTA results are highly sensitive to hyperparameter selection performed
with access to target labels, to the source model's quality, and to the
type of shift, and that no method handles all shift types. Press et
al.~\cite{press2023rdumb} find that periodically resetting the model to its
pretrained weights (RDumb) matches or outperforms many continual-TTA methods
over long horizons. Lim et al.~\cite{lim2026when} refine this into
\emph{adaptive and selective} resets: a reset is triggered when the
concentration of predictions on few classes exceeds its running average, and
only layers from the output side are reset, in proportion to that excess;
the authors report raising accuracy on the CCC-Hard long-horizon benchmark
from 15.41\% (the best prior reset method) to 22.21\%. Amar and
Lindenbaum~\cite{amar2026controlling} take the same collapse indicator---a
narrowing set of predicted classes---as a quantity to control: SCALAR
keeps a buffer of recent predictions, balances their class marginal towards a
target distribution with the Sinkhorn algorithm, and adapts on the balanced
soft targets, which is a label-shift-aware correction (\S\ref{sec:bbse})
applied over a temporal window. Wang et al.~\cite{wang2025search} benchmark online TTA
with Vision Transformers and find that transformers are more resilient to
domain shift, that many methods depend on large batches, and that stability
is critical at batch size one; they also introduce efficiency metrics
(GFLOPs, wall-clock time, memory). Xiao and Snoek~\cite{xiao2024beyond}
broaden the taxonomy to what is adapted---model, inference, normalization,
sample, or prompt.

\smallskip
\noindent\textbf{Optimization-free adaptation.} NEO~\cite{murphy2026neo}
observes that, for ViTs, covariate shift moves penultimate-layer embeddings
along a few directions shared across samples and classes, and adapts by
subtracting the running mean of test embeddings before the linear head; it
uses no backpropagation or hyperparameters, and the authors report gains on
ImageNet-C from a single batch of 64 samples. It is the embedding-space
analogue of test-time BN re-estimation~\cite{schneider2020improving}.

\smallskip
\noindent\textbf{Generative adaptation of the input.} Instead of updating
the model, DDA~\cite{gao2023back} projects corrupted test inputs back toward
the source domain with a diffusion model trained on source data, which is
robust to small batches and non-stationary orders because no parameters
are updated.

\begin{table*}[t]
\centering
\footnotesize
\setlength{\tabcolsep}{4pt}
\renewcommand{\arraystretch}{1.08}
\caption{Representative test-time and source-free adaptation methods.
``Backprop'': whether adaptation requires gradients; ``Batch'': whether the
method relies on batch-level statistics; ``Online'': designed for streams.}
\label{tab:tta}
\begin{tabular}{P{2.3cm}P{1.6cm}P{3.0cm}P{4.3cm}ccc}
\toprule
\textbf{Method} & \textbf{Venue} & \textbf{What is adapted} & \textbf{Objective / signal} & \textbf{Backprop} & \textbf{Batch} & \textbf{Online}\\
\midrule
SHOT~\cite{liang2020we} & ICML'20 & feature extractor & information maximization + pseudo-labels & \cmark & \cmark & \xmark\\
NRC~\cite{yang2021exploiting} & NeurIPS'21 & full network & neighborhood consistency & \cmark & \cmark & \xmark\\
BN-adapt~\cite{schneider2020improving} & NeurIPS'20 & BN statistics & re-estimated moments & \xmark & \cmark & \cmark\\
TTT~\cite{sun2020test} & ICML'20 & shared encoder & self-supervised auxiliary loss & \cmark & \xmark & \cmark\\
Tent~\cite{wang2021tent} & ICLR'21 & BN affine & entropy & \cmark & \cmark & \cmark\\
T3A~\cite{iwasawa2021test} & NeurIPS'21 & classifier prototypes & prototype adjustment & \xmark & \xmark & \cmark\\
EATA~\cite{niu2022efficient} & ICML'22 & BN affine & filtered entropy + Fisher regularizer & \cmark & \cmark & \cmark\\
CoTTA~\cite{wang2022continual} & CVPR'22 & full network & teacher consistency + stochastic restore & \cmark & \cmark & \cmark\\
LAME~\cite{boudiaf2022parameter} & CVPR'22 & outputs only & Laplacian-regularized likelihood & \xmark & \cmark & \cmark\\
MEMO~\cite{zhang2022memo} & NeurIPS'22 & full network & marginal entropy over augmentations & \cmark & \xmark & \cmark\\
SAR~\cite{niu2023towards} & ICLR'23 & norm affine & sharpness-aware reliable entropy & \cmark & \cmark & \cmark\\
DDA~\cite{gao2023back} & CVPR'23 & input (diffusion) & projection to source manifold & \xmark & \xmark & \cmark\\
DeYO~\cite{lee2024entropy} & ICLR'24 & norm affine & shape-aware confidence & \cmark & \cmark & \cmark\\
FOA~\cite{niu2024test} & ICML'24 & input prompt + activations & CMA-ES on entropy + statistic discrepancy & \xmark & \cmark & \cmark\\
\bottomrule
\end{tabular}
\end{table*}

\subsection{Surgical and Selective Fine-Tuning}
\label{sec:surgical}

Adapting all parameters is often suboptimal. Lee et al.~\cite{lee2023surgical}
show that fine-tuning only a subset of layers outperforms full fine-tuning
under several shift types, and that the best block depends on the shift:
early layers for input-level shifts such as corruptions, later layers for
shifts in the label mechanism. This aligns the taxonomy of
\S\ref{sec:taxonomy} with network depth and anticipates the
parameter-efficient strategies used to preserve foundation-model robustness
(\S\ref{sec:foundation}).

\begin{remark}[A disconnect worth naming]
The streaming drift literature (\S\ref{sec:driftdetection}) and the online
TTA literature address structurally similar problems---a non-stationary
stream and a model that must keep pace---with almost entirely disjoint
methods, benchmarks and venues. Drift detectors decide \emph{when} to adapt
and assume labels eventually arrive; TTA methods adapt continuously without
labels and rarely decide whether to adapt at all. RDumb's
finding~\cite{press2023rdumb} that periodic resets are a strong baseline is,
in effect, a drift-agnostic adaptation schedule, suggesting that combining
explicit change detection with TTA is a natural and under-explored direction
(\S\ref{sec:challenges}).
\end{remark}
\section{Distribution Shift in Federated and Distributed Learning}
\label{sec:federated}

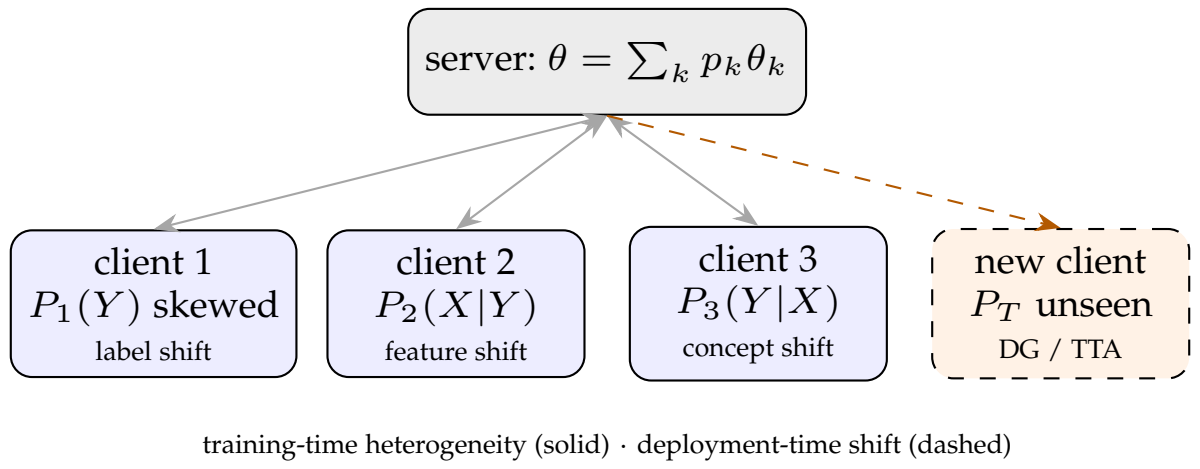
\begin{figure}[t]
\centering
\begin{tikzpicture}[
  font=\scriptsize, >=Stealth,
  srv/.style={draw, rounded corners, fill=gray!15, minimum width=2.6cm, minimum height=0.7cm, align=center},
  cl/.style={draw, rounded corners, fill=blue!7, minimum width=1.7cm, minimum height=0.95cm, align=center},
  nc/.style={draw, dashed, rounded corners, fill=orange!10, minimum width=1.7cm, minimum height=0.95cm, align=center}
]
\node[srv] (s) at (0,0) {server: $\theta=\sum_k p_k\theta_k$};
\node[cl] (c1) at (-3.0,-1.6) {client 1\\$P_1(Y)$ skewed\\{\tiny label shift}};
\node[cl] (c2) at (-1.0,-1.6) {client 2\\$P_2(X|Y)$\\{\tiny feature shift}};
\node[cl] (c3) at (1.0,-1.6) {client 3\\$P_3(Y|X)$\\{\tiny concept shift}};
\node[nc] (c4) at (3.0,-1.6) {new client\\$\Pt$ unseen\\{\tiny DG / TTA}};
\foreach \c in {c1,c2,c3}{\draw[<->, gray!70] (s.south) -- (\c.north);}
\draw[->, dashed, orange!70!black] (s.south) -- (c4.north);
\node[font=\tiny, align=center] at (0,-2.55) {training-time heterogeneity (solid) $\cdot$ deployment-time shift (dashed)};
\end{tikzpicture}
\caption{Federated learning as distribution shift. Clients differ in label,
feature or concept distribution during training; at deployment, the global
model may face an unseen client or a drifting local distribution.}
\label{fig:fl}
\end{figure}

Federated learning (FL) trains a shared model across $K$ clients that keep
their data local~\cite{mcmahan2017communication,kairouz2021advances}. Each
client $k$ draws data from its own distribution $P_k(X,Y)$, so FL is
distribution shift by construction: across clients during training, between
the training federation and new clients or deployment conditions at test
time, and within each client over time. Although usually discussed under
the label ``non-IID data''~\cite{zhao2018federated,zhu2021federated}, the
heterogeneity types studied in FL map one-to-one onto the taxonomy of
\S\ref{sec:taxonomy} (Table~\ref{tab:fl}, Fig.~\ref{fig:fl}).

\subsection{Formulation}
\label{sec:fl-formulation}

The standard objective minimizes a weighted average of client risks,
\begin{equation}
\min_\theta\ F(\theta)=\sum_{k=1}^K p_k\,\E_{P_k}[\ell(h_\theta(X),Y)],
\label{eq:fedavg}
\end{equation}
which is ERM on the mixture $\bar P=\sum_k p_k P_k$. FedAvg~\cite{mcmahan2017communication}
approximates its minimizer by averaging locally updated models. Three
deployment targets must be distinguished: (i) the mixture $\bar P$ itself;
(ii) each participating client's own $P_k$ (\emph{personalization}); and
(iii) an unseen client or condition $\Pt\notin\{P_k\}$ (\emph{federated domain
generalization}). Agnostic FL~\cite{mohri2019agnostic} replaces
\eqref{eq:fedavg} by a minimax over mixture weights,
$\min_\theta\max_{\boldsymbol\lambda\in\Lambda}\sum_k\lambda_k\E_{P_k}[\ell]$,
which is exactly group DRO (\S\ref{sec:dro}) with clients as groups.

\subsection{Heterogeneity Types as Shift Types}
\label{sec:fl-types}

\noindent\textbf{Label distribution skew} ($P_k(Y)$ differs, $P_k(X\mid Y)$
shared) is label shift across clients. It causes local optima to diverge
(``client drift''), degrading FedAvg~\cite{zhao2018federated}; FedProx~\cite{li2020federated}
adds a proximal term to local objectives and SCAFFOLD~\cite{karimireddy2020scaffold}
corrects client drift with control variates.

\noindent\textbf{Feature distribution skew} ($P_k(X\mid Y)$ or $P_k(X)$
differs) is conditional or covariate shift across clients, e.g., different
scanners across hospitals. FedBN~\cite{li2021fedbn} keeps batch-normalization
layers local while sharing the remaining parameters, which the authors show
both empirically and in a simplified analysis to improve convergence under
feature shift. Hsieh et al.~\cite{hsieh2020non} identify batch normalization
as a specific failure point under skewed label partitions and propose group
normalization as a remedy, illustrating that normalization statistics are a
recurring locus of shift sensitivity (cf.\ \S\ref{sec:tta}).

\noindent\textbf{Concept shift across clients} ($P_k(Y\mid X)$ differs) makes a
single global model inappropriate. Clustered FL groups clients with
compatible conditionals---IFCA~\cite{ghosh2020efficient} alternates cluster
assignment and per-cluster optimization, and CFL~\cite{sattler2021clustered}
splits clients recursively by the cosine similarity of their updates---while
personalized FL learns client-specific models via meta-learning
(Per-FedAvg~\cite{fallah2020personalized}) or regularization toward the global
model (Ditto~\cite{li2021ditto}, which the authors show also improves
fairness and robustness).

\noindent\textbf{Test-time and temporal shift.} Clients encounter new
conditions after training. FedTHE+~\cite{jiang2023test} ensembles a global
and a personalized head with test-time adaptive weights and is evaluated
under simultaneous label-distribution shift, natural covariate shift and
corruptions; ATP~\cite{bao2023adaptive} learns module-wise adaptation rates
during training so that an unlabeled test client can adapt in an
unsupervised way, and reports that the learned rates differ by shift type.

\noindent\textbf{Unseen domains.} Federated adversarial domain
adaptation~\cite{peng2020federated} aligns each source client to an
unlabeled target without centralizing data, and FedDG~\cite{liu2021feddg}
exchanges amplitude spectra across clients to simulate domain variation for
medical image segmentation without sharing raw images. The benchmark of Bai
et al.~\cite{bai2024benchmarking}, which evaluates 14 federated DG methods
while controlling the number of clients and their heterogeneity, finds that
significant performance gaps remain, especially with many clients, high
client heterogeneity, or more realistic datasets.

\begin{table}[t]
\centering
\footnotesize
\setlength{\tabcolsep}{2.5pt}
\caption{Federated heterogeneity types mapped onto the shift taxonomy of
Table~\ref{tab:shift-types}.}
\label{tab:fl}
\begin{tabular}{P{1.7cm}P{1.45cm}P{1.6cm}P{3.2cm}}
\toprule
\textbf{FL term} & \textbf{Changes} & \textbf{Shift type} & \textbf{Representative remedy}\\
\midrule
Label skew & $P_k(Y)$ & label & FedProx~\cite{li2020federated}, SCAFFOLD~\cite{karimireddy2020scaffold}\\
Feature skew & $P_k(X\mid Y)$ & conditional / covariate & FedBN~\cite{li2021fedbn}, GroupNorm~\cite{hsieh2020non}\\
Concept skew & $P_k(Y\mid X)$ & concept & IFCA~\cite{ghosh2020efficient}, CFL~\cite{sattler2021clustered}, Ditto~\cite{li2021ditto}\\
Unequal mixtures & $p_k$ & sub-population & Agnostic FL~\cite{mohri2019agnostic}\\
New client / domain & $\Pt\notin\{P_k\}$ & domain (DG) & FedDG~\cite{liu2021feddg}, FADA~\cite{peng2020federated}\\
Test-time drift & $P_k^{(t)}$ & temporal & FedTHE+~\cite{jiang2023test}, ATP~\cite{bao2023adaptive}\\
\bottomrule
\end{tabular}
\end{table}

\subsection{Open Issues Specific to Federated Shift}
\label{sec:fl-open}

Three properties make federated shift harder than its centralized
counterpart. First, \emph{detection and diagnosis are constrained}: the
two-sample tests and decompositions of \S\ref{sec:twosample}--\S\ref{sec:diagnosis}
assume pooled access to samples, which FL forbids, so shift must be
characterized from model updates or privatized summaries. Second,
\emph{importance weighting is ill-posed across clients}, because estimating
$P_k(x)/P_{k'}(x)$ requires either sharing data or training domain
classifiers across clients, which conflicts with privacy constraints.
Third, \emph{model selection for DG is ambiguous}: holding out an entire
client as a validation domain reduces the already limited diversity of
training domains. The centralized evidence that implementation details and
model selection dominate algorithmic choice (\S\ref{sec:yx-shift},
\S\ref{sec:dg}) suggests that controlled federated benchmarks such as that
of Bai et al.~\cite{bai2024benchmarking} are a prerequisite for credible
progress.

\smallskip
\noindent\textbf{Recent directions (2026).} Three lines address
deployment-time shift explicitly. Feroma~\cite{li2026federated} builds
compact, privacy-preserving distribution profiles of clients and uses them
both to choose aggregation (from clustered to personalized) during training
and to assign models to unseen, unlabeled test clients without retraining,
handling shift across clients and drift over time without client identities.
Fed-ADE~\cite{park2026fedade} targets federated post-deployment adaptation
under client-specific, time-varying shift, using lightweight estimators of
distribution dynamics to set per-client learning rates. In federated
continual TTA, where unlabeled clients arrive sequentially, Wang et
al.~\cite{wang2026towards} stabilize the global model against the
forgetting that naive TTA induces. These works operationalize the
``test-time drift'' row of Table~\ref{tab:fl}.
\section{Connections to Out-of-Distribution, Open-Set and Anomaly Detection}
\label{sec:ood}

Everything discussed so far concerns \emph{generalization} under shift:
given $\Pt\neq\Ps$, how can a model perform well on $\Pt$? A distinct but
closely related question is \emph{detection}: given a test input, decide
whether it plausibly came from $\Ps$ at all, so that the system can abstain,
defer to a human, or route the input elsewhere. This is the domain of OOD
detection, anomaly detection (AD), novelty detection (ND), open-set
recognition (OSR) and outlier detection (OD), which developed largely in
isolation before being unified in terms of the shift taxonomy of
\S\ref{sec:taxonomy}.

\subsection{The Generalized OOD Detection Framework}
\label{sec:goodd}

Yang et al.~\cite{yang2024generalized} observe that AD, ND, OSR, OOD
detection and OD share the goal of flagging test samples that deviate from
an in-distribution (ID) set under an open-world assumption, and differ in
(i) the type of shift to be detected (covariate or semantic), (ii) whether
the ID data comprise one or many classes, (iii) whether ID classification is
also required, and (iv) whether the setting is inductive or transductive.

\begin{definition}[Semantic shift]
A test input exhibits semantic shift relative to an ID label space
$\Y_{\mathrm{ID}}$ if its true label satisfies $y\notin\Y_{\mathrm{ID}}$;
this is the open-set shift of \S\ref{sec:modern-types} at the level of a
single sample.
\end{definition}

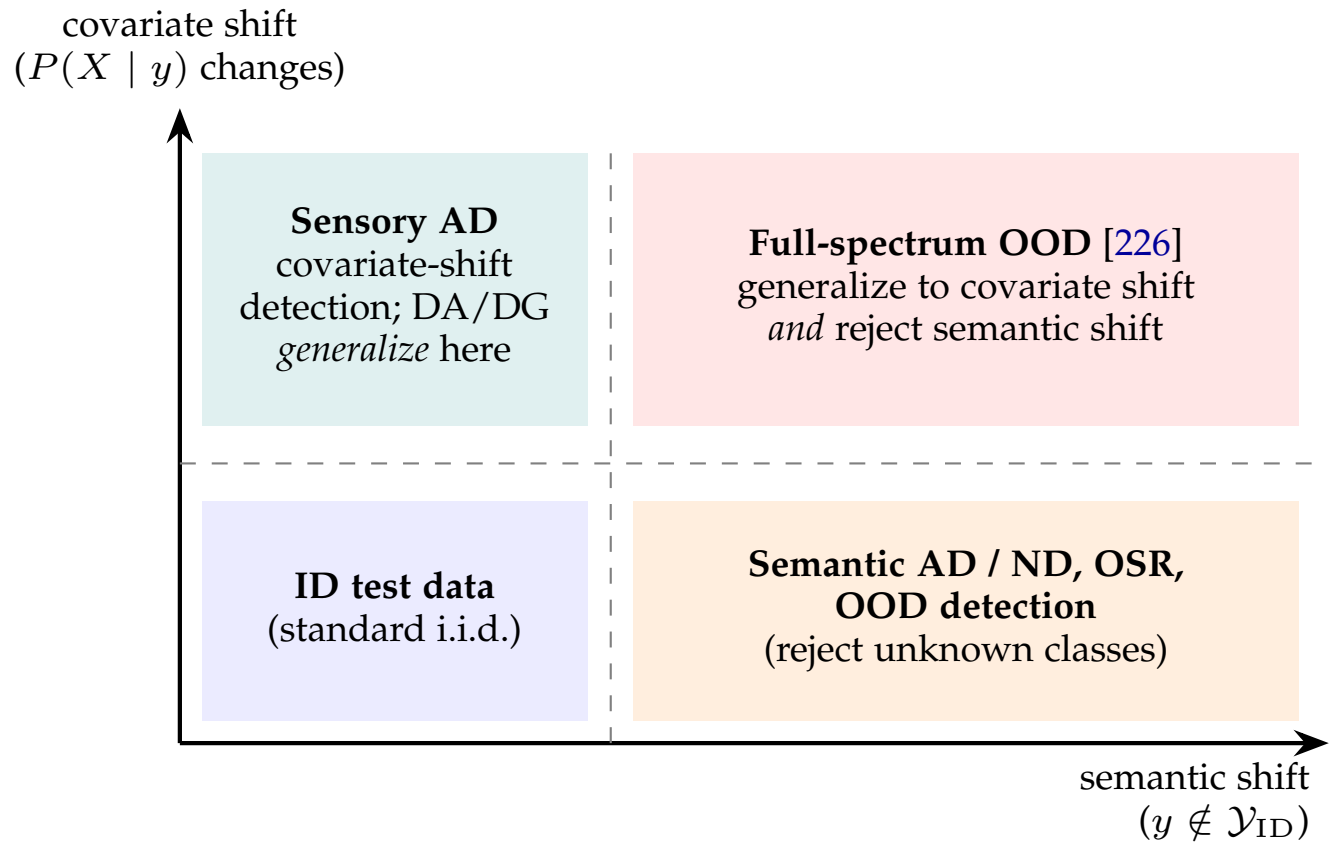
\begin{figure}[t]
\centering
\begin{tikzpicture}[font=\scriptsize, >=Stealth]
  \draw[->, thick] (0,0) -- (7.6,0) node[below left, align=right]{semantic shift\\($y\notin\Y_{\mathrm{ID}}$)};
  \draw[->, thick] (0,0) -- (0,4.2) node[above, align=center]{covariate shift\\($P(X\mid y)$ changes)};
  \fill[blue!8] (0.15,0.15) rectangle (2.7,1.6);
  \node[align=center] at (1.42,0.88) {\textbf{ID test data}\\(standard i.i.d.)};
  \fill[teal!12] (0.15,2.1) rectangle (2.7,3.9);
  \node[align=center] at (1.42,3.0) {\textbf{Sensory AD}\\covariate-shift\\detection; DA/DG\\\emph{generalize} here};
  \fill[orange!13] (3.0,0.15) rectangle (7.4,1.6);
  \node[align=center] at (5.2,0.88) {\textbf{Semantic AD / ND, OSR,}\\\textbf{OOD detection}\\(reject unknown classes)};
  \fill[red!10] (3.0,2.1) rectangle (7.4,3.9);
  \node[align=center] at (5.2,3.0) {\textbf{Full-spectrum OOD}~\cite{yang2023full}\\generalize to covariate shift\\\emph{and} reject semantic shift};
  \draw[dashed, gray] (2.85,0) -- (2.85,4.0);
  \draw[dashed, gray] (0,1.85) -- (7.5,1.85);
\end{tikzpicture}
\caption{Shift-type view of detection problems. The vertical axis is the
covariate axis studied by domain adaptation and generalization
(\S\ref{sec:highdim}); the horizontal axis is semantic (open-set) shift.
Detection methods tuned for one quadrant are not expected to transfer to
another (\S\ref{sec:ood-connection}).}
\label{fig:ood-plane}
\end{figure}

Under this framing (Fig.~\ref{fig:ood-plane}):
\emph{sensory AD} detects covariate shift only (e.g., a defective product
image of a known category); \emph{semantic AD / one-class ND} detects a new
class against a single ID class; \emph{multi-class ND} detects new classes
against several ID classes; \emph{OSR} additionally requires correct
classification among ID classes; \emph{OOD detection} is the broadest
semantic-shift setting, typically evaluated by a scalar score; and \emph{OD}
is transductive, operating on a single contaminated dataset without a
train/test split. Before this unification, papers using ``anomaly
detection'' and ``OOD detection'' frequently addressed the same setting under
different names, while other ``anomaly detection'' papers addressed what is
elsewhere called novelty detection.

\subsection{Connection to the Distribution-Shift Taxonomy}
\label{sec:ood-connection}

The mapping above is, in our reading, a restriction of the taxonomy of
\S\ref{sec:taxonomy} to a per-sample decision problem. Covariate shift in
the OOD sense is covariate shift as defined in \S\ref{sec:basic-types}; the
difference is that detection asks whether a \emph{particular} $x$ is unusual
under $\Ps(X)$, whereas adaptation asks how to perform well on \emph{average}
over $\Pt(X)$. Two corollaries follow. First, a score function implicitly
encodes which axis it targets: classifier-confidence scores are designed for
semantic shift and can be uninformative under covariate-only perturbations
that preserve class identity, so they should not be expected to work as
covariate-shift detectors, and vice versa. Second, there is a direct
\emph{tension} between generalization and detection: a model that
generalizes well to covariate shift (\S\ref{sec:highdim}) should \emph{not}
flag covariate-shifted ID inputs as OOD. This tension is operationalized by
full-spectrum OOD detection~\cite{yang2023full} and by benchmarks that
include covariate-shifted ID test sets~\cite{zhang2024openood}.

\subsection{OOD Detection Methods}
\label{sec:ood-methods}

Table~\ref{tab:ood} summarizes representative methods; we refer to
\cite{yang2024generalized,salehi2022unified} for exhaustive coverage.

\smallskip
\noindent\textbf{Post-hoc output scores.} The maximum softmax probability
(MSP)~\cite{hendrycks2017baseline} is the canonical baseline.
ODIN~\cite{liang2018enhancing} adds temperature scaling and small input
perturbations; the energy score~\cite{liu2020energy},
$E(x)=-T\log\sum_y e^{f_y(x)/T}$, is aligned with the input density up to a
constant and less prone to softmax overconfidence.

\smallskip
\noindent\textbf{Feature- and activation-based scores.} Mahalanobis
distance to class-conditional Gaussians~\cite{lee2018simple}, non-parametric
$k$-nearest-neighbor distance in normalized feature space~\cite{sun2022out},
and ViM~\cite{wang2022vim}, which combines a residual from the principal
feature subspace with logits, exploit information discarded by the softmax.
ReAct~\cite{sun2021react} truncates abnormally high activations and
ASH~\cite{djurisic2023extremely} prunes and rescales activations before
scoring; GradNorm~\cite{huang2021importance} uses the norm of gradients of
the KL divergence to the uniform distribution.

\smallskip
\noindent\textbf{Training-time methods.} Outlier Exposure~\cite{hendrycks2019deep}
trains on auxiliary outliers to produce uniform predictions; VOS~\cite{du2022vos}
synthesizes virtual outliers from low-likelihood regions of class-conditional
feature distributions; LogitNorm~\cite{wei2022mitigating} decouples logit
norm from the cross-entropy objective to reduce overconfidence; and
CIDER~\cite{ming2023exploit} learns hyperspherical embeddings with
intra-class compactness and inter-class dispersion.

\smallskip
\noindent\textbf{Open-set recognition.} Following the formalization of
Scheirer et al.~\cite{scheirer2013open}, OpenMax~\cite{bendale2016towards}
calibrates the rejection of unknowns with extreme value theory on distances
to class means. Vaze et al.~\cite{vaze2022open} show that OSR performance is
highly correlated with closed-set accuracy and that a well-trained closed-set
classifier with a simple max-logit score is competitive with dedicated OSR
methods.

\smallskip
\noindent\textbf{Anomaly detection without labels.} Classical density and
reconstruction approaches are reviewed by Pang et al.~\cite{pang2021deep} and
unified with shallow methods by Ruff et al.~\cite{ruff2021unifying}. In
industrial inspection (MVTec AD~\cite{bergmann2019mvtec}), memory banks of
pretrained patch features (PatchCore~\cite{roth2022towards}) score test patches
by nearest-neighbor distance.

\smallskip
\noindent\textbf{Theory and limits.} Fang et al.~\cite{fang2022out} study the
PAC learnability of OOD detection and show that it is not learnable in some
general settings, while giving conditions under which it is; Fort et
al.~\cite{fort2021exploring} show that large pretrained transformers sharply
improve near-OOD detection, especially when a few outlier examples are
available. Abstention more broadly is formalized by selective
classification~\cite{geifman2017selective}, which trades coverage for risk
and treats shifted inputs and hard ID inputs uniformly.

\begin{table*}[t]
\centering
\footnotesize
\setlength{\tabcolsep}{4pt}
\renewcommand{\arraystretch}{1.08}
\caption{Representative OOD detection methods. ``Retrain'': requires changing
the training of the classifier; ``Aux. OOD'': requires auxiliary outlier data;
``Zero-shot'': applicable without ID training images (VLM-based).}
\label{tab:ood}
\begin{tabular}{P{2.5cm}P{1.6cm}P{2.3cm}P{5.6cm}ccc}
\toprule
\textbf{Method} & \textbf{Venue} & \textbf{Family} & \textbf{Score / mechanism} & \textbf{Retrain} & \textbf{Aux. OOD} & \textbf{Zero-shot}\\
\midrule
MSP~\cite{hendrycks2017baseline} & ICLR'17 & output & max softmax probability & \xmark & \xmark & \xmark\\
ODIN~\cite{liang2018enhancing} & ICLR'18 & output & temperature + input perturbation & \xmark & \xmark & \xmark\\
Mahalanobis~\cite{lee2018simple} & NeurIPS'18 & feature & class-conditional Gaussian distance & \xmark & \xmark & \xmark\\
Outlier Exposure~\cite{hendrycks2019deep} & ICLR'19 & training & uniform targets on outliers & \cmark & \cmark & \xmark\\
Energy~\cite{liu2020energy} & NeurIPS'20 & output & free energy of logits & \xmark & \xmark & \xmark\\
ReAct~\cite{sun2021react} & NeurIPS'21 & activation & clipped activations + energy & \xmark & \xmark & \xmark\\
GradNorm~\cite{huang2021importance} & NeurIPS'21 & gradient & gradient norm of KL to uniform & \xmark & \xmark & \xmark\\
KNN~\cite{sun2022out} & ICML'22 & feature & $k$-NN distance, normalized features & \xmark & \xmark & \xmark\\
ViM~\cite{wang2022vim} & CVPR'22 & feature + output & principal-subspace residual + logits & \xmark & \xmark & \xmark\\
VOS~\cite{du2022vos} & ICLR'22 & training & virtual outlier synthesis & \cmark & \xmark & \xmark\\
LogitNorm~\cite{wei2022mitigating} & ICML'22 & training & logit normalization & \cmark & \xmark & \xmark\\
ASH~\cite{djurisic2023extremely} & ICLR'23 & activation & activation pruning and scaling & \xmark & \xmark & \xmark\\
CIDER~\cite{ming2023exploit} & ICLR'23 & training & hyperspherical prototypes & \cmark & \xmark & \xmark\\
MCM~\cite{ming2022delving} & NeurIPS'22 & VLM & softmax over image--text similarities & \xmark & \xmark & \cmark\\
CLIPN~\cite{wang2023clipn} & ICCV'23 & VLM & learned ``no'' prompts & \cmark & \xmark & \cmark\\
LoCoOp~\cite{miyai2023locoop} & NeurIPS'23 & VLM & few-shot prompts, local OOD regularization & \cmark & \xmark & \xmark\\
NegLabel~\cite{jiang2024negative} & ICLR'24 & VLM & mined negative labels from a lexicon & \xmark & \xmark & \cmark\\
\bottomrule
\end{tabular}
\end{table*}

\subsection{The VLM Era: Blurred Boundaries}
\label{sec:vlm-ood}

VLMs disrupt several implicit assumptions of the generalized framework, and
Miyai et al.~\cite{miyai2025generalized} revise it as \emph{generalized OOD
detection v2}. Three changes are most consequential.

\begin{enumerate}[leftmargin=*,itemsep=1pt]
\item \textbf{The ID set is defined by the prompt, not by training.} A frozen
CLIP-style model has no fixed label space; the ID classes are whatever the
practitioner names at inference. Zero-shot OOD detection therefore needs no
ID training images: MCM~\cite{ming2022delving} applies a softmax to
image--text cosine similarities over ID class names, and
NegLabel~\cite{jiang2024negative} mines a large set of \emph{negative} labels
from a lexicon and scores an image by its affinity to ID versus negative
labels. Few-shot prompt learning can also be regularized for OOD detection
(LoCoOp~\cite{miyai2023locoop} treats ID-irrelevant local regions as OOD
during prompt learning), and CLIPN~\cite{wang2023clipn} trains ``no'' prompts
so that the model can express negation.
\item \textbf{Anomaly detection becomes zero-shot.} WinCLIP~\cite{jeong2023winclip}
compares multi-scale windows with compositional ``normal''/``anomalous''
prompts, and AnomalyCLIP~\cite{zhou2024anomalyclip} learns object-agnostic
normality/abnormality prompts that transfer across object categories.
\item \textbf{Sub-fields consolidate.} Miyai et al.\ find that, in the VLM
era, OOD detection and sensory AD remain the most active and technically
challenging settings, while OSR has become largely subsumed by OOD detection
because VLM-based work uses identical benchmark setups for both. They also
extend the framework to large VLMs, e.g., detecting questions a model cannot
answer.
\end{enumerate}
Large language models extend this further: Xu and Ding~\cite{xu2025large}
categorize LLM-based anomaly and OOD detection by the role the LLM plays,
e.g., serving as the detector itself (via prompting or contrasting
representations) or supplying auxiliary information to another detector.

\smallskip
\noindent\textbf{Detection meets adaptation.} The two halves of the
detection--generalization tension now interact directly. Cao et
al.~\cite{cao2025noisy} define \emph{zero-shot noisy TTA}, in which a VLM is
adapted at test time to a stream containing samples outside the ID label
space, and show that existing TTA methods are harmed by such samples unless
they are detected and excluded online. This is a concrete instance of the
covariate/semantic distinction of Fig.~\ref{fig:ood-plane} governing which
samples should drive adaptation.

\smallskip
\noindent\textbf{Beyond single-object images (2026).} Most VLM detectors
assume one dominant object per image. UNI-OOD~\cite{li2026unood} performs
object- and image-level OOD detection within a single VLM without knowing
which task is posed, and UNO-Adapter~\cite{peng2026learning} addresses OOD object detection,
where no real unknown objects are available in training, by learning latent
concepts rather than synthesizing outliers. Both move the semantic-shift axis of Fig.~\ref{fig:ood-plane} from
the image to its parts.
\section{Distribution Shift in Foundation Models and World Models}
\label{sec:foundation}

Foundation models~\cite{bommasani2021opportunities}---and contrastive
vision--language models such as CLIP~\cite{radford2021learning},
ALIGN~\cite{jia2021scaling} and SigLIP~\cite{zhai2023sigmoid} in
particular---change the shift landscape in two ways that are in tension.
Large-scale pretraining appears to confer robustness that models trained on a
single curated dataset lack; yet the standard way of specializing these
models, fine-tuning, tends to erode it. This section reviews the evidence on
the origins of pretrained robustness (\S\ref{sec:pretraining-robustness}),
the fine-tuning trade-off and robust fine-tuning
(\S\ref{sec:finetuning-tradeoff}), parameter-efficient adaptation
(\S\ref{sec:prompt-tuning}), test-time and in-context adaptation
(\S\ref{sec:fm-tta}), multimodal robustness evaluation
(\S\ref{sec:vlm-benchmarks}), and language models
(\S\ref{sec:llm-robustness}). It then turns to world models
(\S\ref{sec:worldmodels}), where pretrained models are used not only to
predict but to act, and the deployment distribution becomes a function of
the model itself.

\subsection{Pretraining and Effective Robustness}
\label{sec:pretraining-robustness}

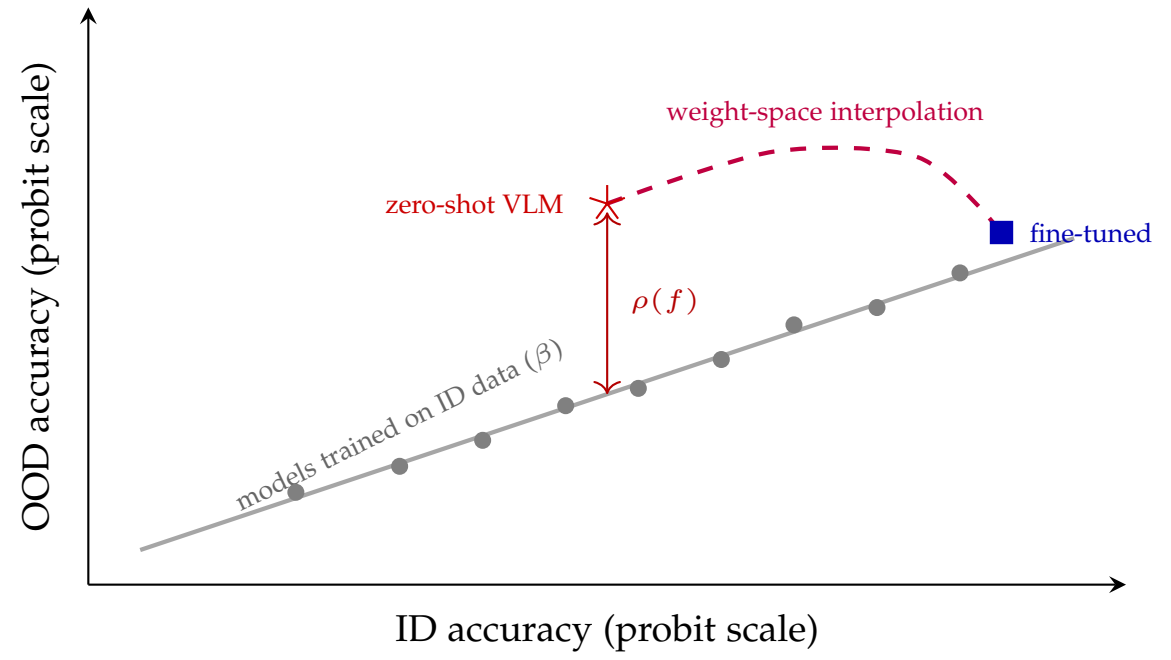
\begin{figure}[t]
\centering
\begin{tikzpicture}
\begin{axis}[
  width=0.95\columnwidth, height=5.4cm,
  xlabel={ID accuracy (probit scale)}, ylabel={OOD accuracy (probit scale)},
  xmin=0, xmax=10, ymin=0, ymax=10,
  xtick=\empty, ytick=\empty,
  axis lines=left, label style={font=\scriptsize},
  clip=false
]
  \addplot[gray!70, thick, domain=0.5:9.5] {0.6*x + 0.3};
  \addplot[only marks, mark=*, mark size=1.4pt, gray] coordinates
    {(2,1.6) (3,2.05) (3.8,2.5) (4.6,3.1) (5.3,3.4) (6.1,3.9) (6.8,4.5) (7.6,4.8) (8.4,5.4)};
  \node[font=\tiny, gray!80!black, rotate=26] at (axis cs:3.0,2.75) {models trained on ID data ($\beta$)};
  \addplot[only marks, mark=star, mark size=3.5pt, red!80!black] coordinates {(5.0,6.6)};
  \node[font=\tiny, red!80!black, anchor=east] at (axis cs:4.75,6.6) {zero-shot VLM};
  \addplot[only marks, mark=square*, mark size=2pt, blue!70!black] coordinates {(8.8,6.1)};
  \node[font=\tiny, blue!70!black, anchor=west] at (axis cs:8.9,6.1) {fine-tuned};
  \addplot[purple, thick, dashed, smooth] coordinates {(5.0,6.6) (6.6,7.5) (8.0,7.4) (8.8,6.1)};
  \node[font=\tiny, purple, anchor=south] at (axis cs:7.1,7.6) {weight-space interpolation};
  \draw[<->, red!70!black] (axis cs:5.0,3.3) -- (axis cs:5.0,6.45);
  \node[font=\tiny, red!70!black, anchor=west] at (axis cs:5.05,4.9) {$\rho(f)$};
\end{axis}
\end{tikzpicture}
\caption{Schematic (not data) of accuracy-on-the-line and effective
robustness~\cite{taori2020measuring,miller2021accuracy}. Models trained on
the ID distribution lie close to a line in probit space; zero-shot VLMs lie
above it. Standard fine-tuning raises ID accuracy but moves the model toward
the line~\cite{andreassen2021evolution}; interpolating the weights of the
zero-shot and fine-tuned models can trace a curve that dominates both
endpoints~\cite{wortsman2022robust}.}
\label{fig:effective-robustness}
\end{figure}

\noindent\textbf{Accuracy on the line.} Taori et al.~\cite{taori2020measuring}
evaluated over 200 ImageNet models on natural distribution shifts and
found that robustness interventions developed for synthetic shifts rarely
help on natural ones, and that OOD accuracy is largely predicted by ID
accuracy. Miller et al.~\cite{miller2021accuracy} show that, across many
architectures and training settings, ID and OOD accuracies are strongly
linearly correlated after a probit (or logit) transformation---\emph{accuracy
on the line}---for many but not all natural shifts. Taori et al.\ define the
\emph{effective robustness} of a model $f$ as its vertical distance above
this trend,
\begin{equation}
\rho(f)=\mathrm{acc}_{\mathrm{OOD}}(f)-\beta\big(\mathrm{acc}_{\mathrm{ID}}(f)\big),
\label{eq:effrob}
\end{equation}
where $\beta$ is the linear fit (in transformed space) obtained from a
reference population of models trained on the ID distribution. Models
trained on substantially more diverse data are the main exception: zero-shot
CLIP lies far above the ImageNet trend on several natural-shift
benchmarks~\cite{radford2021learning,taori2020measuring}
(Fig.~\ref{fig:effective-robustness}).

\noindent\textbf{Where does the robustness come from? A contested
question.} Several lines of evidence, not fully consistent with each other,
now bear on this question.
\begin{itemize}[leftmargin=*,itemsep=1pt]
\item \emph{Data distribution, not language supervision.} Fang et
al.~\cite{fang2022data} vary training data, objective, architecture and
language supervision in controlled experiments and conclude that the
\emph{training distribution} is the primary cause of CLIP's effective
robustness, with the contrastive objective and language supervision playing
little role.
\item \emph{Quality, not quantity, of data mixtures.} Nguyen et
al.~\cite{nguyen2022quality} find that combining pretraining sources does not
add robustness: the robustness of a mixture tends to be diluted toward that
of its less robust components.
\item \emph{Train--test similarity is insufficient\ldots} Mayilvahanan et
al.~\cite{mayilvahanan2024does} retrain CLIP on LAION subsets pruned to match
ImageNet's train--test similarity to common OOD benchmarks; performance drops
on some benchmarks but remains high overall, indicating that near-duplicates
of test data do not by themselves explain CLIP's OOD performance.
\item \emph{\ldots but domain contamination matters.} In follow-up work, the
same group constructs style-filtered subsets (LAION-Natural,
LAION-Rendition) that are strictly OOD in style with respect to ImageNet and
DomainNet test sets, and finds that a significant portion of CLIP's
rendition-domain performance is explained by in-domain training
examples~\cite{mayilvahanan2025search}.
\item \emph{Concept frequency scaling.} Udandarao et al.~\cite{udandarao2024no}
analyze 34 models and five pretraining datasets and find a log-linear
relation between a concept's pretraining frequency and zero-shot
performance on it: linear gains require exponentially more data, even after
controlling for sample-level similarity.
\item \emph{Measurement artifacts.} Shi et al.~\cite{shi2023effective} argue
that effective robustness measured against a single ID test set (ImageNet)
favors models whose training data differ from it; controlling for accuracy on
multiple ID test sets that cover each model's training distribution
substantially reduces the measured effective-robustness advantage of
zero-shot CLIP-like models.
\end{itemize}
Taken together, these results suggest that a substantial part of what has
been reported as distributional robustness of foundation models reflects
\emph{coverage} of test-like domains and concepts by web-scale data, and that
the choice of ``ID'' reference is itself a modeling decision. Classical
notions---support overlap (\S\ref{sec:classical-correction}) and diversity
versus correlation shift~\cite{ye2022ood}---are the right vocabulary for this
debate, which remains open (\S\ref{sec:challenges}).

\noindent\textbf{Architecture.} Vision Transformers pretrained on sufficiently
large data have been reported to be at least as robust as comparable ResNets
to a broad range of perturbations~\cite{bhojanapalli2021understanding}, and online TTA benchmarks find
transformers more resilient to domain shift~\cite{wang2025search}; however,
the accuracy-on-the-line analyses indicate that much of the apparent
architectural benefit follows ID accuracy~\cite{miller2021accuracy}.

\subsection{The Fine-Tuning Robustness Trade-off}
\label{sec:finetuning-tradeoff}

Andreassen et al.~\cite{andreassen2021evolution} track effective robustness
during fine-tuning and find that pretrained models start with high
effective robustness that diminishes as fine-tuning proceeds to convergence.
Kumar et al.~\cite{kumar2022finetuning} explain one mechanism: under a
linear analysis, full fine-tuning with a randomly initialized head changes
features in directions spanned by the ID data while leaving others
unchanged, \emph{distorting} the pretrained features and underperforming
linear probing OOD; initializing with a linear probe before full fine-tuning
(\emph{LP-FT}) mitigates this. They report LP-FT improving OOD accuracy over
both linear probing and full fine-tuning across several shifts.

\smallskip
\noindent\textbf{Robust fine-tuning methods.} WiSE-FT~\cite{wortsman2022robust}
linearly interpolates zero-shot and fine-tuned weights,
$\theta_\alpha=(1-\alpha)\theta_{\mathrm{zs}}+\alpha\theta_{\mathrm{ft}}$,
improving OOD accuracy substantially while preserving or improving ID
accuracy; this exploits linear mode connectivity between a pretrained model
and its fine-tuned descendant. Model soups~\cite{wortsman2022model} average
multiple fine-tuned models with different hyperparameters.
FLYP~\cite{goyal2023finetune} fine-tunes with the same contrastive objective
used in pretraining, treating class names as text prompts.
TPGM~\cite{tian2023trainable} learns per-layer projection radii that
constrain how far fine-tuned weights move from pretrained ones.
Lipsum-FT~\cite{nam2024lipsum} regularizes the change in image--text
energy measured on random text, preserving vision--language alignment.
StarFT~\cite{kim2025starft} targets spurious features: it uses a language model
to generate captions describing potentially spurious attributes and aligns
the fine-tuned model's output distribution on them with that of the zero-shot
model, reporting improvements of 14.30 points in worst-group and 3.02 points
in average accuracy in the Waterbirds group-shift setting. Surgical
fine-tuning~\cite{lee2023surgical} (\S\ref{sec:surgical}) and
mutual-information regularization toward pretrained
features~\cite{cha2022domain} pursue the same goal by restricting what
fine-tuning may change.

\subsection{Prompt and Adapter-Based Adaptation}
\label{sec:prompt-tuning}

Parameter-efficient methods adapt few parameters and leave the backbone
frozen. CoOp~\cite{zhou2022learning} learns continuous context vectors in
place of hand-crafted prompts; it improves accuracy on base (seen) classes
but generalizes worse to unseen classes within the same dataset.
CoCoOp~\cite{zhou2022conditional} conditions the prompt on each image via a
light meta-network, narrowing this base--novel gap.
MaPLe~\cite{khattak2023maple} learns coupled prompts in both the vision and
language branches; KgCoOp~\cite{yao2023visual} and
PromptSRC~\cite{khattak2023self} regularize learned prompts toward the
hand-crafted zero-shot ones to prevent forgetting of general knowledge.
Adapter-based methods learn a light residual on CLIP features
(CLIP-Adapter~\cite{gao2024clip}) or construct a training-free key--value
cache from few-shot features (Tip-Adapter~\cite{zhang2022tip}); low-rank
adaptation~\cite{hu2022lora} is the dominant parameter-efficient method for
large language models. Because these methods leave the backbone untouched,
they inherit more of its zero-shot behavior than full fine-tuning, but the
recurring base-to-novel trade-off shows that even a few parameters can
overfit to the adaptation distribution---the same robustness--plasticity
tension as in \S\ref{sec:finetuning-tradeoff}.

\subsection{Test-Time and In-Context Adaptation}
\label{sec:fm-tta}

\noindent\textbf{Test-time adaptation of VLMs.} Test-time prompt tuning
(TPT)~\cite{shu2022test} optimizes a prompt per test image by minimizing the
entropy of the averaged prediction over confident augmented views;
DiffTPT~\cite{feng2023diverse} uses diffusion-generated augmentations.
C-TPT~\cite{yoon2024ctpt} observes that TPT worsens calibration and that
prompts producing more dispersed text features are better calibrated, and
regularizes for this dispersion. A second line avoids backpropagation
altogether: TDA~\cite{karmanov2024efficient} maintains positive and negative
key--value caches of confident test features and pseudo-labels;
MTA~\cite{zanella2024test} replaces prompt optimization by a robust
mode-seeking procedure over augmented views; DPE~\cite{zhang2024dual} evolves
textual and visual prototypes over the test stream;
BoostAdapter~\cite{zhang2024boostadapter} combines historical samples with
regional bootstraps of the current sample; and
WATT~\cite{osowiechi2024watt} averages weights adapted under multiple text
templates. Zanella et al.~\cite{zanella2025realistic} show that many of these
methods implicitly assume that every class appears in each batch and that
batches are i.i.d., and propose StatA for realistic streams with a variable
number of effective classes; Bayesian class adaptation~\cite{zhou2025bayesian}
adapts the class \emph{prior} as well as class embeddings---a test-time
analogue of label-shift correction (\S\ref{sec:bbse}); and zero-shot noisy
TTA~\cite{cao2025noisy} handles streams containing out-of-label-space
samples (\S\ref{sec:vlm-ood}). Table~\ref{tab:vlm-adapt} summarizes
adaptation strategies for VLMs by data access.

\begin{table*}[t]
\centering
\footnotesize
\setlength{\tabcolsep}{4pt}
\renewcommand{\arraystretch}{1.08}
\caption{Adaptation strategies for vision--language models, organized by
the data they use and what they modify.}
\label{tab:vlm-adapt}
\begin{tabular}{P{2.0cm}P{3.6cm}P{2.7cm}P{2.9cm}P{4.6cm}}
\toprule
\textbf{Strategy} & \textbf{Representative methods} & \textbf{Target data} & \textbf{Modified} & \textbf{Robustness consideration}\\
\midrule
Robust full FT & WiSE-FT~\cite{wortsman2022robust}, LP-FT~\cite{kumar2022finetuning}, FLYP~\cite{goyal2023finetune}, TPGM~\cite{tian2023trainable}, Lipsum-FT~\cite{nam2024lipsum}, StarFT~\cite{kim2025starft} & labeled ID data & all weights (constrained) & preserves effective robustness of the zero-shot model\\
Prompt learning & CoOp~\cite{zhou2022learning}, CoCoOp~\cite{zhou2022conditional}, MaPLe~\cite{khattak2023maple}, KgCoOp~\cite{yao2023visual}, PromptSRC~\cite{khattak2023self} & few-shot labeled & prompt tokens & base-to-novel generalization trade-off\\
Adapters & CLIP-Adapter~\cite{gao2024clip}, Tip-Adapter~\cite{zhang2022tip} & few-shot labeled & residual / cache & small parameter budget limits drift\\
Test-time prompt & TPT~\cite{shu2022test}, DiffTPT~\cite{feng2023diverse}, C-TPT~\cite{yoon2024ctpt} & single test image & prompt (per sample) & calibration can degrade (C-TPT)\\
Training-free TTA & TDA~\cite{karmanov2024efficient}, MTA~\cite{zanella2024test}, DPE~\cite{zhang2024dual}, BoostAdapter~\cite{zhang2024boostadapter}, BCA~\cite{zhou2025bayesian} & unlabeled stream & cache / prototypes / prior & assumptions on class presence and i.i.d.\ batches~\cite{zanella2025realistic}\\
Open-world TTA & ZS-NTTA~\cite{cao2025noisy}, StatA~\cite{zanella2025realistic} & stream with noise / variable classes & detector + adapter & integrates OOD detection with TTA\\
\bottomrule
\end{tabular}
\end{table*}

\noindent\textbf{In-context adaptation.} Foundation models offer an
adaptation modality unavailable to conventional classifiers: conditioning a
frozen model on a context of examples without gradient
updates~\cite{brown2020language}. For tabular data this is the operating
principle of TabPFN~\cite{hollmann2025accurate}, and Drift-Resilient
TabPFN~\cite{helli2024driftresilient} uses it to extrapolate temporal shift
(\S\ref{sec:tabular-remediation}). For language, Calderon et
al.~\cite{calderon2024measuring} measure over 14{,}000 domain shifts across
fine-tuned models and few-shot LLMs and find that fine-tuned models excel
in-domain while few-shot LLMs often surpass them cross-domain, showing
smaller drops. The training-free caches of TDA and Tip-Adapter are a
non-parametric analogue for VLMs.

\smallskip
\noindent\textbf{Long-tailed test streams.} Training-free and prompt-based
VLM TTA typically assumes balanced test classes. Wang et
al.~\cite{wang2026long} show that long-tailed streams bias entropy
minimization towards head classes and propose L-TTA, combining tail-enriching
prototypes, rebalancing cross-attention over shared hyper-class vectors, and
a balanced entropy objective, evaluated on 15 datasets at imbalance ratios up
to 50---a joint label-shift and covariate-shift setting in the terms of
\S\ref{sec:taxonomy}.

\subsection{Evaluating Multimodal Robustness}
\label{sec:vlm-benchmarks}

Because a VLM's failure can originate in either modality or in their
alignment, single-modality corruption benchmarks are insufficient. Qiu et
al.~\cite{qiu2024benchmarking} apply 17 image and 16 text perturbation
methods to five tasks (retrieval, visual reasoning, visual entailment,
captioning and text-to-image generation) and find that multimodal models are
not robust to either, and especially not to image perturbations;
character-level text perturbations and zoom blur were the most severe shifts
in their study. Koddenbrock et al.~\cite{koddenbrock2025domain} propose
DeepBench, which uses an LLM to generate domain-specific, label-free image
corruptions for a described deployment domain, and observe substantial
variability in robustness across six real-world domains and contrastive VLM
variants. Robustness-targeted augmentation during vision--language
pretraining, such as RobustMixGen~\cite{kim2025robustmixgen}, addresses the
same issue at training time. Together with the coverage debate of
\S\ref{sec:pretraining-robustness}, these results indicate that ``VLMs are
robust'' is a graded, domain-dependent claim rather than a guarantee.

\subsection{Language Models}
\label{sec:llm-robustness}

Pretrained transformers show markedly smaller OOD performance drops than
bag-of-words, convolutional or LSTM models and are better at detecting
anomalous inputs, although larger size alone does not guarantee greater
robustness and distillation can reduce it~\cite{hendrycks2020pretrained}.
Benchmarks such as GLUE-X~\cite{yang2023gluex} and BOSS~\cite{yuan2023revisiting}
evaluate OOD generalization of pretrained language models under domain and
style shifts, the latter explicitly including LLMs, and the domain-robustness study of
Calderon et al.~\cite{calderon2024measuring} introduces the complementary
\emph{Source Drop} and \emph{Target Drop} metrics (\S\ref{sec:metrics-performance}).
LLMs exhibit a distinctive form of input-distribution sensitivity: Sclar et
al.~\cite{sclar2024quantifying} report that semantically equivalent prompt
formats can change few-shot accuracy by large margins across several
open-source LLMs. Zhang et al.~\cite{zhang2025evaluating} survey LLM
robustness along adversarial robustness, OOD robustness (including OOD
detection, zero-shot transfer and hallucination) and evaluation. Natural
shift after deployment is now measured directly: Seegmiller and
Preum~\cite{seegmiller2026measuring} quantify shift in real user prompts
across 192 settings over time, user groups and geography, and find that
moderate prompt shift corresponds to large drops in instruction-following
performance, arguing for continuous monitoring of deployed LLMs---the
language-model counterpart of the drift detection of
\S\ref{sec:driftdetection}. Wang et al.~\cite{wang2026agentic} argue, in a
position paper, that the OOD problems of foundation models (knowledge
boundaries, capability ceilings, compositional and open-ended task shift)
call for agentic systems that use tools and interaction rather than static
models alone, a view that connects to the closed-loop setting of
\S\ref{sec:worldmodels}.

\begin{remark}[What is new versus relabeled]
Several phenomena in this section have no precise classical analogue: the
accuracy-on-the-line regularity and its violation by models trained on
different data; the fine-tuning robustness trade-off and its repair by
weight-space interpolation; and in-context adaptation. Others are
instantiations of established principles: robust fine-tuning regularizers are
proximal or invariance constraints (\S\ref{sec:dg}); training-free VLM TTA
methods are prototype or cache-based classifier adjustment
(\S\ref{sec:tta}); prior-adaptive VLM TTA is label-shift correction
(\S\ref{sec:bbse}); and the coverage debate is a question about support
overlap and diversity shift (\S\ref{sec:taxonomy}). Distinguishing the two
categories matters for assessing novelty claims in a fast-moving area.
\end{remark}
\subsection{World Models: Distribution Shift in Closed Loop}
\label{sec:worldmodels}

The preceding sections treat a predictor $h:\X\to\Y$ that is evaluated on
inputs it does not influence. A \emph{world model} breaks this assumption. It
is a learned conditional model of how an environment evolves under
actions---classically a transition model in model-based reinforcement
learning (RL)~\cite{ha2018recurrent,hafner2020dream,chua2018deep}, and more
recently a latent predictive architecture~\cite{lecun2022path,assran2025vjepa2,zhou2025dinowm,maes2026leworldmodel}
or an action-controllable video generator~\cite{bruce2024genie,nvidia2025cosmos,parkerholder2025genie3}---that
an agent queries to imagine the consequences of candidate actions before
committing to one. Because the agent's own choices determine which
state--action pairs the model is asked about, distribution shift in world
models is \emph{endogenous}: the deployed system generates its own test
distribution. This subsection formalizes that observation, maps it onto the
taxonomy of \S\ref{sec:taxonomy}, and organizes the rapidly growing 2024--2026
literature on how world models fail and adapt under shift. We restrict
attention to the shift-specific aspects; comprehensive surveys of world
models in embodied AI~\cite{li2025comprehensivewm} and of distribution shift
in offline RL~\cite{levine2020offline,samani2026distribution} cover the
broader landscape.

\subsubsection{Setup: What Is Being Shifted}
\label{sec:wm-setup}

Let $M=(\mathcal{S},\mathcal{A},P,r,\gamma,\rho_0)$ be a discounted Markov
decision process with transition kernel $P(s'\mid s,a)$, reward
$r(s,a)\in[0,R_{\max}]$, discount $\gamma\in[0,1)$ and initial distribution
$\rho_0$. For pixel observations, an encoder $z=g_\phi(o)$ maps an
observation to a latent state and a predictor
$\hat z_{t+1}=f_\theta(z_t,a_t)$ is trained on logged transitions; we write
$\hat P$ for the induced model kernel and $\hat M$ for the model MDP that
replaces $P$ by $\hat P$. A world model is fit by supervised (or
self-supervised) learning on a dataset
$\D=\{(s_i,a_i,s'_i)\}$ collected by a behavior policy $\pi_\beta$, so its
training ``inputs'' $(s,a)$ follow the data occupancy $d_\D(s,a)$ and its
``labels'' $s'$ follow $P(\cdot\mid s,a)$. For a policy $\pi$ and an MDP
$M$, let $J_M(\pi)=\E[\sum_t\gamma^t r(s_t,a_t)]$ and let
$d^\pi_M(s,a)=(1-\gamma)\sum_{t\ge0}\gamma^t\Pr^{\pi}_{M}(s_t=s,a_t=a)$ be
the normalized discounted occupancy. Viewed as a regression problem from
$X=(s,a)$ to $Y=s'$, a world model is subject to every shift of
Definition~\ref{def:shift}; what is new is \emph{who} moves the input
distribution.

\subsubsection{Endogenous Covariate Shift and Compounding Error}
\label{sec:wm-theory}

The following statement makes precise how model error measured on the
training distribution translates into decision error under the
distribution the agent itself induces.

\begin{proposition}[Model-based return under policy-induced shift]
\label{prop:wm}
Let $V^\pi_M$ be the value function of $\pi$ in $M$, and let
$\mathrm{TV}(s,a)=\mathrm{TV}\big(\hat P(\cdot\mid s,a),P(\cdot\mid s,a)\big)$.
Then
\begin{enumerate}[leftmargin=*,label=(\roman*),itemsep=1pt]
\item (telescoping)
$J_{\hat M}(\pi)-J_M(\pi)=\frac{\gamma}{1-\gamma}\,
\E_{(s,a)\sim d^\pi_{\hat M}}\!\big[\E_{\hat P(\cdot\mid s,a)}V^\pi_M-\E_{P(\cdot\mid s,a)}V^\pi_M\big]$;
\item (horizon amplification)
$|J_{\hat M}(\pi)-J_M(\pi)|\le\frac{\gamma R_{\max}}{(1-\gamma)^2}\,
\E_{d^\pi_{\hat M}}[\mathrm{TV}(s,a)]$;
\item (density-ratio form) if $d^\pi_{\hat M}\ll d_\D$ with
$w=d^\pi_{\hat M}/d_\D$, then
$|J_{\hat M}(\pi)-J_M(\pi)|\le\frac{\gamma R_{\max}}{(1-\gamma)^2}\,
\E_{d_\D}[w\,\mathrm{TV}]\le\frac{\gamma R_{\max}}{(1-\gamma)^2}\,
\|w\|_\infty\,\varepsilon_\D$, where
$\varepsilon_\D=\E_{d_\D}[\mathrm{TV}]$ is the in-distribution model error;
\item (pessimism) if $u(s,a)\ge\mathrm{TV}(s,a)$ for all $(s,a)$ and
$\hat M_u$ denotes $\hat M$ with reward $r-\lambda u$,
$\lambda=\gamma R_{\max}/(1-\gamma)$, then $J_M(\pi)\ge J_{\hat M_u}(\pi)$.
\end{enumerate}
\end{proposition}

\noindent\emph{Proof sketch.} Let $W_j$ be the expected return of $\pi$ when
the first $j$ transitions are drawn from $\hat P$ and the remainder from $P$.
Then $W_0=J_M$, $W_j\to J_{\hat M}$, and $W_{j+1}-W_j=\gamma^{j+1}\E_{(s_j,a_j)}
[\E_{\hat P}V^\pi_M-\E_PV^\pi_M]$ with $(s_j,a_j)$ distributed as in
$\hat M$; summing gives (i). Since $V^\pi_M\in[0,R_{\max}/(1-\gamma)]$, the
inner difference is at most $R_{\max}\mathrm{TV}/(1-\gamma)$, giving (ii);
(iii) is a change of measure, and (iv) follows from (ii) because
$J_{\hat M_u}(\pi)=J_{\hat M}(\pi)-\frac{\lambda}{1-\gamma}\E_{d^\pi_{\hat M}}[u]$.
\hfill$\square$

Part (i) is the standard simulation (telescoping) argument; bounds of the
form (ii) underlie model-based policy optimization~\cite{janner2019trust}, and
(iv) is, with its constant specialized to total variation, the
uncertainty-penalized MDP of MOPO~\cite{yu2020mopo}. We state them together
because, read through the lens of this survey, they say three things.

\smallskip
\noindent\textbf{(1) Planning is covariate-shift correction with an
adversarial weight.} The ratio $w$ in (iii) is an importance weight between
the occupancy the planner induces \emph{inside the model} and the data
occupancy, exactly as in \eqref{eq:iw}; a guarantee requires support overlap
and is governed by $\|w\|_\infty$, as in classical covariate-shift
analysis~\cite{cortes2010learning}. Unlike the static case, $w$ is not
fixed by nature: an optimizer that maximizes $J_{\hat M}$ is rewarded for
steering towards state--action regions where $\hat P$ is optimistically
wrong, i.e., it \emph{selects} the shift that maximizes model error. This is
why offline model-based RL treats pessimism as essential---MOPO subtracts an
uncertainty penalty as in (iv), and MOReL~\cite{kidambi2020morel} constructs
a pessimistic MDP that terminates in unknown regions---and why imitation
learning, which suffers the same endogenous shift between expert and learner
state distributions, is corrected by on-policy data
aggregation~\cite{ross2011reduction}.

\smallskip
\noindent\textbf{(2) Compounding error is a horizon-dependent shift
amplifier.} The $(1-\gamma)^{-2}$ factor reflects that model errors are fed
back as inputs: during a rollout the model is evaluated at predicted states
$\hat s_t$ rather than at the true states on which it was trained, a
self-induced covariate shift of the model's own input. Short model
rollouts branched from real states~\cite{janner2019trust}, ensembles of
probabilistic models~\cite{chua2018deep} and frequent replanning in model
predictive control (MPC) all act on this factor.

\smallskip
\noindent\textbf{(3) Interaction changes what is identifiable.}
Proposition~\ref{prop:nonid} states that a change in $P(Y\mid X)$ cannot be
recovered from unlabeled target inputs. For a world model, the environment
reveals the ``label'' $s'$ of every executed $(s,a)$ one step later, at no
annotation cost. A change of dynamics---concept shift in
$P(s'\mid s,a)$---is therefore identifiable at deployment, but only on the
support of the deployment occupancy; outside it, Proposition~\ref{prop:nonid}
applies unchanged. Test-time adaptation of world models thus has a
genuinely supervised signal, unlike the entropy-type surrogates of
classifier TTA (\S\ref{sec:tta}), which explains why it is effective with
very few updates (\S\ref{sec:wm-adapt}). One caveat applies to latent world
models: when the prediction target is itself an encoder output
$g_\phi(o_{t+1})$, an observation shift moves the target as well as the
input, so the self-supervised signal is not a fixed ground truth.

\subsubsection{A Taxonomy of Shift for World Models}
\label{sec:wm-taxonomy}

\begin{figure}[t]
\centering
\resizebox{\columnwidth}{!}{%
\begin{tikzpicture}[
  font=\small, >=Stealth, line width=0.55pt,
  box/.style={draw, rounded corners=3pt, align=center, inner sep=4pt, minimum height=1.0cm, text width=2.6cm},
  tag/.style={circle, draw=red!70!black, fill=red!8, inner sep=0.8pt, font=\scriptsize\bfseries, text=red!70!black}
]
\node[box, fill=gray!12, draw=gray!60] (env) at (0,0) {Environment\\[-1pt]{\scriptsize $P(s'\!\mid s,a)$, $P(o\!\mid s)$}};
\node[box, fill=blue!7, draw=blue!40] (data) at (6,0) {Logged data $\D$\\[-1pt]{\scriptsize $(s,a)\sim d_\D$ from $\pi_\beta$}};
\node[box, fill=teal!8, draw=teal!45] (wm) at (6,-3) {World model\\[-1pt]{\scriptsize $g_\phi$, $f_\theta$ $\Rightarrow\hat P$}};
\node[box, fill=orange!9, draw=orange!50] (plan) at (0,-3) {Planner / policy\\[-1pt]{\scriptsize MPC, or $\pi$ trained in $\hat M$}};
\draw[->] (env) -- node[above, font=\scriptsize]{collect} (data);
\draw[->] (data) -- node[right, font=\scriptsize]{fit} (wm);
\draw[->] ([yshift=6pt]wm.west) -- node[above, font=\scriptsize]{imagined rollouts $\hat s_{t+1}$} ([yshift=6pt]plan.east);
\draw[->] ([yshift=-6pt]plan.east) -- node[below, font=\scriptsize]{queries $(\hat s_t,a_t)\sim d^\pi$} ([yshift=-6pt]wm.west);
\draw[->] (plan) -- node[left, font=\scriptsize, align=right]{act $a_t$} (env);
\draw[->, dashed, teal!60!black] ([xshift=-4pt]env.south east) -- node[pos=0.5, below, sloped, font=\scriptsize]{observed $(s,a,s')$} ([xshift=4pt]wm.north west);
\node[tag] at ([xshift=-2pt,yshift=2pt]env.north west) {1};
\node[tag] at ([xshift=-2pt,yshift=-2pt]env.south west) {2};
\node[tag] at (1.9,-2.1) {3};
\node[tag] at (1.9,-3.9) {4};
\node[tag] at ([xshift=2pt,yshift=2pt]wm.north east) {5};
\end{tikzpicture}}
\caption{Where shift enters a world-model-based agent. \textbf{1}~dynamics
shift and \textbf{2}~observation shift in the deployment environment;
\textbf{3}~compounding error when imagined states are fed back as inputs;
\textbf{4}~policy-induced occupancy shift, since the planner queries the model
at $(\hat s_t,a_t)\sim d^\pi$ rather than $d_\D$; \textbf{5}~compositional or
extrapolative physical shift relative to the training data. The dashed arrow
is the self-supervised signal used for test-time adaptation, which makes
dynamics shift identifiable at test time (\S\ref{sec:wm-theory}).}
\label{fig:wm-loop}
\end{figure}
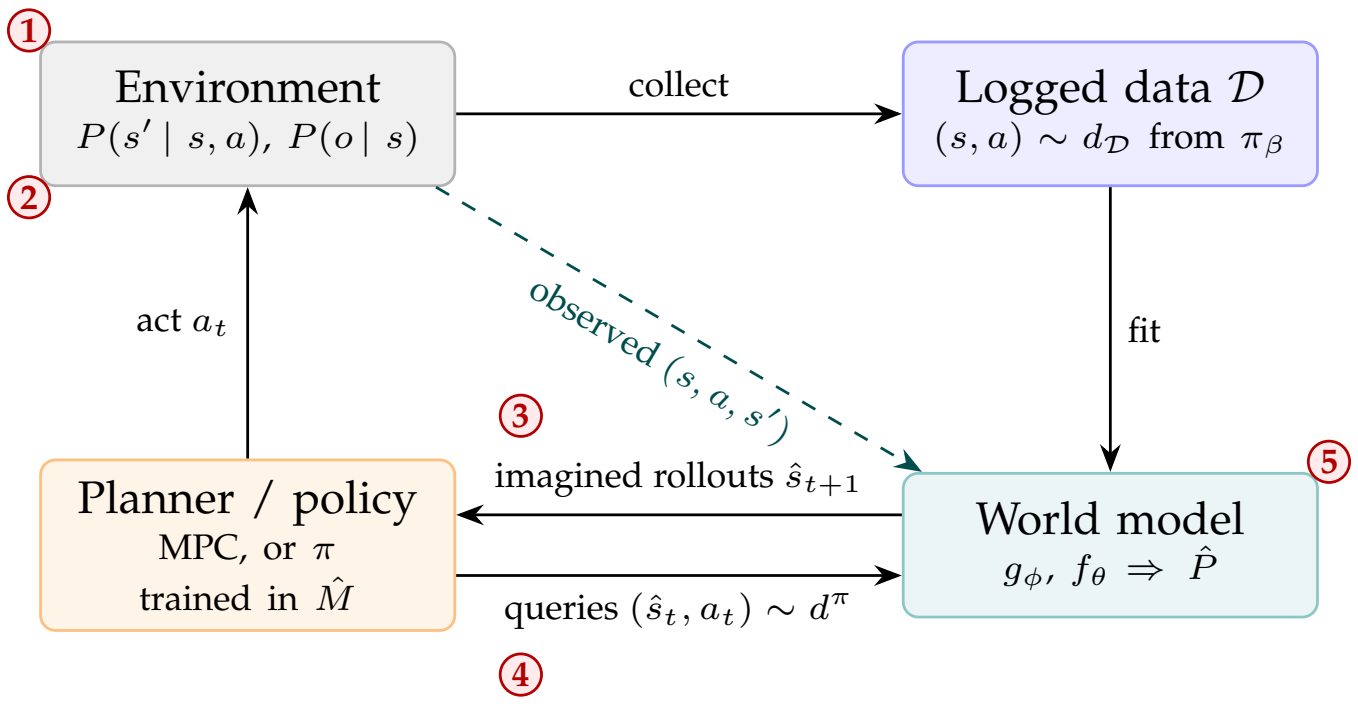

Table~\ref{tab:wm-shift} maps the shifts encountered by world models onto
Table~\ref{tab:shift-types}. Fig.~\ref{fig:wm-loop} shows where each enters
the closed loop.

\begin{table*}[t]
\centering
\footnotesize
\setlength{\tabcolsep}{3pt}
\renewcommand{\arraystretch}{1.12}
\caption{Distribution shift in world models, mapped to the taxonomy of
Table~\ref{tab:shift-types}. $d_\D$: data occupancy of the behavior policy;
$d^\pi$: occupancy induced by the deployed policy or planner; $P(o\mid s)$:
observation (rendering) process; $P(s'\mid s,a)$: dynamics.}
\label{tab:wm-shift}
\begin{tabular}{P{2.5cm}P{3.6cm}P{2.6cm}P{3.6cm}P{4.6cm}}
\toprule
\textbf{World-model shift} & \textbf{What changes} & \textbf{Analogue in Table~\ref{tab:shift-types}} & \textbf{Identifiability / evidence} & \textbf{Representative remedies}\\
\midrule
Policy-induced occupancy shift & $d^\pi(s,a)\neq d_\D(s,a)$; $P(s'\mid s,a)$ fixed & Covariate shift on $(s,a)$, chosen by the learner & Guarantee needs support overlap; $\|w\|_\infty$ in Prop.~\ref{prop:wm}(iii) & Uncertainty penalty~\cite{yu2020mopo}, pessimistic MDP~\cite{kidambi2020morel}, short branched rollouts~\cite{janner2019trust}, data aggregation~\cite{ross2011reduction}\\
Compounding rollout error & Model evaluated at its own predictions $\hat s_t$ & Self-induced covariate shift & Horizon factor $(1-\gamma)^{-2}$ in Prop.~\ref{prop:wm}(ii) & Probabilistic ensembles~\cite{chua2018deep}, MPC replanning, latent-space planning~\cite{zhou2025dinowm,terver2025what}\\
Observation (appearance) shift & $P(o\mid s)$: background, lighting, color, distractors; dynamics fixed & Conditional shift; GLS on the encoder (\S\ref{sec:basic-types}) & Planning collapses under visual factors~\cite{maes2026stable}; DINO-WM sensitive to task-irrelevant features~\cite{toso2026learning} & Domain randomization~\cite{tobin2017domain}, bisimulation-invariant encoders~\cite{toso2026learning}, encoder/predictor TTA~\cite{wang2026adajepa,soni2026sandwich}\\
Dynamics shift & $P_T(s'\mid s,a)\neq P_S(s'\mid s,a)$: mass, friction, damping, actuation & Concept shift with $X=(s,a)$ & Identifiable on visited support from interaction (\S\ref{sec:wm-theory}) & Online self-supervised TTA~\cite{wang2026adajepa,soni2026sandwich,soni2026alra,wan2026worldagen}, mixtures of world models~\cite{yoo2025world,jang2026test}\\
Compositional / extrapolative physical shift & Unseen combinations or out-of-range values of object attributes and interaction regimes & Sub-population / support mismatch & ID $\gg$ combinatorial $\gg$ OOD~\cite{kang2025far}; regime- and complexity-dependent~\cite{xue2026acwm} & Data diversity and scale help combinatorially, not OOD~\cite{kang2025far}; no general remedy\\
Novel hazards & Failure modes absent from training data & Semantic (open-set) shift (\S\ref{sec:ood}) & Optimistic safety estimates in OOD imagination~\cite{seo2025uncertainty} & Epistemic uncertainty with conformal threshold~\cite{seo2025uncertainty}; system-level monitoring~\cite{sinha2022system}\\
Task / reward shift & $r$ or goal changes; dynamics fixed & Task shift (\S\ref{sec:relation-transfer}) & Not a shift of the world model itself & Reuse the model for new goals; reward-free JEPA world models~\cite{zhou2025dinowm,maes2026leworldmodel}\\
\bottomrule
\end{tabular}
\end{table*}

\subsubsection{Foundation World Models and Their Generalization}
\label{sec:wm-foundation}

World models have followed the trajectory of \S\ref{sec:foundation} from
task-specific to pretrained-then-adapted. DreamerV3 learns a latent world
model and policy in imagination and, with a single hyperparameter
configuration, outperforms specialized methods across over 150
tasks~\cite{hafner2025mastering}, removing per-domain tuning as a confound
when comparing domains. Genie~\cite{bruce2024genie} is an 11B-parameter
action-controllable world model learned from unlabeled Internet videos via a
latent action model; Genie~3 generates interactive environments in real time
at 720p and 24 frames per second that remain consistent for a few minutes,
and supports text-prompted changes to the world~\cite{parkerholder2025genie3};
AdaWorld~\cite{gao2025adaworld} likewise uses latent actions extracted from
video during pretraining to make world models adaptable to new action spaces
from limited interaction. The Cosmos platform~\cite{nvidia2025cosmos}
formalizes a pretrain-then-post-train workflow, in which a general world
foundation model is fine-tuned into application-specific models. In the
latent (non-generative) line, DINO-WM~\cite{zhou2025dinowm} plans in the
patch-feature space of a frozen DINOv2 encoder; V-JEPA~2~\cite{assran2025vjepa2}
pretrains on over one million hours of video and post-trains an
action-conditioned predictor on less than 62 hours of unlabeled robot video,
which is then deployed zero-shot on Franka arms in two different labs; and
LeWorldModel~\cite{maes2026leworldmodel} trains a JEPA end-to-end from pixels
with only a prediction loss and a Gaussian-embedding regularizer, using about
15M parameters and planning up to $48\times$ faster than world models built
on foundation-model encoders. Terver et al.~\cite{terver2025what}
systematically ablate architecture, objective and planner for such
joint-embedding predictive world models and obtain a model that outperforms
DINO-WM and V-JEPA-2-AC on navigation and manipulation.

The zero-shot deployment of V-JEPA~2-AC on new robots and labs is, in our
taxonomy, a combined observation and dynamics shift absorbed by pretraining.
How far such absorption extends is exactly the question of
\S\ref{sec:pretraining-robustness}, and the evidence is so far negative
outside the training support:
\begin{itemize}[leftmargin=*,itemsep=1pt]
\item Kang et al.~\cite{kang2025far} train video generators on synthetic 2D
physics and separate in-distribution, out-of-distribution (unseen attribute
values) and combinatorial (unseen combinations of seen values) evaluation.
They report near-perfect in-distribution generalization, measurable scaling
for combinatorial generalization, but failure out of distribution that more
data and parameters do not repair; models appear to generalize by retrieving
the closest training case rather than abstracting rules, and prioritize
color over size, size over velocity and velocity over shape when matching.
This is the world-model counterpart of the coverage hypothesis for VLM
robustness~\cite{udandarao2024no,mayilvahanan2025search}.
\item Physics-IQ~\cite{motamed2026physicsiq} (396 videos, 66 scenarios)
finds physical understanding in current video models severely limited and
unrelated to visual realism.
\item ACWM-Phys~\cite{xue2026acwm} evaluates action-conditioned video world
models across eight simulated environments covering rigid, deformable,
particle and kinematic interaction with controlled ID/OOD splits; OOD
degradation depends on interaction regime and task complexity, models rely
on appearance statistics, and increasing model size helps OOD more than ID.
\end{itemize}
Whether the robust fine-tuning lessons of \S\ref{sec:finetuning-tradeoff}
(e.g., that standard fine-tuning erodes the robustness of the pretrained
model~\cite{kumar2022finetuning,wortsman2022robust}) transfer to the
post-training of world foundation models is, to our knowledge, untested.

\subsubsection{Adaptation and Robustness Methods}
\label{sec:wm-adapt}

Table~\ref{tab:wm-works} summarizes the methods discussed here.

\smallskip
\noindent\textbf{Pessimism and uncertainty penalties.} Offline model-based RL
operationalizes Proposition~\ref{prop:wm}(iv): MOPO~\cite{yu2020mopo}
penalizes reward by an estimate of model error, typically ensemble
disagreement, and MOReL~\cite{kidambi2020morel} partitions state--action
space into known and unknown regions. Both are only as good as the error
estimate $u$. Berger et al.~\cite{berger2026biased} show that in
recurrent latent dynamics models ensemble disagreement captures local
uncertainty but can become decoupled from trajectory-level error, which
undermines its use as $u$ over long rollouts. A recent survey reviews the
offline-RL side of this literature in depth~\cite{samani2026distribution}.

\smallskip
\noindent\textbf{Invariant representations.} Observation shift is addressed,
as in DG (\S\ref{sec:dg}), by learning encoders invariant to nuisance
variation. Domain randomization~\cite{tobin2017domain} randomizes rendering
in simulation so that the real world appears as one more variation. Toso et
al.~\cite{toso2026learning} show that DINO-WM is sensitive to slowly varying,
task-irrelevant features such as background changes and distractors, and
train a bisimulation-based encoder whose latent distances respect
behavioral equivalence, yielding robustness to these perturbations.

\smallskip
\noindent\textbf{Test-time adaptation in the control loop.} AdaJEPA~\cite{wang2026adajepa}
plans with the current model, executes the first action chunk, and uses the
observed next-state transition as a self-supervised signal to update the
model before replanning; the authors report substantial gains in planning
success with as few as one gradient step per replanning step.
Sandwich-Residuals~\cite{soni2026sandwich} keeps the world model frozen and
learns small residual corrections around the predictor (about $10^5$
parameters) from the model's own prediction error, reporting
$1.3\times$ the success rate of the frozen model across 21 shift conditions
of the AdaJEPA benchmark (65.0\% versus 49.2\%); the authors explicitly
distinguish appearance from dynamics shift, and report a much smaller gain
on a DINO-WM manipulation task. ALRA~\cite{soni2026alra} goes further for
shifts that rescale how commanded actions translate into motion (damping and
mass changes in PointMaze): a 210-parameter affine residual on the action
embedding matches the strongest AdaJEPA baseline while leaving encoders and
predictor frozen. In vision--language--action models, WorldAgen~\cite{wan2026worldagen}
executes exploratory actions to collect ground-truth state transitions and
applies lightweight test-time training updates to its world-model head.
These results instantiate point (3) of \S\ref{sec:wm-theory}: when the label
is observed, parameter-efficient TTA (\S\ref{sec:surgical}) has a
well-posed objective, and where to adapt (action embedding, predictor,
encoder) can be matched to the shift type.

\smallskip
\noindent\textbf{What prediction-based TTA corrects.} Two late-2026 studies
sharpen the support caveat of \S\ref{sec:wm-theory}. Write the latent
prediction error at state $z$ as
$e(z,a)=f_\theta(z,a)-z^\star_{+}(z,a)=b(z)+r(z,a)$, where
$z^\star_{+}(z,a)$ is the embedding of the observation actually reached,
$b(z)=\E_{a}[e(z,a)]$, the mean over the planner's candidate actions, is the component they all share, and $r(z,a)$ the action-dependent
remainder. For a planner that scores actions by
$\|f_\theta(z,a)-z_g\|^2$, the shared component acts exactly as a shifted
goal $z_g-b(z)$, since
$\|z^\star_{+}+b+r-z_g\|^2=\|z^\star_{+}+r-(z_g-b)\|^2$, whereas only $r$ changes the relative accuracy of competing
actions. Niu et al.~\cite{niu2026what} estimate this decomposition
empirically by executing several actions from the same state on an
object-pushing task, and find that AdaJEPA mainly tracks the shared offset:
correcting only that offset recovers 91\% of AdaJEPA's planning gain. Because
part of the offset is episode-specific while its recurring part and the
action-dependent error are not, they propose a fast learner that tracks the
offset and is reset every episode alongside a slow low-rank learner that
accumulates across episodes, improving on AdaJEPA by up to 22 points of
planning success. Mete et al.~\cite{mete2026adapt} make the complementary
point about the objective: a prediction-only loss fits the model only on the
outcomes of the actions the planner chose, i.e.\ on the support of the
deployment occupancy, and can therefore remain wrong about the actions that
would reach the goal. Their OpTTA adds the planner's goal-reaching cost on
candidate rollouts to the prediction loss, a form of optimism inspired by
reward-biased maximum likelihood in adaptive control, and reports average
gains over the frozen model of 13.5 points versus 6.9 for prediction-only
adaptation across three JEPA world models. Together, these results recast
closed-loop TTA as an exploration problem: the identifiability that
interaction provides is only as broad as the actions the adapted model is
willing to try.

\smallskip
\noindent\textbf{Policy- and planner-side adaptation.} When the world model
is implicit, the same closed-loop logic applies to the policy. TEMPO~\cite{alfarra2026efficient}
adapts locomotion policies online without rewards, using a frozen critic as
a performance proxy with an entropy regularizer against overconfident
extrapolation, and is demonstrated on a physical quadruped carrying an added
payload. In end-to-end driving, P2-TTA~\cite{fang2026coordinating} couples
perception entropy with a retrospective planning-consistency loss computed
from the observed ego motion and rejects updates that increase historical
planning loss---a use of the observed future as a label that parallels the
world-model case.

\smallskip
\noindent\textbf{Modular composition.} For language-model-based embodied
agents, WorMI~\cite{yoo2025world} retrieves domain-specific world models by
prototype matching and composes them with the reasoning model at test time,
and TMoW~\cite{jang2026test} updates the routing function of a mixture of
world models at test time, evaluated on VirtualHome, ALFWorld and RLBench.
These are the world-model counterparts of multi-source and mixture-of-experts
DG (\S\ref{sec:dg}).

\smallskip
\noindent\textbf{Detection and safe fallback.} UNISafe~\cite{seo2025uncertainty}
uses the world model's epistemic uncertainty as a proxy for unseen hazards,
calibrates an uncertainty threshold with conformal prediction to flag OOD
imagined rollouts, and synthesizes a latent safety filter by reachability
analysis in a state space augmented with that uncertainty, evaluated in
simulation and on a Franka manipulator. The calibration step is the
split-conformal machinery of \S\ref{sec:perf-est}; because it is calibrated
on in-distribution data, its coverage guarantee holds only under
exchangeability, and the weighted and adaptive variants
there~\cite{tibshirani2019conformal,gibbs2021adaptive} are the natural
extensions under policy-induced shift. Sinha et al.~\cite{sinha2022system}
argue that OOD data in robotics should be handled at the level of the full
system and its lifecycle rather than by component-level detectors alone.
Detection can also be read out of the generator itself: on NoPeBench, none
of the open-source interactive video world models reliably prevents objects
from passing through obstacles, yet a small head on the frozen diffusion
transformer's hidden states predicts such violations, and replacing flagged
actions reduces the violation rate averaged over six models from 74\% to
16\%~\cite{dai2026nonpenetrating}.

\begin{table*}[t]
\centering
\footnotesize
\setlength{\tabcolsep}{3pt}
\renewcommand{\arraystretch}{1.1}
\caption{Representative works on world models under distribution shift.
Findings are as reported by the authors under their experimental
conditions. ``Obs.'' = observation (appearance) shift; ``Dyn.'' = dynamics
shift; ``Pol.'' = policy-induced occupancy shift; ``Phys.'' =
compositional/extrapolative physical shift.}
\label{tab:wm-works}
\begin{tabular}{P{3.1cm}P{2.6cm}P{1.6cm}P{9.4cm}}
\toprule
\textbf{Work} & \textbf{Model class} & \textbf{Shift} & \textbf{Mechanism or reported finding}\\
\midrule
\multicolumn{4}{l}{\emph{Pessimism and uncertainty}}\\
MBPO~\cite{janner2019trust} (2019) & Ensemble dynamics & Pol. & Short model rollouts branched from real states; return bound in model error and policy shift\\
MOPO~\cite{yu2020mopo}, MOReL~\cite{kidambi2020morel} (2020) & Ensemble dynamics & Pol. & Uncertainty-penalized reward / pessimistic MDP with unknown-region termination\\
Biased Dreams~\cite{berger2026biased} (2026) & RSSM ensembles & Pol. & Ensemble disagreement captures local uncertainty but can decouple from trajectory-level error\\
UNISafe~\cite{seo2025uncertainty} (2025) & Latent dynamics & Novel hazards & Conformal threshold on epistemic uncertainty defines an OOD failure set for a latent safety filter\\
\midrule
\multicolumn{4}{l}{\emph{Representation invariance}}\\
Domain randomization~\cite{tobin2017domain} (2017) & Perception for control & Obs. & Randomized rendering in simulation for sim-to-real transfer\\
Toso et al.~\cite{toso2026learning} (2026) & JEPA world model & Obs. & Bisimulation encoder counters DINO-WM's sensitivity to background changes and distractors\\
\midrule
\multicolumn{4}{l}{\emph{Test-time adaptation and composition}}\\
AdaJEPA~\cite{wang2026adajepa} (2026) & JEPA world model & Obs., Dyn. & Plan--act--adapt--replan; observed transition as self-supervised signal; gains with one gradient step per replanning step\\
Sandwich-Residuals~\cite{soni2026sandwich} (2026) & JEPA world model & Obs., Dyn. & Frozen model plus small residuals around the predictor; $1.3\times$ frozen success on 21 conditions\\
ALRA~\cite{soni2026alra} (2026) & JEPA world model & Dyn. & 210-parameter affine residual on action embedding matches AdaJEPA under damping/mass shifts\\
WorldAgen~\cite{wan2026worldagen} (2026) & VLA with world-model head & Dyn. & Exploratory actions collect ground-truth transitions for lightweight test-time training\\
sft-JEPA~\cite{niu2026what} (2026) & JEPA world models & Obs., Dyn. & Error splits into action-shared offset and action-dependent part; fast/slow learners; up to $+22$ points over AdaJEPA\\
OpTTA~\cite{mete2026adapt} (2026) & JEPA world models & Obs., Dyn. & Prediction loss plus goal-reaching cost (optimism); $+13.5$ vs.\ $+6.9$ points for prediction-only TTA\\
WorMI~\cite{yoo2025world} (2025), TMoW~\cite{jang2026test} (2026) & LM agent + world models & Domain & Test-time retrieval / routing over a mixture of domain-specific world models\\
\midrule
\multicolumn{4}{l}{\emph{Evaluation of generalization}}\\
Kang et al.~\cite{kang2025far} (2025) & Video generation & Phys. & Near-perfect ID, scaling on combinatorial, failure OOD; case-based rather than rule-based generalization\\
Physics-IQ~\cite{motamed2026physicsiq} (2026) & Video generation & Phys. & 396 videos, 66 scenarios; physical understanding limited and unrelated to visual realism\\
ACWM-Phys~\cite{xue2026acwm} (2026) & Action-cond.\ video & Phys. & 8 environments, 4 interaction regimes; OOD gap tracks regime and task complexity\\
WorldLens~\cite{liang2025worldlens} (2025) & Driving world models & Closed loop & 5 aspects; no model excels on all; high open-loop realism does not ensure safe closed-loop control\\
stable-worldmodel~\cite{maes2026stable} (2026) & JEPA world models & Obs., Pol. & Controlled factors of variation; success falls with distance from training data; prediction error weakly predictive of planning success\\
\bottomrule
\end{tabular}
\end{table*}

\subsubsection{Evaluating World Models under Shift}
\label{sec:wm-eval}

A recurring finding of the 2025--2026 evaluation literature is that
open-loop fidelity and closed-loop utility come apart---a sequential
analogue of the gap between average-case metrics and deployment-relevant
ones such as calibration discussed in
\S\ref{sec:metrics-robustness}. WorldLens~\cite{liang2025worldlens}
evaluates driving world models along generation, reconstruction,
action-following, downstream-task and human-preference aspects and finds
that no model excels on all of them and that high open-loop realism does not
guarantee safe closed-loop control. DrivingGen~\cite{zhou2026drivinggen}
benchmarks 14 generative driving world models over weather, time of day,
region and maneuver diversity and reports that general-purpose video models
look better but break physics, while driving-specific ones capture motion
realistically but lag in visual quality. The stable-worldmodel
platform~\cite{maes2026stable} equips control environments with controllable
visual, geometric and physical factors of variation. Under progressively
stronger shift on Push-T, the authors report LeWorldModel's planning success
falling from 88\% on expert training trajectories to 84\% on held-out expert
trajectories, 51\% on random-policy trajectories and 30\% when all factors
of variation are also varied; a single visual factor (canvas color) reduces
success from 50.8\% to 6.0\%; and prediction error correlates poorly with
planning success, leading the authors to attribute failures to OOD inputs
rather than to error magnitude. The drop from expert to random-policy
trajectories with fixed visuals is policy-induced occupancy shift in
isolation; the canvas-color result is pure observation shift.

These results suggest a reporting protocol that mirrors the one we advocate
for static shift (\S\ref{sec:benchmarks}): (i) report ID, combinatorial and
OOD splits separately, per factor of variation, following
Kang et al.~\cite{kang2025far}; (ii) separate observation, dynamics and
occupancy shift, which stable-worldmodel and ACWM-Phys make controllable;
(iii) report closed-loop success, not only prediction error or visual
quality; and (iv) for safety-critical use, report the calibration of the
uncertainty used for detection or pessimism, given the decoupling documented
by Berger et al.~\cite{berger2026biased}.

\begin{remark}[What is new versus relabeled]
Several world-model phenomena are classical shift types with new
parameters: observation shift is conditional shift on the encoder; dynamics
shift is concept shift; compositional physical shift is sub-population shift
with unseen groups. Two features are genuinely different. First, the shift
is endogenous and adversarially selected by the planner
(Proposition~\ref{prop:wm}), which has no counterpart in passive prediction.
Second, interaction supplies labels for the conditional that changed, so
concept shift---the least tractable case of
\S\ref{sec:identifiability}---becomes the most tractable case at test time,
within the visited support.
\end{remark}
\section{Benchmarks, Metrics and Open Problems}
\label{sec:benchmarks}

Every empirical claim in \S\ref{sec:classical}--\S\ref{sec:foundation} rests on
a benchmark and a metric, yet these are usually treated as implementation
details. This section consolidates the benchmarking landscape across
modalities and problems (Table~\ref{tab:benchmarks}) and the metrics used to
score it, making explicit the assumptions each encodes, and closes with the
open problems that follow from the survey as a whole
(\S\ref{sec:challenges}).

\begin{table*}[t]
\centering
\footnotesize
\setlength{\tabcolsep}{3.5pt}
\renewcommand{\arraystretch}{1.07}
\caption{Representative benchmarks by modality and problem. ``Shift type''
uses the taxonomy of Table~\ref{tab:shift-types}. ``Natural'' indicates
shifts arising from real data collection rather than synthetic injection.
This list is representative, not exhaustive.}
\label{tab:benchmarks}
\begin{tabular}{P{1.55cm}P{4.1cm}P{2.45cm}P{3.6cm}cP{3.0cm}}
\toprule
\textbf{Modality} & \textbf{Benchmark} & \textbf{Problem} & \textbf{Shift type} & \textbf{Natural} & \textbf{Reference}\\
\midrule
\multirow{4}{*}{Tabular}
 & WhyShift & DG, shift diagnosis & $Y\mid X$ and covariate & \cmark & \cite{liu2023need,wang2024rethinking}\\
 & TableShift (15 tasks) & DG & label, covariate & \cmark & \cite{gardner2023tableshift}\\
 & TabFSBench & feature shift & feature / schema & \halfmark & \cite{cheng2025tabfsbench}\\
 & Shifts (weather) & uncertainty under shift & temporal, geographic & \cmark & \cite{malinin2021shifts}\\
\midrule
\multirow{6}{*}{\makecell[l]{Image\\(DA/DG)}}
 & Office-31, Office-Home, VisDA, DomainNet & UDA, SFDA & domain (style, synthetic-to-real) & \halfmark & \cite{saenko2010adapting,venkateswara2017deep,peng2017visda,peng2019moment}\\
 & DomainBed (CMNIST, RMNIST, VLCS, PACS, OfficeHome, TerraIncognita, DomainNet) & DG & domain, correlation & \halfmark & \cite{gulrajani2021search,li2017deeper,fang2013unbiased,beery2018recognition}\\
 & NICO++ & DG & context / background & \cmark & \cite{zhang2023nico}\\
 & WILDS (10 datasets) & DG, sub-population & domain, sub-population & \cmark & \cite{koh2021wilds}\\
 & Wild-Time & temporal DG & temporal & \cmark & \cite{yao2022wildtime}\\
 & ACDC & segmentation DA & weather, illumination & \cmark & \cite{sakaridis2021acdc}\\
\midrule
\multirow{3}{*}{\makecell[l]{Image\\(robustness)}}
 & ImageNet-C / -P; CIFAR-10/100-C & robustness, TTA & synthetic covariate & \xmark & \cite{hendrycks2019benchmarking}\\
 & ImageNet-V2, -A, -R, -Sketch, ObjectNet & effective robustness & natural covariate & \cmark & \cite{recht2019imagenetv2,hendrycks2021nae,hendrycks2021many,wang2019imagenetsketch,barbu2019objectnet}\\
 & LAION-Natural / -Rendition & style-strict OOD for web-scale training & domain (style) & \cmark & \cite{mayilvahanan2025search}\\
\midrule
\multirow{2}{*}{\makecell[l]{Sub-\\population}}
 & Waterbirds, CelebA, MultiNLI, CivilComments, MetaShift & group robustness & spurious correlation, imbalance & \halfmark & \cite{sagawa2020distributionally,liang2022metashift,yang2023change}\\
 & BREEDS & sub-population & unseen subclasses & \halfmark & \cite{santurkar2021breeds}\\
\midrule
\multirow{3}{*}{OOD / AD}
 & OpenOOD v1 / v1.5 (near/far, full-spectrum) & OOD detection & semantic + covariate & \halfmark & \cite{yang2022openood,zhang2024openood}\\
 & NINCO & ImageNet OOD (clean) & semantic & \cmark & \cite{bitterwolf2023or}\\
 & MVTec AD & industrial AD & sensory anomaly & \cmark & \cite{bergmann2019mvtec}\\
\midrule
\multirow{4}{*}{Text}
 & Hendrycks et al.\ (7 tasks) & OOD generalization + detection & domain & \cmark & \cite{hendrycks2020pretrained}\\
 & GLUE-X; BOSS & OOD generalization & domain, style & \cmark & \cite{yang2023gluex,yuan2023revisiting}\\
 & Calderon et al.\ (7 tasks, $>$14k shifts) & domain robustness, SD/TD & topical domain & \cmark & \cite{calderon2024measuring}\\
 & Baran et al. & text OOD detection & semantic, background, shuffled & \halfmark & \cite{baran2023classical}\\
\midrule
\multirow{2}{*}{Multimodal}
 & Qiu et al.\ (5 tasks, 33 perturbations) & VLM robustness & image / text perturbation & \xmark & \cite{qiu2024benchmarking}\\
 & DeepBench & domain-specific VLM robustness & LLM-generated corruptions & \xmark & \cite{koddenbrock2025domain}\\
\midrule
Federated & Federated DG benchmark (14 methods) & FedDG & client and domain heterogeneity & \halfmark & \cite{bai2024benchmarking}\\
Graph & GOOD & graph OOD generalization & covariate, concept & \halfmark & \cite{gui2022good}\\
\bottomrule
\end{tabular}
\end{table*}

\subsection{Tabular Benchmarks}
\label{sec:bench-tabular}

WhyShift~\cite{liu2023need,wang2024rethinking} is, to our knowledge, the most
direct attempt to measure shift \emph{type} rather than magnitude on natural
data, which is what makes the $Y\mid X$ finding of \S\ref{sec:yx-shift}
measurable rather than asserted. TableShift~\cite{gardner2023tableshift}
packages 15 tasks with natural shifts behind a common API, and
TabFSBench~\cite{cheng2025tabfsbench} manipulates which features are present
rather than their distribution. All three construct shift from real
partitions (states, years, hospitals, survey waves), distinguishing them from
older tabular DRO evaluations on synthetically subgroup-imbalanced data---part
of why we treat their findings as a corrective to that literature.

\subsection{Image Benchmarks}
\label{sec:bench-image}

\noindent\textbf{Domain adaptation and generalization.}
DomainBed~\cite{gulrajani2021search} packages seven datasets with a fixed
hyperparameter-search and model-selection protocol and shows that, under that
protocol, ERM is competitive with or outperforms most DG algorithms of the
preceding decade---paralleling the tabular finding that DRO does not
consistently beat ERM. We read the two results together as a recurring
pattern: \emph{methodological sophistication that helps on a benchmark
engineered to isolate one shift type does not reliably transfer once
evaluation protocol and model selection are held to the same standard as the
baseline}. WILDS~\cite{koh2021wilds} complements DomainBed with natural domain
and sub-population shifts across ten datasets in medical imaging, wildlife
monitoring, satellite imagery and text, and Wild-Time~\cite{yao2022wildtime}
re-indexes several such datasets by time to operationalize temporal shift.

\smallskip
\noindent\textbf{Natural and synthetic robustness.}
ImageNet-C/P~\cite{hendrycks2019benchmarking} apply 15 algorithmically
generated corruptions (plus held-out ones) at five severities, and are the
standard testbeds for TTA. ImageNet-V2~\cite{recht2019imagenetv2} replicates
the original collection process; ImageNet-A~\cite{hendrycks2021nae} contains
natural adversarial examples; ImageNet-R~\cite{hendrycks2021many} and
ImageNet-Sketch~\cite{wang2019imagenetsketch} contain renditions; and
ObjectNet~\cite{barbu2019objectnet} controls object pose, background and
viewpoint. These are the test sets on which accuracy-on-the-line and CLIP's
departure from it were established. As web-scale corpora increasingly contain
images similar in style to these benchmarks, their ``OOD'' status for
foundation models is no longer guaranteed; LAION-Natural and
LAION-Rendition~\cite{mayilvahanan2025search} were constructed specifically to
restore style-strict OOD evaluation at scale.

\smallskip
\noindent\textbf{OOD detection.} OpenOOD~\cite{yang2022openood} unified the
evaluation of OOD methods and popularized the near/far distinction:
\emph{near-OOD} sets are semantically close to ID (e.g., CIFAR-100 for a
CIFAR-10 model) and \emph{far-OOD} sets are distant (e.g., SVHN or Textures).
OpenOOD v1.5~\cite{zhang2024openood} extends evaluation to ImageNet-scale ID
data and foundation-model backbones (CLIP, DINOv2) and to
\emph{full-spectrum} detection, in which covariate-shifted ID data (e.g.,
ImageNet-C/R/V2) must be accepted while semantically shifted data are
rejected---directly operationalizing Fig.~\ref{fig:ood-plane}.
NINCO~\cite{bitterwolf2023or} addresses a different validity problem: many
standard ImageNet OOD sets contain ID objects, and NINCO provides manually
verified OOD samples.

\subsection{Text and Multimodal Benchmarks}
\label{sec:bench-text}

``Domain'' in text is graded and harder to isolate: topic, register,
dialect and time co-vary. Hendrycks et al.~\cite{hendrycks2020pretrained}
built one of the first systematic NLP OOD benchmarks (seven tasks), and
GLUE-X~\cite{yang2023gluex} and BOSS~\cite{yuan2023revisiting} provide OOD
splits for standard understanding tasks. Calderon et
al.~\cite{calderon2024measuring} construct seven tasks with natural topical
domains that permit measuring both Source Drop and Target Drop. For text OOD
\emph{detection}, Arora et al.~\cite{arora2021types} distinguish background
from semantic shift, and Baran et al.~\cite{baran2023classical} find that
eight classical detectors transferred from vision are not sufficiently
sensitive to all text shift types, with background shift and randomly
shuffled word order within in-domain texts particularly challenging. For
multimodal models, see \S\ref{sec:vlm-benchmarks}.

\subsection{Metrics: Predictive Performance Under Shift}
\label{sec:metrics-performance}

\noindent\textbf{ID/OOD accuracy and effective robustness.} Reporting ID and
OOD accuracy as a pair, and their difference as the \emph{shift gap}, is the
most common practice. Effective robustness~\eqref{eq:effrob} removes the part
of OOD accuracy predicted by ID accuracy~\cite{taori2020measuring}; its value
depends on the reference population defining $\beta$ and, as Shi et
al.~\cite{shi2023effective} show, on the choice of ID test set(s).

\smallskip
\noindent\textbf{Worst-group accuracy.} For sub-population shift, the
standard metric is $\min_g\mathrm{acc}(\mathcal{D}_g)$ over a partition of the
test set into known groups~\cite{sagawa2020distributionally}. It inherits the
group definition as a modeling assumption and is undefined without it;
worst-class accuracy is a group-free surrogate~\cite{yang2023change}.

\smallskip
\noindent\textbf{Source Drop and Target Drop.} Let $SS$ and $TT$ denote
in-domain performance on source and target (training and testing on the same
domain), and $ST$ the cross-domain performance (train on source, test on
target). Calderon et al.~\cite{calderon2024measuring} define
\begin{equation}
\mathrm{SD}=SS-ST,\qquad \mathrm{TD}=TT-ST.
\label{eq:sdtd}
\end{equation}
SD conflates genuine robustness loss with the target domain being
intrinsically harder; TD measures the loss relative to what is achievable on
the target. In a significant proportion of shifts only one of the two is
positive, so reporting either alone can lead to incorrect conclusions.

\smallskip
\noindent\textbf{Label-free estimation error.} Performance-estimation methods
(\S\ref{sec:perf-est}) are scored by the absolute error between predicted and
true target accuracy, averaged over shifts~\cite{garg2022leveraging,baek2022agreement}.

\smallskip
\noindent\textbf{Efficiency.} For TTA, accuracy alone is insufficient;
Wang et al.~\cite{wang2025search} report GFLOPs, wall-clock time and memory,
which separate training-free (\S\ref{sec:fm-tta}) and forward-only methods
(\S\ref{sec:tta}) from backpropagation-based ones.

\subsection{Metrics: OOD and Anomaly Detection}
\label{sec:metrics-detection}

Detection is a binary decision and is evaluated with threshold-free and
fixed-operating-point metrics: \textbf{AUROC}, the probability that a random
OOD sample scores above a random ID sample; \textbf{FPR@95TPR}, the fraction
of OOD samples accepted when 95\% of ID samples are accepted, which reflects a
deployment-relevant operating point; and \textbf{AUPR}, reported with ID or
OOD samples as the positive class, preferable under class imbalance. These
metrics are common to OpenOOD~\cite{yang2022openood,zhang2024openood} and most
of the methods in Table~\ref{tab:ood}; what differs across papers---and what
near/far and full-spectrum protocols make visible---is \emph{which} ID/OOD
pair the metric is computed on, an implicit statement about which axis of
shift is being targeted (\S\ref{sec:ood-connection}).

\subsection{Metrics: Calibration and Uncertainty Under Shift}
\label{sec:metrics-robustness}

A model can be accurate on average while being confidently wrong on the
inputs it misclassifies. The expected calibration error,
\begin{equation}
\mathrm{ECE}=\sum_{m=1}^M\frac{|B_m|}{n}\,\big|\mathrm{acc}(B_m)-\mathrm{conf}(B_m)\big|,
\label{eq:ece}
\end{equation}
bins predictions by confidence~\cite{naeini2015obtaining,guo2017calibration};
Guo et al.\ also popularized temperature scaling as a post-hoc correction.
ECE is sensitive to the binning scheme and biased in finite
samples~\cite{nixon2019measuring,kumar2019verified}, motivating adaptive
binning and debiased estimators; reported values should be read together with
the variant used. Under shift, Ovadia et al.~\cite{ovadia2019can} find that
calibration of many uncertainty methods degrades substantially as shift
intensity increases, and that post-hoc calibration on ID validation data does
not transfer; Minderer et al.~\cite{minderer2021revisiting} find that some
recent architectures (non-convolutional, large-scale pretrained) are better
calibrated and that their calibration degrades more slowly under shift.
Calibration can be explicitly adapted using importance weights from
unlabeled target data~\cite{park2020calibrated}, and coverage guarantees
under shift are provided by the conformal methods of \S\ref{sec:perf-est}.
For VLMs, test-time prompt tuning can harm calibration, which motivates
calibration-aware objectives~\cite{yoon2024ctpt}.

\subsection{A Cross-Cutting Observation}
\label{sec:benchmark-observation}

In tabular data (WhyShift/TableShift versus earlier DRO evaluations), image
DG (DomainBed), sub-population shift (``Change is hard''), test-time
adaptation (TTAB, RDumb) and federated DG, the introduction of a more
rigorous protocol has coincided with a downward revision of the benefit of
specialized methods over well-tuned
baselines~\cite{wang2024rethinking,gulrajani2021search,yang2023change,zhao2023pitfalls,press2023rdumb,bai2024benchmarking}.
We do not conclude that robustness methods do not work---OpenOOD's finding
that no single detector wins across all configurations shows that method
choice matters, just not uniformly---but we do conclude that a reported
robustness gain should be assessed against a baseline evaluated under the
\emph{same} hyperparameter-search and model-selection rigor, and that the
model-selection procedure (which data, labeled or not, from which domain)
must be reported as part of the method.
\subsection{Open Challenges and Future Directions}
\label{sec:challenges}

We close with open problems that follow from the gaps of \S\ref{sec:gaps}
and the technical content of \S\ref{sec:taxonomy}--\S\ref{sec:worldmodels}.
They are grouped by the section from which they arise rather than ranked,
since we do not believe the field has reached consensus on priorities.

\subsubsection{Diagnosing Shift Type at Deployment}
\label{sec:challenge-yx}

The $Y\mid X$ finding (\S\ref{sec:yx-shift}) was established post hoc with
target labels. Two directions follow. First, replicate the
covariate/conditional decomposition~\cite{cai2023diagnosing,liu2023need} on
high-dimensional vision, language and VLM benchmarks to test whether the
imbalance holds outside tabular data. Second, develop \emph{deployment-time}
diagnostics that use unlabeled target data plus a minimal labeling budget to
decide between reweighting, retraining and abstention. Proposition~\ref{prop:nonid}
implies that some labels (or an explicit structural assumption) are
unavoidable; the open question is how few, and chosen how. Active-labeling
strategies that target the covariate regions with the largest estimated
$Y\mid X$ shift~\cite{liu2023need} are a natural starting point.

\subsubsection{A Shared Benchmark for Drift Detection and Test-Time Adaptation}
\label{sec:challenge-streaming-tta}

Classical detectors (DDM, ADWIN, risk monitoring~\cite{podkopaev2022tracking})
decide \emph{when} to adapt; online TTA methods (Tent, EATA, SAR, CoTTA)
decide \emph{how}. The strength of periodic resets~\cite{press2023rdumb}
suggests that adaptation schedules matter as much as adaptation objectives.
A shared benchmark spanning tabular streams and visual streams, with delayed
labels, recurring concepts and explicit compute budgets~\cite{wang2025search},
would allow detector-gated TTA to be evaluated against continuous adaptation.
The main barrier is infrastructural.

\subsubsection{Predicting the Robustness Cost of Adapting Foundation Models}
\label{sec:challenge-fm}

Full fine-tuning, LP-FT, weight interpolation, prompt and adapter tuning,
training-free caches and in-context conditioning trade task accuracy against
retained effective robustness in different ways (Table~\ref{tab:vlm-adapt}).
Is there a principled way to predict, \emph{before} adaptation, how much
robustness a mechanism will sacrifice for a given ID gain, given the
pretraining distribution, the adaptation data and the anticipated shift? The
feature-distortion analysis of LP-FT~\cite{kumar2022finetuning} and linear mode
connectivity behind WiSE-FT~\cite{wortsman2022robust} are partial answers for
specific mechanisms; a unified account is missing.

\subsubsection{What Is ``Out-of-Distribution'' for a Foundation Model?}
\label{sec:challenge-coverage}

The evidence of \S\ref{sec:pretraining-robustness}---log-linear concept-frequency
scaling~\cite{udandarao2024no}, domain contamination~\cite{mayilvahanan2025search},
and the dependence of effective robustness on the ID reference~\cite{shi2023effective}---implies
that OOD evaluation of web-scale models requires measuring the overlap between
pretraining data and test domains. Open problems include scalable estimators of
domain and concept coverage, benchmarks that are certified to be
style- and concept-disjoint from common pretraining corpora, and a theory
that connects coverage to the support-overlap conditions of classical
importance weighting.

\subsubsection{Evaluation Validity for OOD Detection Under Mixed Shift}
\label{sec:challenge-ood-eval}

Detection scores implicitly target one axis of Fig.~\ref{fig:ood-plane}.
Protocols that report performance separately under covariate-only,
semantic-only and mixed conditions---as full-spectrum
evaluation~\cite{yang2023full,zhang2024openood} begins to do---should become
standard, and OOD datasets should be verified to be free of ID
content~\cite{bitterwolf2023or}. For VLMs, where the ID set is defined by the
prompt, it is open how to evaluate detectors whose behavior depends on the
choice of class names and negative vocabularies~\cite{jiang2024negative}.

\subsubsection{Open-World Test-Time Adaptation}
\label{sec:challenge-openworld-tta}

Real streams combine covariate shift, label-distribution shift, temporal
correlation and semantically novel inputs. Recent work begins to relax the
i.i.d., all-classes-present assumption~\cite{zanella2025realistic,niu2023towards}
and to integrate detection with adaptation~\cite{cao2025noisy}, and
prior-adaptive VLM TTA~\cite{zhou2025bayesian} reconnects TTA with label-shift
correction. A principled framework that jointly estimates the target prior,
detects semantic novelty and adapts to covariate shift---with guarantees
analogous to BBSE~\cite{lipton2018detecting}---remains open.

\subsubsection{Federated Shift: Detection, Diagnosis and Model Selection}
\label{sec:challenge-fl}

Shift detection, density-ratio estimation and DG model selection all assume
pooled data that FL forbids (\S\ref{sec:fl-open}). Privacy-preserving
two-sample tests and shift decompositions computed from model updates,
federated analogues of label-free performance estimation (\S\ref{sec:perf-est})
for unseen clients, and controlled benchmarks that vary client count and
heterogeneity~\cite{bai2024benchmarking} are concrete, tractable directions.
Test-time adaptation at clients~\cite{jiang2023test,bao2023adaptive} raises the
additional question of whether locally adapted models should feed back into
the global model.

\subsubsection{Feature Shift and Open-Environment Tabular Learning}
\label{sec:challenge-feature-shift}

Feature shift has a benchmark~\cite{cheng2025tabfsbench} but no mature
methodology. Natural directions are to extend invariant feature selection
(\S\ref{sec:tabular-remediation}) to settings where features can disappear,
and to condition in-context tabular models~\cite{hollmann2025accurate,helli2024driftresilient}
on the set of features present at inference rather than assuming a fixed
schema.

\subsubsection{Certifiable Guarantees Under Realistic Shift}
\label{sec:challenge-certification}

Domain-adaptation bounds depend on divergences that are unknown and only
loosely estimable. Two complementary tools offer a path to auditable
guarantees for regulated deployments: distribution-free coverage under
specified shift classes~\cite{tibshirani2019conformal,podkopaev2021distribution,barber2023conformal}
and sequential risk monitoring~\cite{podkopaev2022tracking}. Extending these
from covariate and label shift to bounded conditional shift, and to
foundation-model predictions with prompt-defined label spaces, is open.

\subsubsection{Distribution Shift in World Models}
\label{sec:challenge-wm}

Section~\ref{sec:worldmodels} suggests four concrete problems. (i)
\emph{Estimating the policy-induced ratio.} Proposition~\ref{prop:wm}(iii)
reduces model-based decision error to a density ratio between planner and
data occupancies, but planners do not estimate it; the ratio-estimation and
support-overlap tools of \S\ref{sec:classical-correction} have not been
brought into planning in latent world models, where the relevant ratio is
over learned representations. (ii) \emph{Calibrated pessimism.} Pessimism is
only as good as the error proxy $u$, and ensemble disagreement can decouple
from trajectory-level error~\cite{berger2026biased}; weighted and adaptive
conformal methods~\cite{tibshirani2019conformal,gibbs2021adaptive} offer
guarantees under covariate shift and drift, but extending them to the
endogenous, adversarially selected shift of planning is open. (iii)
\emph{When to adapt, and what.} World-model TTA has a supervised signal
(\S\ref{sec:wm-theory}), and residuals placed in action space suffice for
actuation shifts~\cite{soni2026alra}; deciding online whether an error stems
from observation or dynamics shift, and therefore which component to adapt,
is a shift-diagnosis problem in the sense of \S\ref{sec:challenge-yx}. When
the prediction target is an encoder output, preventing adaptation from
drifting the target itself is also unresolved. Because a prediction loss is
informative only about executed actions, closed-loop TTA inherits an
exploration problem; optimistic objectives~\cite{mete2026adapt} and
fast/slow decompositions of the error~\cite{niu2026what} are first answers,
but neither comes with guarantees on when adaptation helps rather than
reinforces the planner's own bias. (iv) \emph{Physical OOD
generalization.} The finding that video world models generalize by case
rather than by rule~\cite{kang2025far}, and that physical understanding is
unrelated to visual realism~\cite{motamed2026physicsiq}, poses the coverage
question of \S\ref{sec:challenge-coverage} for dynamics: whether any amount of
passive video makes extrapolation possible, or whether interaction and
structural priors are necessary.

\subsubsection{Beyond Vision, Tabular and Language}
\label{sec:challenge-modalities}

Graph data (where topology itself can shift~\cite{zhang2024survey,gui2022good,wu2022handling}),
time series~\cite{wu2025tsood}, and model-free offline reinforcement learning (where the
shift is between the behavior policy's and the learned policy's state--action
distributions~\cite{levine2020offline,prudencio2024survey,samani2026distribution}) present related
but distinct phenomena. Because the taxonomy of \S\ref{sec:taxonomy} is
stated at the level of factorizations of a joint distribution, extending it
to these settings---e.g., to factorizations over graph structure and node
attributes, or over states, actions and returns---is a natural next step.
\section{Conclusion}
\label{sec:conclusion}

We have argued that classical correction, domain adaptation and
generalization, test-time adaptation, federated learning, OOD detection, and
the robustness of foundation and world models study one object---a change
in $P(X,Y)$ between training and deployment---under different
factorizations and different assumptions about what target information is
available, and when. Three consequences stand out. Every adaptation method
commits to a structural assumption that the data cannot verify
(Proposition~\ref{prop:nonid}), so diagnosing the shift type, not choosing a
method, is the bottleneck; the dominance of $Y\mid X$ shift in natural
tabular benchmarks and the repeated shrinkage of method gains under
rigorous evaluation make this concrete. Foundation models change the
landscape rather than adding a domain, yet the debate over the origin of
their robustness is best settled in classical terms of support overlap.
Finally, world models make shift endogenous: the planner selects its own
test distribution (Proposition~\ref{prop:wm}), while interaction makes
dynamics shift identifiable on the states it visits. We hope the taxonomy,
the comparison of \S\ref{sec:gaps} and the problems of
\S\ref{sec:challenges} help connect these communities.

\ifCLASSOPTIONcompsoc
  \section*{Acknowledgments}
\else
  \section*{Acknowledgment}
\fi
The authors thank the anonymous reviewers for their feedback.

\bibliographystyle{IEEEtran}
\bibliography{references}

\begin{IEEEbiographynophoto}{Behraj Khan}
is a Ph.D. candidate at the Institute of Business Administration (IBA),
Karachi, Pakistan. His research focuses on learning under distribution
shift, including federated learning under covariate shift, test-time
adaptation, and the robustness of vision--language models.
\end{IEEEbiographynophoto}

\begin{IEEEbiographynophoto}{Tahir Qasim Syed}
is with the Institute of Business Administration (IBA), Karachi, Pakistan.
\end{IEEEbiographynophoto}

\end{document}